\documentclass[final,5p,times,twocolumn]{elsarticle}

\usepackage{amssymb}
\usepackage{amsmath}

\usepackage[english]{babel}
\usepackage{hyperref}
\usepackage{subcaption}
\usepackage{footnote} 
\usepackage{xcolor} 
\usepackage{float}
\usepackage{placeins}
\usepackage{tabularx}

\journal{Engineering Applications of Artificial Intelligence}

\begin{document}

\begin{frontmatter}



\title{A Survey on State-of-the-art Deep Learning Applications and Challenges} 


\author[1]{Mohd Halim Mohd Noor \corref{cor1}}
\ead{halimnoor@usm.my}

\author[2]{Ayokunle Olalekan Ige}
\ead{ayo.ige@aaua.edu.ng}

\cortext[cor1]{Corresponding author}

\affiliation[1]{organization={School of Computer Sciences, Universiti Sains Malaysia},
            addressline={Jalan Universiti}, 
            city={Gelugor},
            postcode={11800}, 
            state={Pulau Pinang},
            country={Malaysia}}
			
\affiliation[2]{organization={Department of Computer Science, Adekunle Ajasin University},
            addressline={Ikare Akoko-Ipeme Rd}, 
            city={Akungba-Akoko},
            postcode={P.M.B 001}, 
            state={Ondo State},
            country={Nigeria}}

\begin{abstract}
Deep learning, a branch of artificial intelligence, is a data-driven method that uses multiple layers of interconnected units or neurons to learn intricate patterns and representations directly from raw input data. Empowered by this learning capability, it has become a powerful tool for solving complex problems and is the core driver of many groundbreaking technologies and innovations. Building a deep learning model is challenging due to the algorithm’s complexity and the dynamic nature of real-world problems. Several studies have reviewed deep learning concepts and applications. However, the studies mostly focused on the types of deep learning models and convolutional neural network architectures, offering limited coverage of the state-of-the-art deep learning models and their applications in solving complex problems across different domains. Therefore, motivated by the limitations, this study aims to comprehensively review the state-of-the-art deep learning models in computer vision, natural language processing, time series analysis and pervasive computing, and robotics. We highlight the key features of the models and their effectiveness in solving the problems within each domain. Furthermore, this study presents the fundamentals of deep learning, various deep learning model types and prominent convolutional neural network architectures. Finally, challenges and future directions in deep learning research are discussed to offer a broader perspective for future researchers.
\end{abstract}




\end{frontmatter}



\section{Introduction}\label{sec1}
Deep learning has revolutionized many applications across a variety of industries and research. The application of deep learning can be found in healthcare \cite{shamshirband_review_2021}, smart manufacturing \cite{wang_deep_2018}, robotics \cite{pierson_deep_2017} and cybersecurity \cite{dixit_deep_2021}, solving challenging and complex problems such as disease diagnosis, anomaly detection, object detection and malware attack detection. Deep learning is a subset of machine learning that focuses on learning from data using artificial neural networks with many layers, known as deep neural networks. An artificial neural network is a computational model that imitates the working principles of a human brain. The computational models are composed of an input layer which receives the input data, multiple processing layers that learn the representation of data and the output layer which produces the output of the model.

Prior to the reintroduction of deep learning (DL) into the research trend, pattern recognition tasks involved a transformation of the raw input data such as pixel values of an image into a feature vector that represents the internal representation of the data. The feature vector can be used by a machine learning model to detect or classify patterns in the data. This process requires feature engineering and considerable domain knowledge to design a suitable feature representation. With deep learning, this cumbersome process can be performed automatically whereby at each processing layer known as hidden layers, the internal representation of the input data is learned or extracted in a hierarchical manner. The first layer learns the presence of basic primitive features such as edges, dots, and lines. The second layer learns patterns or motifs by recognizing the combinations of the edges, dots and lines, and the subsequent layers combine the motifs to produce more sophisticated features that correspond to the input data. This feature learning process takes place in the sequence of hidden layers until the prediction is finally produced. 

Deep learning has seen numerous breakthroughs in various industries, transforming how complex problems are solved and how businesses operate. For instance, protein folding is a complex process, influenced by the intricate sequence of amino acids. Using traditional methods to determine how a protein folds would take an immense amount of computational power and time due to its complex 3D structure and vast number of possible configurations. AlphaFold \citep{noauthor_alphafold_2025} has made significant progress in this area, using deep learning to predict protein structures with remarkable accuracy, helping researchers to comprehend the complete structure of a key protein associated with diseases like malaria and Parkinson's disease. In natural language processing, systems like OpenAI GPT, Google Gemini and IBM Watsonx have revolutionized chatbots and virtual assistants, enabling them to understand and respond to human language with remarkable accuracy and contextual awareness. These systems have overcome the challenge of processing vast amounts of unstructured text, allowing them to engage in more natural conversations and handle a wide range of topics. Businesses can use these systems to automate customer service, provide real-time support and improve user experience.

Several studies have been conducted to discuss the concept and application of deep learning in the last few years, as listed in Table 1. The studies addressed or focused on several aspects of deep learning, such as types of deep learning models, learning approaches and strategies, convolutional neural network (CNN) architectures, deep learning applications and challenges. In \citep{dong_survey_2021}, the authors provided fundamentals of deep learning and highlighted different types of deep learning models, such as convolutional neural networks, autoencoder and generative adversarial networks. Then, the applications of deep learning in various domains are discussed, and some challenges associated with deep learning applications are presented. Another survey \citep{talaei_khoei_deep_2023} provided a comprehensive analysis of supervised, unsupervised and reinforcement learning approaches and compared the different learning strategies such as online, federated and transfer learning. Finally, the current challenges of deep learning and future direction are discussed.

In \citep{alzubaidi_review_2021}, the authors provided a comprehensive review of the popular CNN architectures used in computer vision tasks, highlighting their key features and advantages. Then, the applications of deep learning in medical imaging and the challenges are discussed. A similar survey is reported in \citep{alom_state_art_2019}, where the different supervised and unsupervised deep learning models are highlighted, and the popular CNN architectures are compared and discussed. In another survey \citep{pouyanfar_survey_2018}, the authors focused on the applications of deep learning in computer vision, natural language processing and speech and audio processing. The different types of deep learning models are also discussed. In \citep{sarker_deep_2021}, the authors focused on the different types of deep learning models and provided a summary of deep learning applications in various domains.

Despite the existing surveys on deep learning that offer valuable insights, the increasing amount of deep learning applications and the existing limitations in the current studies motivated us to explore this topic in depth. In general, to the best of our knowledge, no survey paper focuses on the emerging trends in state-of-the-art applications and the current challenges associated with deep learning. Furthermore, the surveys do not discuss the issues and how deep learning addresses them by highlighting the key features and components in the models. Furthermore, most surveys either ignore or provide minimal coverage of the fundamentals of deep learning, which is crucial for understanding the state-of-the-art models. The main objective of this paper is to present the most important aspects of deep learning, making it accessible to a wide audience and facilitating researchers and practitioners in advancing and leveraging its capabilities to solve complex problems across diverse domains. Specifically, we present the fundamentals of deep learning and the various types of deep learning models, including popular deep learning architectures. Then, we discuss the progress of deep learning in state-of-the-art applications, highlighting the key features of the models and their problem-solving approaches. Finally, we discuss the challenges faced by deep learning and the future research directions.

To this end, we conducted automatic search strategies using ``deep learning'' keyword with keywords related to deep learning applications such as ``image classification'', ``neural machine translation'', ``text generation'', ``human activity recognition'' and ``robotics''. Then, a manual search was carried out by scanning the references produced by the automatic search, selecting relevant works and discarding the irrelevant ones. This survey paper collected primary studies from journals, conference proceedings, and books in English only. We use scientific databases such as IEEE Explore, ScienceDirect, SpringerLink, ACM Digital Library, Scopus and ArXiv. The ArXiv repository is used because it hosts numerous manuscripts that are highly relevant to the topics discussed in this survey. These preprints often contain novel findings and methodologies that may not be available in other repositories at the time of writing. Nevertheless, we ensure that the selected preprints are of high quality, authored by researchers from reputable institutions.

\renewcommand{\arraystretch}{1.2}  
\begin{table*}[htbp]
\centering
\caption{Summary of related works.}
\scriptsize
\begin{tabularx}{\textwidth}{p{1.5cm} X X}
 \hline
 Ref. & Focus & Concepts not covered \\ 
 \hline
 \citep{dong_survey_2021}, 2021 & A short review of the fundamentals of DL networks and discusses different types of neural networks, DL applications and challenges. & Lack of analysis of CNN architectures and limited coverage of deep learning fundamentals. \\ 
 \hline
 \citep{talaei_khoei_deep_2023}, 2023 & Discusses the learning approaches (supervised, unsupervised and reinforcement learnings), learning strategies, and DL challenges & Lack of fundamentals of deep learning, CNN architectures and DL applications. \\
 \hline
\citep{alzubaidi_review_2021}, 2021 & Discusses different types of DL networks, CNN fundamentals and architectures, DL challenges and medical imaging applications & Limited discussion on DL applications such as natural language processing and time series analysis. \\
 \hline
 \citep{alom_state_art_2019}, 2019 & A short review of the fundamentals of neural networks and discusses different types of DL networks, CNN architectures and applications. & Limited discussion on DL applications and no discussion of DL challenges. \\
 \hline
 \citep{pouyanfar_survey_2018}, 2018 & Discusses different types of DL networks and DL applications and challenges & Lack of analysis of CNN architectures and limited coverage of deep learning fundamentals. \\ 
 \hline
 \citep{sarker_deep_2021}, 2021 & Discusses different types of DL networks and provides a summary of DL applications & Lack of fundamentals of deep learning, analysis of CNN architectures and limited discussion on DL applications. \\ 
 \hline
\end{tabularx}
\label{table_summary_related_works}
\end{table*}

The remainder of this paper is organized as follows: Section 2 describes the fundamentals of deep learning which includes layers and attention mechanisms, activation functions, model optimization and loss functions, and regularization methods. Section 3 presents the types of deep learning models, including the CNN architectures. Section 4 discusses the state-of-the-art applications of deep learning. Section 5 discusses the challenges and future directions in the field of deep learning. The conclusion is given in Section 6.

\section{Fundamentals of Deep Learning}\label{sec2}
This section describes the fundamental concepts such as layer types, activation functions, training algorithms and regularization methods to provide a comprehensive understanding of the underlying principles in advancing the field of deep learning.

\subsection{Layers}
\begin{figure}[h!]
\centering
\includegraphics[scale=0.5]{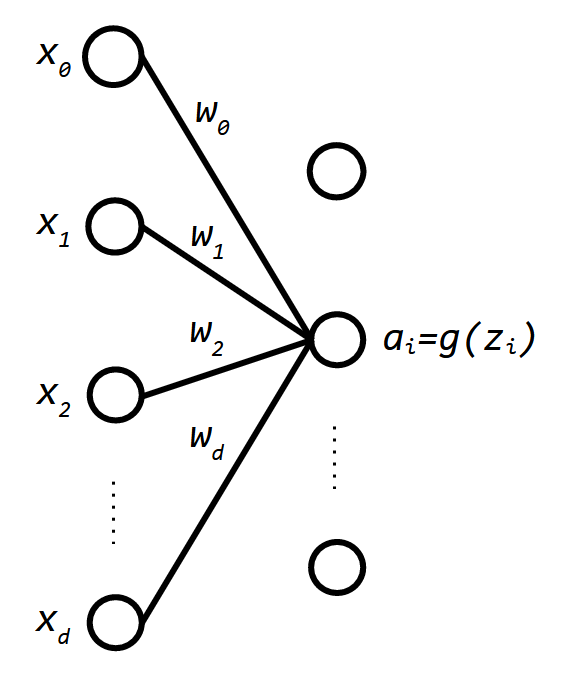}
\caption{A graphical representation of a neuron.}
\label{fig_layers_neuron}
\end{figure}

A deep learning model is characterized by having numerous hidden layers. The hidden layers are responsible for learning and extracting complex features from the input data. A hidden layer is composed of an arbitrary number of neurons which serves as the fundamental building block of a neural network as shown in Fig. \ref{fig_layers_neuron}. A neuron consists of an arbitrary number of inputs, each associated with a weight, which controls the flow of information into the neuron during the forward pass. The flow of information, or forward pass, involves the computation of summation of the weighted input, followed by the application of a transformation function to the weighted sum. Let $\mathbf{x} = {x_1, x_2, ...,x_d}, j=0,1,...,d$ be the input vector with $d$ dimensions and $\mathbf{w}_i^l$ denotes the weights that are connected to neuron $i$ in layer $l$. The forward pass to neuron $i$ in layer $l$ is defined as  

\begin{equation}
z_i^l=\mathbf{w}^{l}_i \cdot \mathbf{x}=\sum_{j=0}^{d} w_{i,j}^{l} \cdot x_{j}
\end{equation}

\begin{equation}
a_i^l=g(z_i^l)
\end{equation}

\noindent where $x_{j}$ is the input vector of size $d$, $w_{i,j}^{l}$ is the weight associated with $x_j$, connecting the input to neuron $i$ at layer $l$, and $g$ is the transformation function also known as activation function. It is worth noting that $w_0$ is called bias and $x_0=1$. A hidden layer wherein each neuron is connected to all neurons of the previous layer is known as the fully connected layer. The forward pass computation of a neuron $a_i^{l}$ in layer $l$ receiving a set of input from layer $l-1$  can be generalized as follows:

\begin{equation}
a_i^l=g(\sum_{j=0}^{d} w_{i,j}^{l} \cdot a_{j}^{l-1})=\mathbf{w}^{l}_i \cdot \mathbf{a}^{l-1}
\end{equation}

\noindent where $\mathbf{a}^{l-1}$ is the input vector from layer $l-1$.

\begin{figure}[h!]
\centering
\includegraphics[scale=0.4]{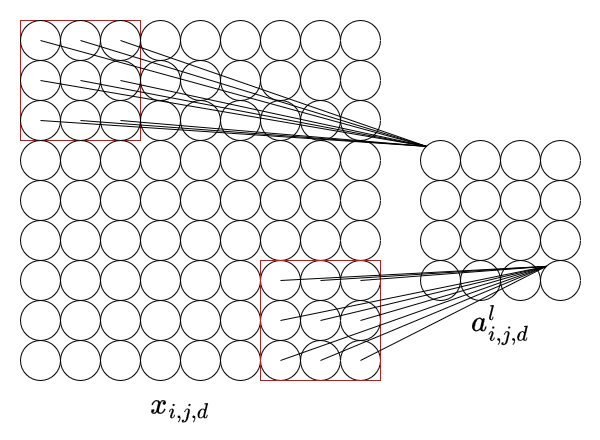}
\caption{A neuron is connected to a local region of the input data.}
\label{fig_layers_convolution}
\end{figure}

Another crucial layer type is the convolutional layer, which is primarily used for processing data that has underlying structures, such as spatial patterns in image data or temporal patterns in time series data. Unlike a fully connected layer, each neuron in a convolutional layer is connected only to a subset of neurons in the previous layer, as shown in Fig. \ref{fig_layers_convolution}. As shown in the figure, each neuron in the hidden layer is connected only to a local region, a subset of nine input neurons, and the weights are shared across the input data. The weight-sharing not only significantly reduces the number of parameters of the neural network, but also allows the network to learn the same features across different spatial locations in the input \citep{lecun_deep_2015}. Let $\mathbf{x}$ be a two-dimensional input data with $H \times W \times C$ such as a grayscale image, where $H$ is the height, $W$ is the width and $C$ is the channel of the image data. The computation of a neuron in the convolutional layer $l$ is defined as

\begin{equation}
z_{i,j,d}^l=\sum_{m=0}^{k_1} \sum_{n=0}^{k_2} w_{m,n,d}^{l} \cdot x_{i\cdot s +m,j\cdot s+n,d}
\end{equation}

\begin{equation}
a_{i,j,d}^l=g(z_{i,j,d}^l)
\end{equation}

\noindent where $w_{m,n,d}$ is the weight connecting the input data $x_{i,j,d}$ within the window to the neuron in layer $l$ and $s$ is the stride or the step size of the window as it moves across the input data. This convolution operation where the filter is represented by $w_{m,n,d}$ slides over the input data $x_{i,j,d}$, producing a set of output values called feature map. For a more hands-on understanding of this convolutional operation, readers are encouraged to explore CNN Explainer \citep{noauthor_poloclubcnn-explainer_2025}, an interactive tool that demonstrates how convolutional layers process input data and generate feature maps.

\begin{figure}[h!]
\centering
\includegraphics[scale=0.4]{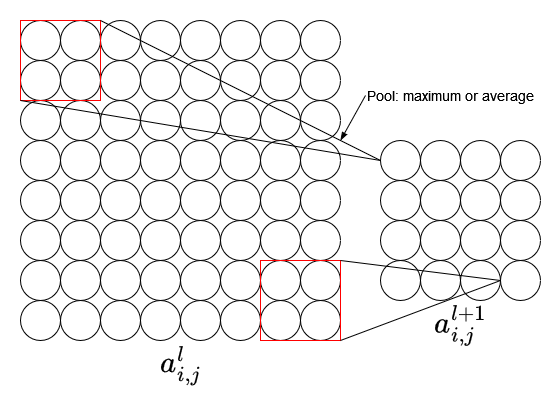}
\caption{Summaries of the feature map using maximum or average pooling to produce a reduced feature map.}
\label{fig_layers_pooling}
\end{figure}

Pooling layers are commonly applied after successive convolutional layers to progressively reduce the spatial dimensions of the feature maps. The spatial reduction is performed by computing the summary of a subset of the feature values in the feature maps as shown in Fig. \ref{fig_layers_pooling}. The pooling operation, as shown in the figure, can use either the maximum or average method with a pooling size of $2 \times 2$ applied across the entire feature map, thereby reducing the size of the feature map. In addition to spatial reduction, pooling layers decrease the number of parameters and provide translation-invariant features \citep{lecun_deep_2015}. Let $\mathbf{a}^l$ denotes feature map (hidden layer) $l$, a pooling operation with a pooling size of $m \times n$ is defined as

\begin{equation}
\mathbf{a}^{l+1}=\texttt{pool}(\mathbf{a}^l_{i\cdot s+m,j\cdot s+n})
\end{equation}

\noindent where $\texttt{pool}$ is either maximum or average function and $s$ is the stride or step size of the window as it moves across the feature map.

\subsection{Attention Mechanisms}
One of the important concepts in pattern recognition is the ability to attend to and neglect certain parts of the input data based on their relevance. This is because not all parts of the input hold equal importance for making predictions. Certain features exhibit a stronger correlation with the output while others are less relevant. To provide a simple analogy, consider a set of sensors to measure room temperature. The sensors are deployed across different corners of the room. Each sensor measures temperature in its own area, and these local measurements are used to estimate the room temperature. However, not all sensors are equally important due to various factors. For instance, a sensor near an air-conditioning or a heater might give inaccurate readings due to the external temperature fluctuation caused by these systems. A sensor near the center of the room, away from the heating or cooling sources might provide more reliable readings. Therefore, when estimating the room temperature, we give more attention to the sensors that are less affected by the localized factors. In convolutional layers, all extracted features are treated uniformly, without consideration of the varying degree of the importance of the different parts of the input data. This limitation is addressed by the introduction of attention mechanism, which can dynamically assign varying levels of significance (weights) to the different features. This flexibility enables the deep learning models to prioritize the more relevant aspects of the input data, enhancing its ability to capture the intricate dependencies for accurate prediction. Given an input data $x$, the process of attending to the important components of the input is given as

\begin{equation}
A=f(g(x),x)
\end{equation}

\noindent where $g$ is a composite function that performs a sequence of operations to generate the attention or the weights and $f$ applies the generated attention $g(x)$ on the input $x$. 

For instance, the squeeze-and-excitation (SE) attention generates the attention through five consecutive operations \citep{hu_squeeze-and-excitation_2019}. First, the input is vectorized using global average pooling. Then the vector is passed to two fully connected layers, where the first one with ReLU activation and the second one with sigmoid activation. SE attention was a pioneer in channel attention. The attention module assigns varying weights to the channels of the feature maps. SE attention suffers from computational cost and the use of global average, which may cause information loss at the spatial level. Several efforts have been made to improve SE attention. Global Second-order Pooling (GSoP) attention performs $1 \times 1$ convolution on the feature maps to reduce the number of channels, and then computes the pairwise channel correlation, which is used to generate the weights \citep{gao_global_2018}. Efficient Channel Attention (ECA) replaced the fully connected layers with 1D convolution to reduce the number of parameters and the computational cost \citep{wang_eca-net_2020}.

Temporal attention is an attention module that focuses on specific time steps in a sequence of data such as time series and video (sequence of images). In video processing such as recognizing human actions, temporal attention is used to focus on key frames at different point in time that contain crucial information for predicting the ongoing activity. Temporal adaptive module (TAM) is a temporal attention that can focus on short-term (local) information and global context information of the data \citep{liu_tam_2021}. The composite function consists of a local branch for generating attention weights and a global branch for generating a channel-wise adaptive kernel. First, the input feature map is squeezed using global average pooling to reduce the computational cost. Subsequently, the local branch executes two 1D convolution operations, with the first convolution using ReLU, and sigmoid activation for the second to generate the local weights. The local weights are then multiplied with the feature map. Meanwhile, the global branch is composed of two fully connected layers, with the first layer using ReLU and the second layer employing softmax function to generate the adaptive kernel (weights). Self-attention is a form of temporal attention, initially proposed for machine translation to enable the deep learning models to attend different words in a sequence relative to other words \citep{vaswani_attention_2023}. The attention module has become a fundamental building block in various natural language processing applications. To generate the attention weights, the input (word embeddings) is transformed by linear projection to compute query, key, and value. Then, the dot product between query and key is computed, and the resultant is normalized by the square root of the size of the key. Finally, the attention weights are obtained by applying the softmax function. Self-attention is the fundamental building block of the Transformer architecture, a key deep learning model in natural language processing. For a more hand-on understanding of self-attention and Transformer, readers are encouraged to explore Transformer Explainer \citep{noauthor_poloclubtransformer-explainer_2025}, an interactive tool that demonstrates how self-attention mechanisms work in Transformer by visualizing the attention scores and how different inputs interact with each other.

Spatial attention focuses on specific regions or spatial location of the input data, enabling the deep learning models to selectively emphasize and ignore certain features. In the context of computer vision, spatial attention is crucial in capturing the spatial relationships and context within an image for accurate prediction. Attention gate is a spatial attention that can identify and focus the salient regions and suppress feature responses of the insignificant ones. The composite function consists of ReLU activation followed by $1 \times 1$ convolution to reduce channel dimension of the feature maps to a singular feature map. Finally, sigmoid is applied to the feature map to generate attention weights \citep{oktay_attention_2018}. The self-attention in the standard Transformer is not effective in handling image data due to its inherent sequential processing nature and lacks the ability to capture spatial dependencies and local patterns. To address this limitation, the Vision Transformer (ViT) treats images as a sequence of non-overlapping patches. A similar computational pipeline is used to generate the attention weights, the sequence of patches is transformed by linear projection to compute the query, key, and value \citep{dosovitskiy_image_2021}. The same operations are employed to generate the attention weights. Self-attention is computationally costly due to its quadratic complexity, especially when dealing with image data. To reduce the complexity, two learnable linear layers, independent of the input data, are adopted as the key and value vectors \citep{guo_beyond_2023}. 

\subsection{Activation Functions}
\begin{figure}[h!]
    \centering
    \begin{subfigure}[b]{0.3\textwidth}
        \centering
        \includegraphics[scale=.25]{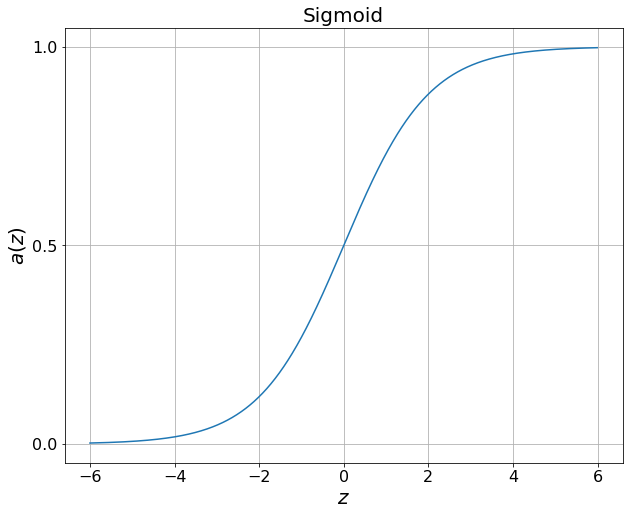}
        \caption{Sigmoid activation function.}
        \label{fig_activation_functions_sigmoid}
    \end{subfigure}
    \hfill
    \begin{subfigure}[b]{0.3\textwidth}
        \centering
        \includegraphics[scale=.25]{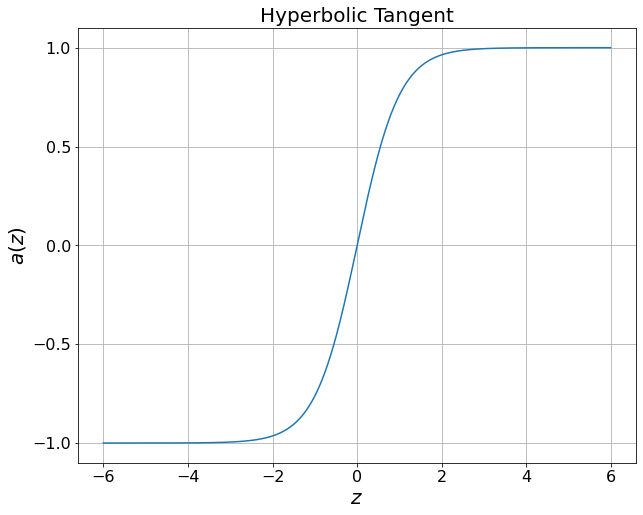}
        \caption{Hyperbolic tangent activation function.}
        \label{fig_activation_functions_tanh}
    \end{subfigure}
    \hfill
    \begin{subfigure}[b]{0.3\textwidth}
        \centering
        \includegraphics[scale=.25]{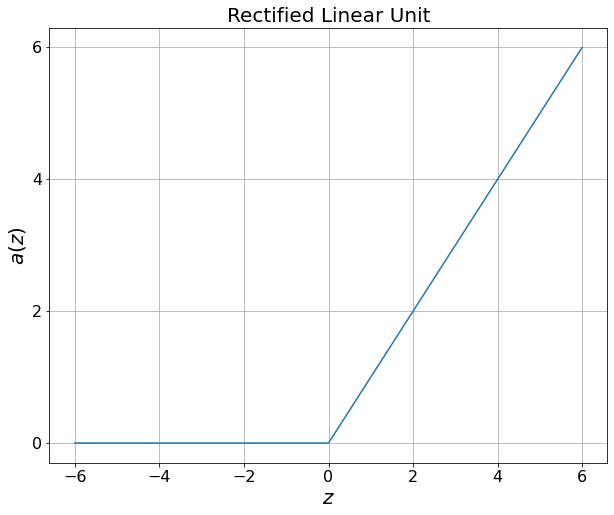}
        \caption{Rectified linear unit activation function.}
        \label{fig_activation_functions_relu}
    \end{subfigure}
\end{figure}

The role of the activation function is to transform the weighted sum into a more classifiable form. This is crucial to the learning behavior of the deep learning model, generating non-linear relationships between the input and the output of the model. The activation function, combined with many hidden layers, allows the neural network to approximate highly complex, non-linear functions. Many activation functions are available for use in neural networks, and some of the functions are shown in Fig. \ref{fig_activation_functions_sigmoid}\textemdash Fig \ref{fig_activation_functions_relu}. The figures show the plots of the three popular activation functions. The sigmoid is a classic example of an activation function used in logistic regression. It maps the weighted sum to a value in the range of 0 to 1 which can be used for classification. The hyperbolic tangent (tanh) is another popular choice of bounded activation function, which produces an output between -1 and 1. Since it has a stronger gradient, neural network training often converges faster than with the sigmoid function \citep{lecun_efficient_2012}. For many years, sigmoid and hyperbolic tangent functions were the commonly used activation functions. However, both suffer from the vanishing gradient problem, which hinders the efficient training of deep neural networks with many layers \citep{hochreiter_gradient_2001}. 

It was shown that neural networks with unbounded activation functions have the universal approximation property and reduce the problem of vanishing gradients. In recent years, numerous unbounded activation functions have been proposed for neural networks, with the softplus \citep{dugas_incorporating_2000} and the rectified linear unit (ReLU) \citep{glorot_deep_2011} activation function being notable examples. These activation functions, especially ReLU has a pivotal role in improving the training and performance of deep learning models. ReLU has been a cornerstone of deep learning models due to its computational efficiency and effectiveness in addressing the vanishing gradient problem. Since then, several variants of ReLU have been proposed including Leaky ReLU \citep{maas_rectifier_2013}, sigmoid linear unit \citep{misra_mish_2020} and exponential linear unit \citep{clevert_fast_2016}, each offering unique advantages for building deep learning-based applications. 

\subsection{Parameter Learning and Loss Functions}
The weights (parameters) of deep learning models are often optimized using an optimization algorithm called gradient descent, though other optimization algorithms may also be used. However, it has to be noted that gradient descent is a generic algorithm which can be used to solve a wide range of optimization problems. In general, gradient descent finds the optimal weights by iteratively updating the weights such that the weights will result in a minimum prediction error over all instances in the training set. This process is synonymous with a person at the top of a hill who wants to climb down to the ground. Just like the person may choose a path that leads to the lowest point by taking small steps based on the steepness of the slope, gradient descent makes small adjustments to the weights, moving them in the direction that reduces the error the most. The prediction error is quantified by a loss function. For classification problems, the commonly used loss function is the negative log-likelihood loss or cross entropy loss, while the square loss and absolute loss are used for regression problems \citep{wang_comprehensive_2022}. The weights of layer $l$ are updated as

\begin{equation}
{w_{i,j}^l = w_{i,j}^l - \alpha \nabla_{w_{i,j}}}
\end{equation}

\noindent where $\alpha$ is a hyperparameter called learning rate and $\nabla_{w_{i,j}}$ is the gradient or the derivative of the loss function $J$ with respect to the weight $\frac{\partial J}{\partial w_{i,j}^l}$.

The gradient can be computed across all training set instances, an approach known as batch gradient descent. However, this approach does not always guarantee convergence to the optimal solution, as it may get stuck in local minima or saddle points, and the same gradient is used for every weight update. An alternative approach is to perform the weight update on the basis of a single instance, but the approach results in a noisy gradient and becomes computationally intensive due to the frequent weight update. A more commonly used approach is to perform the weight update over a set of training instances, known as mini-batch gradient descent. This approach strikes a balance, providing a less noisy gradient and a more stable training process. For a more hands-on understanding of neural network training, readers are encouraged to explore TF playground \citep{carter_tensorflow_nodate} and Initializing Neural Networks \citep{guo_ai_nodate}, interactive tools demonstrating how neural networks are trained and the impact of hyperparameters.

Several efforts have been made to improve the efficiency of gradient descent. One of the earlier efforts is the inclusion of past rate of change in the weight update to speed up the training of deep learning models, the algorithm is called gradient descent with momentum \citep{qian_momentum_1999}. Another effort is to improve the training convergence by adapting the learning rate based on the occurrence of the features \citep{duchi_adaptive_2011}. A more recent work utilizes both adaptive learning rate and momentum to improve the training efficiency and convergence of deep learning models \citep{kingma_adam_2017}. 

\subsection{Regularization Methods}
Regularization methods are employed to prevent overfitting in deep learning models and improve their generalization performance. Early stopping is a method that can detect the onset of overfitting during training by continuously monitoring the validation error. The model is considered overfitting if the validation error starts to increase at some point of the training while the training error is decreasing. However, detecting the onset of overfitting during the training of deep learning models is challenging due to the inherent stochasticity and the presence of noisy data. Several stopping criteria can be considered, such as using a threshold to check if the decrease of (average) validation error is significant and counting the number of successive increases of validation error \citep{prechelt_early_2012}.

Dropout is a regularization method that randomly switches off some neurons in the hidden layers during training with a predefined drop probability (dropout rate) \citep{srivastava_dropout_2014}. Dropout has the effect of training and evaluating a large number of different subnetworks within the models. The dropout rate is a hyperparameter that needs to be carefully tuned to balance regularization and model capacity. Different ranges of dropout rate have been suggested. The original author suggested a dropout rate between 0.5 and 0.8 \citep{srivastava_dropout_2014} while others recommended a lower dropout rate between 0.1 and 0.3 \citep{park_analysis_2017}. Furthermore, it has been suggested a low dropout rate due to the exponential increase in the volume of training data \citep{liu_dropout_2023}.

Parameter norm penalty is a regularization method that adds a penalty term consisting of the network’s weights to the loss function. During the training, the penalty term discourages large weight values and hence, constraining the model’s capacity and reducing the chance of overfitting. The common penalty terms are $L^1$ norm penalty \citep{tibshirani_regression_1996}, $L^2$ norm penalty, also known as weight decay and a combination of $L^1$ and $L^2$ \citep{zou_regularization_2005}. An adaptive weight decay is proposed, allowing the regularization strength for each weight to be dynamically adjusted \citep{nakamura_adaptive_2019}.

Despite the advantages of the mini-batch gradient descent, each mini-batch may comprise data from different distributions. Furthermore, the data distribution may change after each weight update, which could slow down the training process. Batch normalization overcomes this issue by normalizing the summed input to a neuron over a mini batch of training instances \citep{ioffe_batch_2015}. An alternative method is to perform normalization across the neurons instead of the mini batch, a method known as layer normalization \citep{ba_layer_2016}. Layer normalization is applicable in recurrent neural networks and overcomes the dependencies on the mini batch size.

\section{Types of Deep Learning}\label{sec3}
Deep learning models can be categorized into deep supervised learning and deep unsupervised learning.

\subsection{Deep Supervised Learning}
Deep supervised models are trained with a labelled dataset. The learning process of these models involves calculating the prediction error through a loss function and utilizing the error to adjust the weights iteratively until the prediction error is minimized. Among the deep supervised models, three important models are identified, namely multilayer perceptron, convolutional neural network and recurrent neural network. 

\subsubsection{Multilayer Perceptron} 
\begin{figure}[h!]
    \centering
    \includegraphics[scale=0.6]{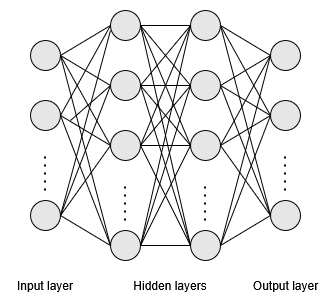}
    \caption{A fully connected neural network.}
    \label{fig_deep_sv_learning_mlp}
\end{figure}

Multilayer perceptron is a neural network model with one or more hidden fully connected layers stacked between the input and output layers as shown in Fig. \ref{fig_deep_sv_learning_mlp}. The width (number of neurons) of the hidden layers and the depth (number of layers) of the network influence the model’s ability to learn patterns in the data. Specifically, the width affects the network’s ability to capture a broader range of features, while the depth facilitates the learning of hierarchical representations. Nevertheless, studies indicated that a multilayer perceptron with a single hidden layer can approximate any continuous function \citep{cybenko_approximation_1989}, \citep{hornik_multilayer_1989}. Multilayer perceptron is effective in various industries and applications from healthcare to finance \citep{widrow_neural_1994}. However, a multilayer perceptron requires the input data to be structured in a one-dimensional format (e.g., tabular data), making it less suitable for unstructured data such as images, text, and speech. To leverage multilayer perceptron for unstructured data, a feature extraction or transformation into structured data is necessary.

\subsubsection{Recurrent Neural Network} 
\begin{figure}[h!]
    \centering
    \includegraphics[scale=0.6]{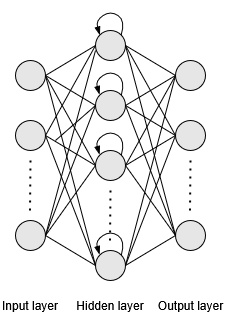}
    \caption{A neural network with recurrent connection.}
    \label{fig_deep_sv_learning_rnn}
\end{figure}

Recurrent neural network (RNN) is a neural network model that leverages the sequential information and memory through the use of recurrent connections, allowing it to effectively process data such as time series, text, speech and other sequential patterns. As shown in Fig. \ref{fig_deep_sv_learning_rnn}, a recurrent neural network is characterized by the recurrent connection, which enables the network to loop back and use internal state from the previous time step to the next time step. The internal state is parameterized by a set of weights shared across the sequence of data. The training of recurrent neural networks suffers from the issue of vanishing gradient due to the challenges of propagation of gradients over a long sequence of data. Variants of recurrent neural networks are introduced to overcome the problem of vanishing gradient, such as long short-term memory (LSTM) \citep{hochreiter_long_1997} and gated recurrent memory (GRU) \citep{cho_learning_2014}. The improved recurrent neural networks introduce memory cell and gating mechanisms to retain and discard information in every time step, allowing for more effective learning dependencies in long sequence. The network architecture can be built using fully connected and convolutional layers \citep{shi_convolutional_2015}.

\begin{figure}[h!]
    \centering
    \includegraphics[scale=0.45]{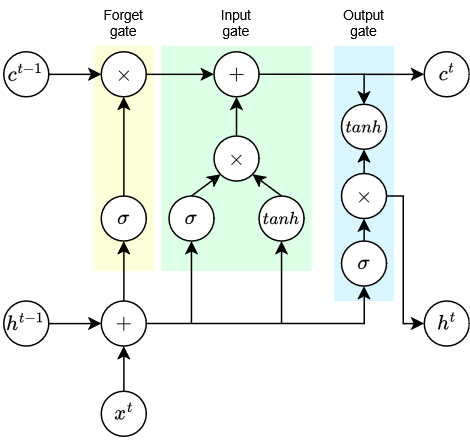}
    \caption{The architecture of an LSTM unit.}
    \label{fig_deep_sv_learning_lstm}
\end{figure}

LSTM is arguably the most widely used variant and has been applied in various applications such as image captioning, machine translation and sentiment analysis \citep{van_houdt_review_2020}. Similar to the standard RNNs, LSTMs process information steps by steps using a chain of repeating units. Each LSTM unit consists of several gates, namely the forget gate, input gate and output gate that control the flow of information from one time step to the next time step. Figure \ref{fig_deep_sv_learning_lstm} illustrates how information flows through these gates. The key component of an LSTM unit is the cell, represented by the horizontal line running through the top of the block. This cell maintains the state at each time step and is updated by the gates. Specifically, at each time step, the forget gate determines which information to be retained or discarded from the cell state based on the current input and the previous hidden state. The input gate decides what new information should be added to the cell state, while the output gate uses the updated cell state, current input and previous hidden state to produce the hidden state for the current time step.

\begin{figure}[h!]
    \centering
    \includegraphics[scale=0.45]{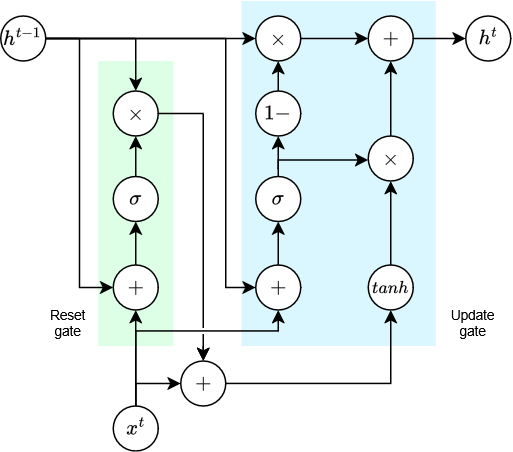}
    \caption{The architecture of an LSTM unit.}
    \label{fig_deep_sv_learning_gru}
\end{figure}

GRU is a popular variant of RNN that is similar to LSTM but has a relatively simpler architecture. The network has two gates called reset and update to control the information flow. Unlike LSTM, GRU does not maintain a separate internal cell state, but uses the reset gate to determine which parts of the information to be retained and discarded, and the update gate to control how much of the previous hidden state should be passed to the current hidden state. Figure \ref{fig_deep_sv_learning_gru} illustrates the interconnection of the gates in a GRU unit.

\subsubsection{Convolutional Neural Network} 
\begin{figure*}[h]
    \centering
    \includegraphics[scale=0.6]{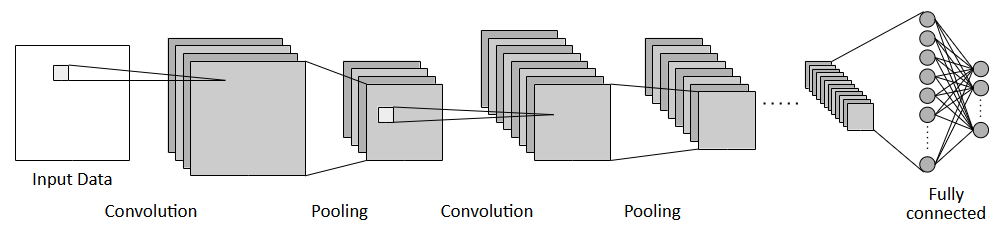}
    \caption{A neural network with convolutional and pooling layers followed by fully connected layers.}
    \label{fig_deep_sv_learning_cnn}
\end{figure*}

Convolutional neural network (CNN) is a neural network model that preserves and leverages the spatial local information in the data through the use of convolutional layers. Fig. \ref{fig_deep_sv_learning_cnn} shows a typical architecture of a convolutional neural network which consists of convolutional, pooling and fully connected layers. The convolutional and pooling layers are stacked alternately to automatically extract salient features in a hierarchical manner. The extracted features are then fed to fully connected layers to predict the outputs. The final feature maps need to be converted to a one-dimensional vector before they are fed to the fully connected layers. The conversion can be performed by flattening the feature maps. CNN architecture is crucial in increasing the performance of the prediction, as it is designed to efficiently extract the feature representation of the input data, enabling more accurate and robust pattern recognition. Over the last decade, several CNN architectures have been proposed, whereby the focus of the improvements has been on enhancing the feature learning capabilities and addressing challenges such as vanishing gradient and diminishing feature reuse.

AlexNet is among the first CNN models that gained widespread recognition and success, marking a significant achievement in the field of deep learning for computer vision tasks \citep{krizhevsky_imagenet_2012}. The model consists of five convolutional layers with maximum pooling operation performed after the first and second convolutional layers, followed by three fully connected layers. The first and second convolutional layers utilize a filter size of $11 \times 11$ and $5 \times 5$ respectively, and $3 \times 3$ filter size is used for the remaining convolutional layers. ReLU activation function is used to mitigate the vanishing gradient. AlexNet is the first deep learning architecture that demonstrated CNN's capability for large-scale image recognition. While 8 layers was considered deep for its time, later deep learning architectures demonstrated that even deeper networks could achieve better performance.

ZFNet is a classic CNN model which has a similar architectural principle as AlexNet, featuring five convolutional layers with maximum pooling layers after the first and second convolution, followed by three fully connected layers \citep{zeiler_visualizing_2013}. The significant differences are the use of smaller filter size and stride in the convolutional layers and contrast normalization of the feature maps, which allows the model to capture better features and improve the overall performance. This configuration improves feature extraction and performance compared to AlexNet. However, like AlexNet, its limited depth restricted its ability to learn more complex hierarchical features.

Network-in-network introduces two innovative concepts to enhance the performance of the model \citep{lin_network_2014}. The first was introducing a block of convolutional layers consisting of $k \times k$ convolution followed by two $1 \times 1$ convolution operations. The pointwise convolutions are similar to applying a multilayer perceptron on the feature maps, allowing the model to approximate more abstract feature representations. In the preceding models, the final feature maps are vectorized by flattening operation for classification by the fully connected layers. Instead of flattening, network-in-network model calculates the spatial average of each feature map, and the resulting vector is fed to softmax function for classification. This approach is parameter-less, significantly reducing the number of parameters. Although this approach makes it more computationally efficient, its general performance does not always surpass models with deeper architecture.

VGGNet attempts to improve the CNN architecture by adding more convolutional layers, specifically up to 19 layers to capture more intricate feature representation from input data, followed by three fully connected layers \citep{simonyan_very_2015}. ReLU activation function is used to reduce vanishing gradient. Unlike AlexNet, all convolutional layers utilize a small fix filter size of $3 \times 3$ and maximum pooling layer is added after a stack of two or three convolutional layers. This configuration allows the model to extract more discriminative features and decreases the number of parameters. The architecture provides better generalization and finer feature extraction, but it comes at the cost of a large number of parameters (138 million), making it more computationally expensive for practical applications.

\begin{figure*}[h]
    \centering
    \includegraphics[scale=0.5]{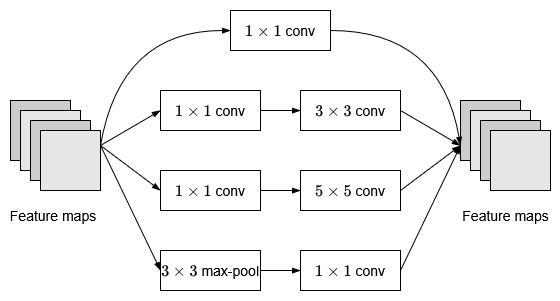}
    \caption{The inception module. Adapted from \citep{szegedy_going_2014}.}
    \label{fig_deep_sv_learning_cnn_inception}
\end{figure*}

GoogleNet or Inception v1 leverages the fact that visual data can be represented at different scales by incorporating a module which consists of multiple convolutional pipelines with different filter sizes \citep{szegedy_going_2014}. The module known as inception utilizes three kernel sizes ($5 \times 5$, $3 \times 3$, $1 \times 1$) to capture spatial and channel information at different scales of resolution as shown in Fig. \ref{fig_deep_sv_learning_cnn_inception}. This configuration enables a more effective feature extraction at both fine-grained and coarse-grained information from input data while maintaining computational efficiency. The model architecture utilizes the inception module at the higher layers, while the traditional convolution and maximum pooling block is used to extract primitive and basic features. The inception modules are stacked upon each other, with maximum pooling operation is performed occasionally to reduce the spatial resolution of the feature maps. GoogleNet utilizes global average pooling to vectorize the final feature maps before passing it to a fully connected layer for classification. However, the complexity of the model architecture may hinder its interpretability and tuning compared to simple models. GoogleNet has been further enhanced in later Inception models by introducing batch normalization, auxiliary classifier and deeper architecture.

Increasing the number of layers enhances the model performance, mainly for solving complex tasks. However, training a very deep neural network is challenging due to the vanishing gradient problem, where the gradients that are used to update the network become insignificant or extremely small as they are backpropagated from the output layer to the earlier layers. A model called Highway Network overcomes this issue by introducing a gating mechanism that regulates the information flow of the layers, enabling the flow of information from the earlier layers to the later layers \citep{srivastava_highway_2015}. Consequently, this not only mitigates the vanishing gradient problem, but also renders the gradient-based training more tractable, enabling the training of very deep neural networks consisting as many as 100 layers. However, the gating mechanism increases the model complexity, making the model mode resource-intensive and less suitable for real-time applications.

\begin{figure}[h!]
    \centering
    \includegraphics[scale=0.5]{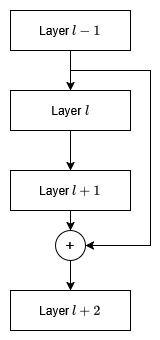}
    \caption{A residual connection which explicitly incorporates features from previous layers into the later layers. Adapted from \citep{he_deep_2015}.}
    \label{fig_deep_sv_learning_cnn_resnet}
\end{figure}

The gating mechanism of Highway Network increases the number of parameters for regulating the information flow. ResNet is a CNN architecture that incorporates residual (skip) connection that allows information to bypass certain layers, mitigating the vanishing gradient problem \citep{he_deep_2015}. ResNet architecture stacks residual blocks, which consists of two or three of convolutional layers with batch normalization and ReLU, and a skip connection which adds the input to the output of the final convolutional layer as shown in Fig. \ref{fig_deep_sv_learning_cnn_resnet}. If the input dimension does not match with the residual output dimension, a linear projection is performed by the residual connection to match the dimensions. This concept of "feature reuse" is central to ResNet's design, as the skip connection allows the features learned in the previous layers to be directly reused in the layer layers, enhancing the model's ability to learn hierarchical features. In comparison to the gating mechanism of Highway Network, the residual connection is parameter-free, and thus does not incur additional computational costs. Furthermore, the connections are never closed whereby all information is always passed through the layers. This innovative concept enables the training of very deep neural networks boasting as many as 152 layers. However, as the model gets deeper, the effectiveness of skip connection diminishes and the performance gain is negligible. Furthermore, ResNet can be more complex to implement and fine-tune compared to simpler architecture.

\begin{figure*}[h]
    \centering
    \includegraphics[scale=0.45]{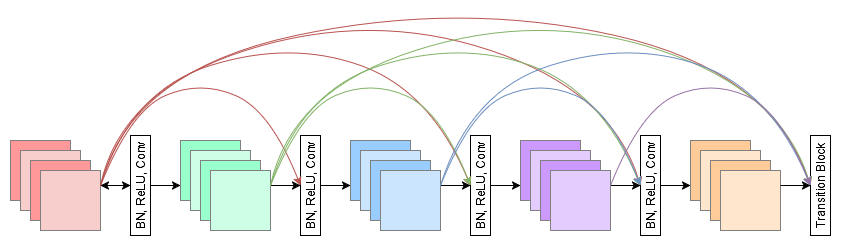}
    \caption{A 5-layer dense block. Adapted from \citep{huang_densely_2017}}
    \label{fig_deep_sv_learning_cnn_densenet}
\end{figure*}

DenseNet is another CNN architecture that overcomes the vanishing gradient problem. DenseNet follows the same approach as ResNet and Highway Network, utilizing skip connection to allow information flow from the earlier layers to later layers. However, DenseNet takes this concept one step further, by introducing a dense block consisting of multiple convolution functions (layers) with each convolution function performs batch normalization followed by ReLU and $3 \times 3$ convolution. Each convolutional layer in the dense block receives feature maps from all its preceding layers. Hence, the connection is referred to as a dense connection \citep{huang_densely_2017}. This configuration as shown in Fig. \ref{fig_deep_sv_learning_cnn_densenet} maximizes information flow, feature reuse and preserves the feed-forward nature of the network, improving the feature learning. To reduce the computational costs, a block of $1 \times 1$ convolutional with batch normalization and maximum pooling layers known as transition block is used to reduce the spatial dimension of the feature maps. The model architecture integrates these dense and transition blocks, stacking them alternately. The network depth can reach up to 264 layers. However, Dense blocks can become computationally expensive due to the increasing number of feature maps. Furthermore, it is prone to overfitting due to the dense connection between the layers.

Although skip connections in ResNet effectively mitigate the vanishing gradient problem, a new challenge arises in the form of diminishing feature reuse as the network becomes deeper. Diminishing feature reuse refers to the diminishing effectiveness of the previously learned feature maps in subsequent layers, impacting the final prediction. WideResNet is a CNN architecture that is based on ResNet with the aim to mitigate diminishing feature reuse problem. Instead of making the network deeper, WideResNet makes the network wider by increasing the number of channels by $k$ factor \citep{zagoruyko_wide_2017}. The increased width allows the model to capture more diverse features, enhancing its ability to learn complex relationships in the input data. This configuration improves feature learning efficiency while mitigating depth-related issues such as vanishing gradients and overfitting. However, wider models may increase memory usage and computational costs.

\begin{figure*}[h]
    \centering
    \includegraphics[scale=0.45]{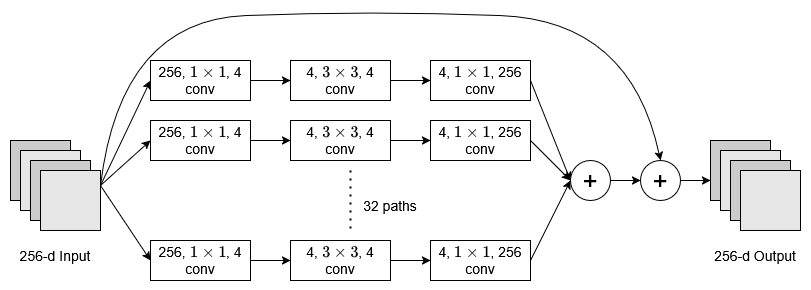}
    \caption{A cardinal block. Adapted from \citep{xie_aggregated_2017}}
    \label{fig_deep_sv_learning_cnn_resnext}
\end{figure*}

ResNeXt addresses the diminishing feature reuse by capturing more efficient and diverse features of the input data. ResNeXt introduces a concept of cardinality which is loosely based on the inception module as shown in Fig. \ref{fig_deep_sv_learning_cnn_resnext}. Cardinality refers to the number of independent and identical paths, where each path performs transformation of the input data, divided along the channel dimension  \citep{xie_aggregated_2017}. In other words, instead of solely relying on increasing the depth of the model, ResNeXt enhances the feature learning by parallelizing the feature extraction through this cardinal path. In the proposed architecture, each path configuration is similar to the residual block of ResNet. The output from each path is then aggregated to form a comprehensive and diverse representation of the input data. The skip connection is used to mitigate the vanishing gradient problem. ResNeXt improves feature learning efficiency by balancing the depth and the width of the model, but its increased architectural complexity introduces additional parameters such as the number of cardinality, and increases memory usage and computational costs.

These CNN models were trained using the ImageNet \citep{noauthor_imagenet_nodate}  and/or CIFAR \citep{noauthor_cifar-10_nodate} datasets. ImageNet is considered as the most influential and important dataset in deep learning for computer vision research. The dataset was used to train all popular CNN models such as VGGNet, ResNet and DenseNet due to its large number of labeled images. The dataset contains 1.2 million training images, 50,000 validation images and 100,000 test images across 1000 object classes. CIFAR is a comparatively smaller dataset and includes two variants: CIFAR-10 and CIFAR-100. Both contain 50,000 training images and 10,000 test images, with CIFAR-10 covering 10 object classes and CIFAR-100 covering 100 classes.

\begin{figure*}[h!]
    \centering
    \includegraphics[scale=0.6]{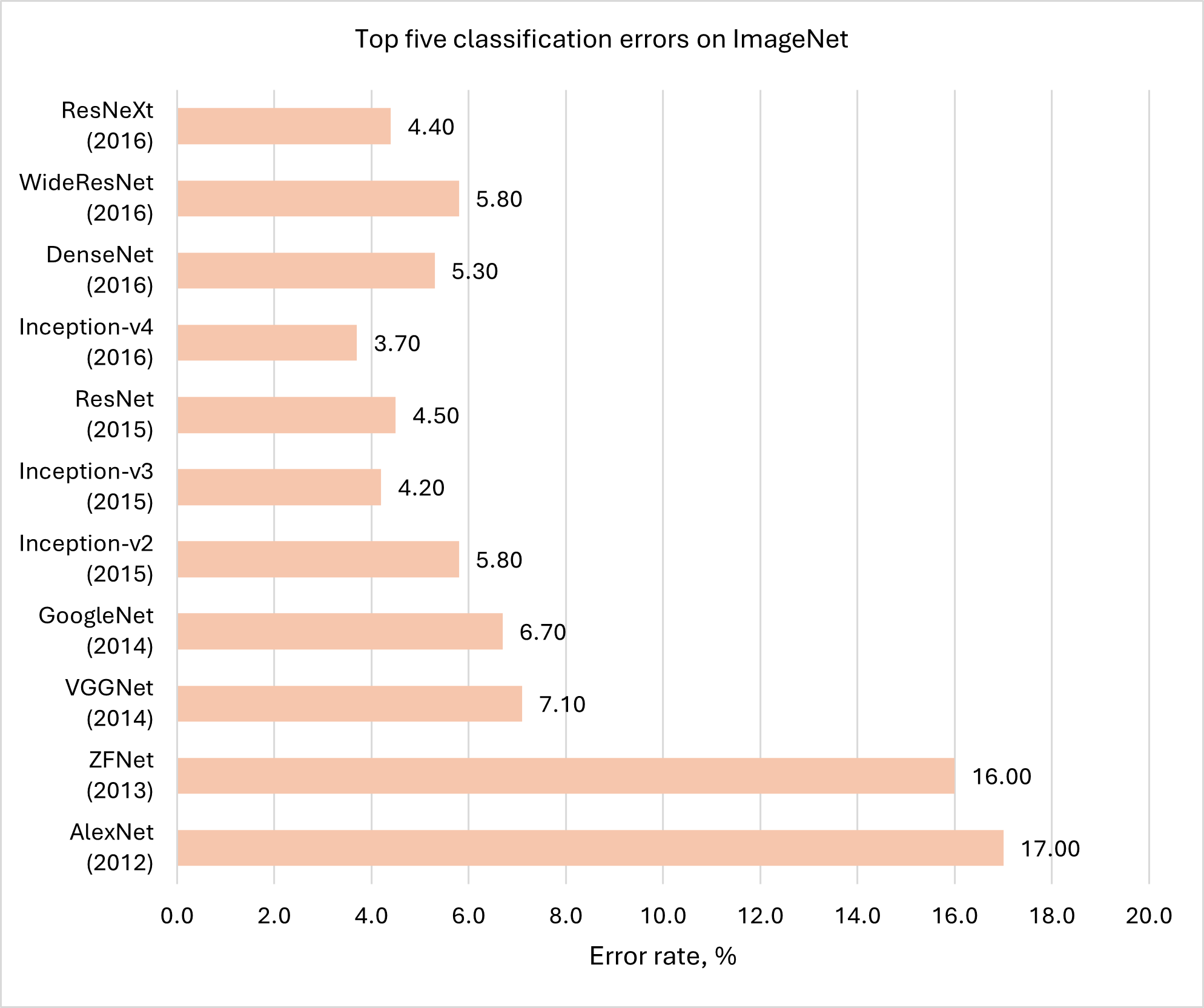}
    \caption{Top five classification errors on ImageNet.}
    \label{fig_cnn_top_five_errors_imagenet}
\end{figure*}

\begin{figure*}[h!]
    \centering
    \includegraphics[scale=0.6]{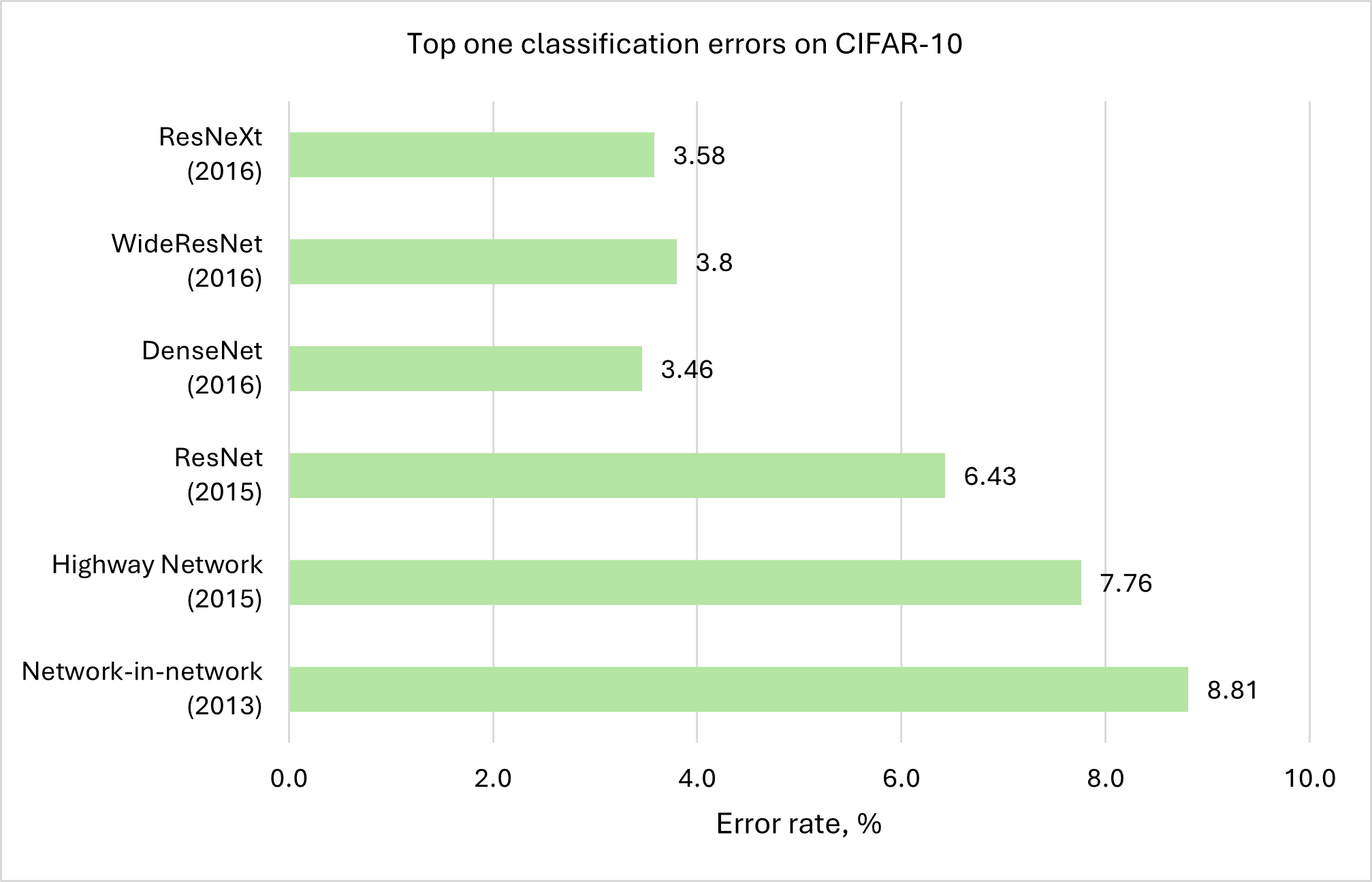}
    \caption{Top one classification errors on CIFAR-10.}
    \label{fig_cnn_top_one_errors_cifar10}
\end{figure*}

Figures \ref{fig_cnn_top_five_errors_imagenet} and \ref{fig_cnn_top_one_errors_cifar10} show the top five classification error rates of the CNN models on ImageNet and CIFAR-10, respectively. Error rates for CIFAR-100 are not reported, as it was used only by Network-in-network, DenseNet, Wide ResNet and ResNeXt. It is also noted that the error rates on ImageNet are sourced from the models' published manuscripts, which may differ from the values reported on the ILSVRC website. As shown in the figures, error rates are generally similar between the datasets, despite their differences in complexity. This could be attributed to ImageNet's large number of training images, allowing the models to generalize well across its 1000 classes. Furthermore, it is evident that the newer models, such as ResNeXt, DenseNet, Inception-v4 and ResNet significantly outperform older models due to their more advanced architectures. Among the evaluated models, DenseNet achieved the best performance on CIFAR-10 with a 3.46\% error rate, while Inception-v4 achieved the lowest error rate on ImageNet at 3.70\%.

\subsubsection{Applications of Deep Supervised Learning Models}
The deep supervised learning models have become integral to advancements in various applications such as medical imaging, internet of things and robotics due to their ability to extract complex hierarchical features. Typically, the studies utilize transfer learning to build the predictive models for solving the problem at hand, especially in scenarios with limited labelled data. This approach leverages pre-trained models to accelerate model development and improve accuracy. In \citep{chougrad_deep_2018}, VGGNet-16, Inception-v3 and ResNet-50 are employed as pre-trained models for breast cancer screening. Specifically, the weights of the convolutional layers are frozen, and the fully connected predicting layers are replaced with new ones. The models are then fine-tuned to adapt to the specific characteristics of the breast cancer dataset, improving their performance on the target task. In medical image segmentation, ResNet \citep{zhao_novel_2020} and DenseNet \citep{cinar_hybrid_2022} have been utilized to progressively capture rich spatial features from the input images, systematically reducing their dimensions while preserving essential information. These compressed feature representations are then gradually expanded and refined through upsampling operations, followed by localizing and predicting the segmentation regions.

In other studies, the architecture of the CNN models have been enhanced to improve their feature extraction capability. In \citep{sun_self-attentional_2023}, ResNet architecture is modified by replacing ReLU with LeakyReLU and incorporating self-attention before the predicting layers to assign weights based on the feature importance. In another study, the residual block of ResNet-18 is enhanced by integrating a channel attention before the skip connection addition \citep{dong_improved_2023}. Additionally, a channel attention featuring two parallel processing pipelines is introduced before the predicting layers. One pipeline applies maximum pooling while the other uses average pooling, allowing the model to capture diverse features. Each pooled output is then processed by convolutional layers, and the resulting feature maps are concatenated to combine the extracted features for prediction. In \citep{hou_application_2024}, DenseNet is modified to reduce feature reuse by controlling the number of feature maps inputs to each layer using a parameter. This parameter controls the number of feature maps based on the distance between layers, whereby if the distance between two layers is large, the input feature maps are reduced to half of the original number. Additionally, the network width is modified based on depth, leading to a gradual widening of the network as it deepens. The modification reduces the complexity of DenseNet, improving both memory usage and computational efficiency.

The LSTMs and GRUs have been exploited to capture temporal information in time series data. In \citep{narotamo_deep_2024}, AlexNet, VGG16 and ResNet50 are used to extract spatial features from ECG images, while LSTM and GRU are used to extract temporal features from the ECG signals. The features are then combined and passed to a self-attention module, followed by fully connected layers for predictions. In \citep{zhang_heart_2024}, 1D-CNN is hybridized with LSTM to exploit the spatial feature extraction capabilities of convolutional layers and the temporal sequence learning capabilities of LSTM. The time series data is first processed by a series of convolutional and maximum pooling layers to extract spatial features. These features are then fed into 10 layers of LSTM capture temporal dependencies and sequential patterns within the data, improving the performance of the predictions. A hybrid deep learning model with parallel feature learning pipelines is introduced for classifying motion signals into human activities \citep{mohd_noor_deep_2022}. The feature learning pipelines consist of convolutional and maximum pooling layers to extract local features from the sequence of segmented signals (windows). The local features are concatenated to form a sequence, which is then fed into an LSTM to capture temporal dependencies and patterns. A similar hybrid model is proposed, combining ResNet-18 with an LSTM for EEG signal classification. The feature maps produced by ResNet-18 are flattened and fed into the LSTM, followed by a fully connected layer for prediction.

\subsection{Deep Unsupervised Learning}
Deep unsupervised models are trained with an unlabeled dataset. The learning process of these models discovering patterns, structures, and representations within the data without relying on explicit labels or supervision. Instead, these models often learn by optimizing objective functions that capture the underlying data characteristics such as clustering, learning useful feature embeddings and reconstructing input data from compressed representations. Examples of deep unsupervised models are autoencoders, generative adversarial networks and restricted Boltzmann machines.

\begin{figure}[h!]
    \centering
    \includegraphics[scale=0.6]{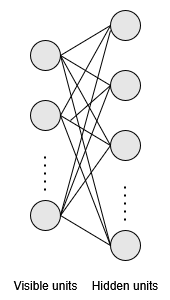}
    \caption{A restricted Boltzmann machine.}
    \label{fig_deep_unsv_learning_rbm}
\end{figure}

Restricted Boltzmann Machine is a generative neural network model that learns a probability distribution based on a set of inputs. The model consists of a visible (input) layer and a hidden layer with symmetrically weighted connections as shown in Fig. \ref{fig_deep_unsv_learning_rbm}. The input layer represents the input data with each node corresponding to a feature or variable while the hidden layer learns the abstract representation of the input data. Restricted Boltzmann machine model is trained using contrastive divergence, an algorithm that is based on a modified form of gradient descent, utilizing a sampling-based approach to estimate the gradient \citep{hinton_practical_2012}. It has found success in solving combinative problems such as dimensionality reduction, collaborative filtering and topic modelling.

\begin{figure}[h!]
    \centering
    \includegraphics[scale=0.6]{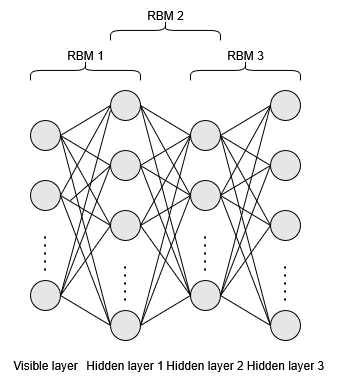}
    \caption{A deep belief network.}
    \label{fig_deep_unsv_learning_dbn}
\end{figure}

Deep Belief Network can be viewed as a stack of restricted Boltzmann machines, comprising a visible layer and multiple hidden layers \citep{hinton_fast_2006} as shown in Fig. \ref{fig_deep_unsv_learning_dbn}. Deep belief network has two training phases. The initial phase is known as pretraining in which the network is trained layer by layer, with each layer serves as a pretraining layer of the subsequent layers. This sequential learning allows the hidden layers learn complex hierarchical feature representation of the data. The second phase is called fine-tuning whereby the deep belief network model can be further trained with supervision to perform tasks such as classification and regression \citep{hinton_deep_2009}.

\begin{figure}[h!]
    \centering
    \includegraphics[scale=0.6]{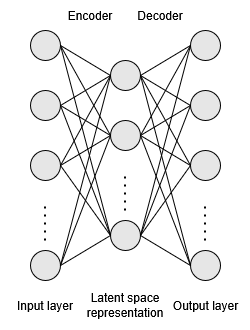}
    \caption{An autoencoder.}
    \label{fig_deep_unsv_learning_ae}
\end{figure}

An autoencoder is a generative neural network model that learns to encode the input data into a compressed representation and then reconstructs the original data from this representation. The layers that encode the input data are known as encoder while the layers that responsible for the reconstruction are referred to as the decoder as shown in Fig. \ref{fig_deep_unsv_learning_ae}. The encoded data (hidden layer) represents the abstract features of the input data, also known as latent space or encoding. The decoder can be removed from the autoencoder, creating a standalone model that can be used for data compression and dimensionality reduction \citep{romero_quantum_2017}, \citep{li_guided_2020}. The decoder can also be replaced with predictive layers for classification task \citep{mohd_noor_feature_2021}. The network architecture can be built using fully connected and convolutional layers \citep{li_comprehensive_2023}. 

Several autoencoder variants have been introduced to improve the autoencoder’s ability to capture better feature representation. Some introduced penalty terms to the loss function such as sparsity penalty (sparse autoencoder) \citep{ng_sparse_2011} to encourage sparse representation and Jacobian Frobenius norm (contractive autoencoder) \citep{rifai_higher_2011} to be less sensitive to small and insignificant variations in the input data while encoding the feature representation. Others trained the autoencoder to recover original data from corrupted data with noise \citep{vincent_extracting_2008}. An improved denoising autoencoder knowns marginalized denoising autoencoder has been proposed which marginalizes the noise by adding a term that is linked to the encoding layer \citep{chen_marginalized_2012}. Variational autoencoder is a variant of autoencoder that has similar architecture: encoder, latent space and decoder. Despite the similarity, instead of learning a fixed encoding, variational autoencoder learns the probability distribution of the input data in the latent space \citep{kingma_auto-encoding_2013}. The model can be used to generate data by sampling from the learned probability distribution. The network architecture can be built by stacking more than one fully connected layer and convolutional layer.

\begin{figure}[h!]
    \centering
    \includegraphics[scale=0.4]{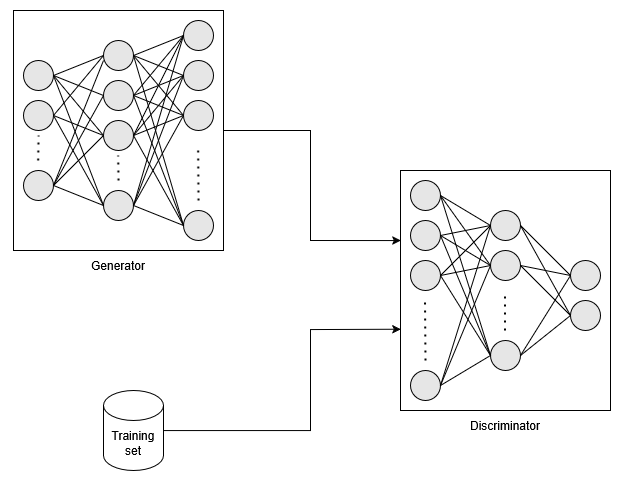}
    \caption{A generative adversarial network.}
    \label{fig_deep_unsv_learning_gan}
\end{figure}

Generative Adversarial Network (GAN) is another generative neural network model that is designed for generating data that adheres closely to the distribution of the original training set. The model consists of two different neural networks namely generator and discriminator as shown in Fig. \ref{fig_deep_unsv_learning_gan}. The generator learns to imitate the distribution of the training set given a noise vector, effectively outsmarting the discriminator. Simultaneously, during the training, the discriminator is trained to differentiate between the real data from the training set and synthetic data generated by the generator \citep{goodfellow_generative_2014}. This intricate dynamic between the networks drives an iterative learning process whereby the generator continually refines its ability to create synthetic data that closely resembles the real data, while the discriminator enhances its ability to distinguish between authentic and fake data.

The model can be extended by providing the labels to both generator and discriminator in which the model known as conditional GAN, capable of generating 1000 image classes \citep{odena_conditional_2017}. Conditional GANs require a labelled dataset, which might limit its application. InfoGAN is similar to conditional GAN, but the labels are substituted with latent codes, which allows the model to be trained in an unsupervised manner \citep{chen_infogan_2016}. GANs often suffer from mode collapse, where the model can only generate a single or small set of outputs. Wasserstein GAN improves the training by utilizing Wasserstein loss function, which measures the difference between the real and synthesized data distribution \citep{weng_gan_2019}. ProGAN tackles the training instability of GAN by progressively growing the generator and discriminator. The idea is that the model is scaled up gradually, starting with the simplest form of the problem, and little by little the problem’s complexity is increased as the training progresses \citep{karras_progressive_2018}. StyleGAN leverages the progressive GAN’s approach and neural style transfer to improve the quality of the generated data \citep{karras_style-based_2019}. The model is characterized by the independent manipulation of both style and content, allowing it to generate diverse styles and high-quality data.

\section{State-of-the-art Deep Learning Applications}\label{sec4}
As discussed in the previous section, the application of deep learning ranges from computer vision \citep{Tan_2020_CVPR}, natural language processing \citep{Otter}, healthcare \citep{esteva2019guide}, robotics \citep{soori2023artificial}, education \citep{hernandez2019systematic}, and many others. This section presents the applications of deep learning across several areas.

\subsection{Computer Vision}
Computer vision is an essential field in artificial intelligence (AI). It is a field of study that focuses on enabling computers to acquire, analyze and interpret visual inputs to derive meaningful information. The visual inputs can take many forms such as digital images, sequence of images or video and point cloud, and the source of these inputs can be camera, LiDAR and medical scanning machine. Deep learning, specifically CNN models have been widely used in real-world computer vision applications including image classification, object detection and image segmentation. This section discusses more details about the recent advancements in deep learning models that have been achieved over the past few years.

\subsubsection{Image Classification}
Image classification is a fundamental task in computer vision, which involves categorizing an image into one of predefined classes based on the visual content. The objective of image classification is to enable computers or machines to differentiate between objects within images, in a manner similar to how humans interpret visual information. Image classification is a crucial component in various applications such as robotics, manufacturing, and healthcare. LeNet-5, introduced in 1998, is one of the earliest convolutional neural networks that was successfully trained to classify handwritten digits. The model underwent a series of improvements, including the use of tanh and average pooling, which enhanced its ability to extract hierarchical features, ultimately improving overall performance. The model architecture comprises two convolutional layers, each with an average pooling layer, followed by two fully connected layers, including the output layer \citep{lecun_gradient-based_1998}. Since then, numerous CNN models have been proposed based on LeNet-5 for image classification \citep{simard_best_2003, matsugu_subject_2003} but the most significant one is AlexNet in 2012, which saw a transformative breakthrough in deep learning. AlexNet is considered the first CNN model with a large number of parameters that significantly improved the performance of image classification on a very large dataset (ImageNet). The model won first place in ILSVRC 2012, improving the test error from the previous year by almost 10\% \citep{krizhevsky_imagenet_2012}. Numerous significant CNN models have been introduced in subsequent ILSVRC competitions including ZFNet, VGG16, GoogleNet, ResNet and ResNext. In general, the research focused on increasing the number of layers, addressing the problem of vanishing gradient and diminishing of feature reuse.

Research in image classification continues to evolve with a focus on addressing key challenges to improve the classification performance. One notable trend is the formulation of the loss function to address problems such as neglecting well-classified instances and imbalance distribution of class labels. In a particular study, an additive term is introduced to the cross-entropy loss to reward the models for the correctly classified instances. This formulation encourages the models to also pay attention to well-classified instances while focusing on the bad-classified ones \citep{zhao_well-classified_2022}. Another study proposes an asymmetric polynomial loss function using the Taylor series expansion. The loss function allows the training to selectively prioritize contributions of positive instances to mitigate the issue of imbalance between negative and positive classes \citep{huang_asymmetric_2023}. The asymmetric polynomial loss requires a large number of parameters to be fine-tuned and may lead to overfitting. A robust asymmetric loss is formulated by introducing a multiplicative term to control the contribution of the negative gradient and making it less sensitive to parameter optimization \citep{park_robust_2023}. Combining multiple deep learning models improves the overall performance by leveraging the diverse strengths of individual models. However, identifying the optimal combination is non-trivial due to the large number of hyperparameters. A straightforward method is to employ the weighted sum rule \citep{nanni_building_2023}. To enhance the overall performance, an algorithm, named greedy soups, adds a model based on the validation accuracy \citep{wortsman_model_2022}. The final prediction is produced via averaging. Multi-symmetry ensembles framework improves the building of diverse deep learning models by utilizing contrastive learning \citep{loh_multi-symmetry_2023}. Then, the diverse models are sequentially combined based on their validation accuracy.

Vision transformers (ViT) offers an alternative to convolutional neural networks that have long been the dominant architecture for image classification, by leveraging self-attention mechanisms for scalable global representation learning. Despite its effectiveness, ViT is sensitive to hyperparameter optimization and substandard performance on smaller datasets \citep{xiao_early_2021}. Furthermore, ViT lacks the ability to leverage local spatial features which is inherent in convolutional neural networks \citep{wu_cvt_2021}. Therefore, several studies attempt to incorporate convolutional layers into ViT architecture to improve its performance and robustness. In particular, conformer is a network architecture with two branches: CNN branch and a transformer branch to extract local and global features respectively \citep{peng_conformer_2023}. Both branches are connected by two “bridges” of $1 \times 1$ convolution and up or down sampling operations, allowing the branches to share their features and enhance the feature representation. Both branches output predictions which are combined to produce the final prediction. 

\begin{figure}[h!]
    \centering
    \includegraphics[scale=0.65]{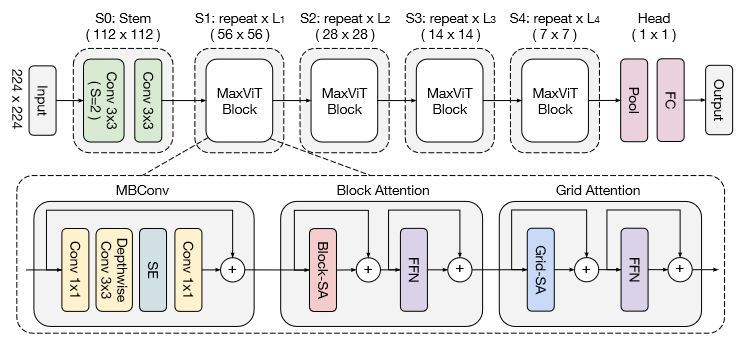}
    \caption{The architecture of MaxViT \citep{tu_maxvit_2022}.}
    \label{fig_img_clf_MaxViT}
\end{figure}

A hybrid architecture, named MaxViT, combines convolutional networks and vision transformer to address the lack of scalability issues of self-attention mechanisms when the model is trained on large input size \citep{tu_maxvit_2022}. The improved vision transformer is composed of two modules whereby the first module attends local features in non-overlapping image patches and the global features are attended by processing a grid of sparse and uniform pixels. The transformer is stacked with a block of convolutional layers to extract local spatial features. The architecture of MaxViT is shown in Figure \ref{fig_img_clf_MaxViT}. Another study proposes a convolutional transformer network, introducing the depthwise convolutional block into the ViT \citep{ma_convolutional_2024}. This configuration allows the model to exploit the ability of convolutional networks to extract local spatial features while the ViT attends the extracted local features to focus on relevant information, enhancing the model’s ability to capture complex patterns and relationships. Specifically, instead of linear mapping, depth-wise convolutional mapping is used to generate query, key, and value matrices. Table \ref{table_summary_imgclf_studies} lists the summary of state-of-the-art image classification.

\begin{table*}[h!]
\centering
\caption{Summary of state-of-the-art image classification.}
\scriptsize
\begin{tabularx}{\textwidth}{p{1.5cm} X X}
 \hline
 Ref. & Description & Results (Datasets and Metrics) \\ 
 \hline
 \citep{zhao_well-classified_2022}, 2022 & The study introduced an encouraged loss that rewards well-classified examples with additive bonuses to enhance their contribution during training, addressing issues in representation learning, energy optimization, and makes the decision margin grows faster & CIFAR-10\newline Accuracy: 92.97\% \newline ImageNet\newline Accuracy: 76.43\% \\ 
 \hline
 \citep{huang_asymmetric_2023}, 2023 & The study introduced asymmetric polynomial loss to address class imbalance by decoupling gradient contributions from positive and negative instances, increasing the impact of updates for minority instances and helping the model focus on the more challenging, less frequent cases & MS-COCO\newline mAP: 90.08\newline NUS-WIDE\newline mAP: 31.27 \\
 \hline
 \citep{park_robust_2023}, 2023 & The study introduced robust asymmetric loss to address class imbalance by emphasizing minority instances through an asymmetric weighting mechanism, allowing the model to focus more on the less frequent, yet critical minority instances & ISIC2018\newline Accuracy: 0.852\newline APTOS2019\newline Accuracy: 0.826 \\
 \hline 
 \citep{peng_conformer_2023}, 2023 & This study introduced dual network structure to extract local and global feature representations using convolutional layers and transformer & ImageNet\newline Accuracy: 81.3 (Conformer-Ti)\newline Accuracy: 83.4 (Conformer-S)\newline Accuracy: 84.1 (Conformer-B) \\
 \hline 
 \citep{tu_maxvit_2022}, 2022 & This study introduced MaxViT, a hybrid attention model, combining convolutional layers with transformer-based architecture whereby it uses convolutional layers in the early stages to capture local patterns efficiently, and then applies a multi-axis attention mechanism to capture both local and global dependencies more effectively & ImageNet\newline Accuracy: 85.72 (MaxViT-T)\newline Accuracy: 86.19 (MaxViT-S)\newline Accuracy: 86.66 (MaxViT-B)\newline Accuracy: 86.70 (MaxViT-L) \\
 \hline 
 \citep{ma_convolutional_2024}, 2024 & This study introduced convolutional transformer network that integrates convolutional layers and transformer encoder blocks whereby a transformer encoder block consists of multi-head self attention and feed-forward network layers with skip connections & Diving 48\newline Accuracy: 77.3 (CTN-a), 78.2 (CTN-b)\newline Epic-Kitchens\newline Accuracy: 44.1 (CTN-a), 45.9 (CTN-b) \\
 \hline 
\end{tabularx}
\label{table_summary_imgclf_studies}
\end{table*}

\subsubsection{Object Detection}
Deep learning plays a major role in significantly advancing the state-of-the-art in object detection performance. Region-based CNN (R-CNN) is the first breakthrough in object detection that combines CNN with selective region proposals \citep{girshick_rich_2014}. The region proposals are the candidate bounding boxes serving as the potential region of interests (objects) within the input image, and the CNN are used to extract features from the region proposals and classify the regions for object detection. An improved model, named Fast R-CNN introduces two prediction branches: object classification and bounding box regression which improves the overall performance of object detection \citep{girshick_fast_2015}. However, R-CNN and Fast R-CNN models are computationally expensive and slow, thus practically infeasible for real-time applications. Addressing this issue, Fast R-CNN is integrated with a region proposal network, referred to as Faster R-CNN \citep{ren_faster_2015}. The region proposal network (RPN) is used to efficiently generate region proposals for object detection. RPN takes an input image and output a set of rectangle object proposals, each with a confidence score to indicate the likelihood of an object’s presence. To this end, RPN introduces the concept of anchor boxes, whereby multiple bounding boxes of different aspect ratios are defined over the feature maps produced by the convolutional networks. These anchor boxes are then regressed over the feature maps to localize the objects, contributing to the improved speed and effectiveness of Faster R-CNN. The training of Faster R-CNN is divided into two stages. First, the RPN is pre-trained to generate the region proposals and then, the Fast R-CNN is trained using the region proposals generated by the RPN for object detection. The backbone network responsible for extracting the features for Faster R-CNN is either ZFNet or VGG16.

In two-stage object detectors, the region proposals are generated first, and then used for object detection. The two-stage process is computationally intensive and infeasible for real-time object detection applications. You Only Look Once or YOLO proposes a one-stage detection by directly predicting bounding boxes and object’s confidence score in a single forward pass through the neural network \citep{redmon_you_2016}. This single pass architecture significantly reduces the computational complexity, making YOLO suitable for real-time object detection applications. In YOLO, the input image is divided into $S×S$ grids, each grid cell is responsible for detecting the objects present in the cell. Specifically, each grid cell predicts multiple bounding boxes and associated object’s confidence score, enabling simultaneous object detection across the entire image. Subsequent enhancements such as YOLOv3 \citep{redmon_yolov3_2018} and YOLOv4 \citep{bochkovskiy_yolov4_2020} are proposed, improving the model’s capability and accuracy. Single Shot Multibox Detector (SSD) is another one-stage detector, which aims to address the issue of real-time object detection \citep{liu_ssd_2016}. SSD also eliminates the region proposal generation and directly predicts bounding boxes and confidence scores, reducing the computational complexity. To improve the overall performance, SSD produces the predictions from different levels of feature maps, allowing detection of objects of different sizes in the input image. 

A common issue in object detection problems is the extremely imbalanced ratio of foreground to background classes. Addressing this issue, RetinaNet introduces a loss function that is based on the cross entropy called focal loss. Focal loss reduces the loss contribution of easily classified objects, allowing the model training to focus on the difficult objects \citep{lin_focal_2020}. RetinaNet adopts the Feature Pyramid Network (FPN) \citep{lin_feature_2017} with ResNet as the backbone network for extracting the feature maps. FPN is a network architecture with a pyramid structure that efficiently captures multiscale feature representation, facilitating object detection across various sizes. To further improve the overall performance, EfficientDet introduces bidirectional FPN, which incorporates multi-level feature fusion to better capture multiscale feature representation \citep{tan_efficientdet_2020}. Also, the model utilizes EfficientNet \citep{tan_efficientnet_2019} as the backbone network to achieve a balance between computational efficiency and accuracy. 

Object detection performance often relies on a post-processing step called non-maximum suppression (NMS) to eliminate duplicate detections and select the most relevant bounding boxes. Specifically, NMS sorts all detection boxes based on their confidence scores, selects a box with the maximum score and discards the other boxes with a significant overlap with the selected box. This process is repeated on the remaining detection boxes. However, due to the inconsistency between the confidence score and the quality of object localization, NMS retains poorly localized bounding boxes with high confidence score while discarding more accurate predictions with poor confidence score. To mitigate this limitation, instead of discarding the neighboring boxes with significant overlap, soft-NMS applies Gaussian function to lower their confidence scores \citep{bodla_soft-nmsimproving_2017}. The idea is not to discard the neighboring bounding boxes, but gradually decline their scores based on the extent of the overlap with the selected box. This results in a smoother suppression, preserving the better-localized bounding boxes. 

Adaptive NMS introduces an adaptive threshold for the suppression of bounding boxes \citep{liu_adaptive_2019}. The algorithm dynamically adjusts the threshold based on the level of overlapping of the selected box with the other bounding boxes. Similar method is reported in \citep{husham_al-badri_adaptive_2023} whereby an adaptive NMS is proposed by dynamically adjusting the suppression criteria based on intersection over union (IoU) values. The IoU values are compared with a threshold and applies an additional iteration to vote for each detected proposal, ensuring better distinction between closely positioned objects. The method helps preserve multiple high-confidence bounding boxes, mitigating the issue of incorrectly merging adjacent objects into one. Redundant bounding boxes are often not filtered out due to their low IoU values with the best bounding box, leading to high false positives. In \citep{jiang_non-maximum_2024}, a learning-based NMS is proposed to reduce false positives by integrating the NMS process into the model's learning framework. Specifically, a novel NMS-aware loss is formulated that incorporates IoU to adjust bounding box weights, boosting negative attention for low IoU bounding boxes and enhancing positive weights for high IoU bounding boxes. Furthermore, regression assisted classification branch is introduced to aid classification by leveraging regression prediction relationships between bounding boxes and their best counterparts. In \citep{shi_similarity_2024}, the authors formulated similarity distance metric is proposed to evaluate the similarity between bounding boxes. The proposed method considers the location and shape of the boxes, and adapts dynamically to different datasets and object sizes. This method improves label assignment by learning hyperparameters automatically, eliminating the need for manual tuning and enhancing detection performance for tiny objects detection.

Detection Transformer (DETR) is an end-to-end trainable object detection model that leverages the transformer architecture to eliminate the need for handcrafted components such as anchor boxes and non-maximum suppression \citep{carion_end_end_2020}. The self-attention mechanism of the transformer captures the global context and relationships between different parts of the image, allowing it to localize the objects and remove duplicate predictions. The model is trained with a set of loss functions that perform bipartite matching between the predicted and ground truth objects. DETR uses ResNet as backbone network. Despite the success of DETR in simplifying and improving object detection tasks, DeTR suffers from a long training time and low performance at detecting small objects due to its reliance on the self-attention mechanism of the transformer, which lacks a multiscale feature representation. To mitigate this limitation, Deformable DETR introduces a multiscale deformable attention module which can effectively capture feature representation at different scales \citep{zhu_deformable_2020}. Furthermore, the attention module leverages deformable convolution, allowing the model to adapt to spatial variation and capture more informative features in the input data. 

\begin{figure*}[h!]
    \centering
    \includegraphics[scale=0.65]{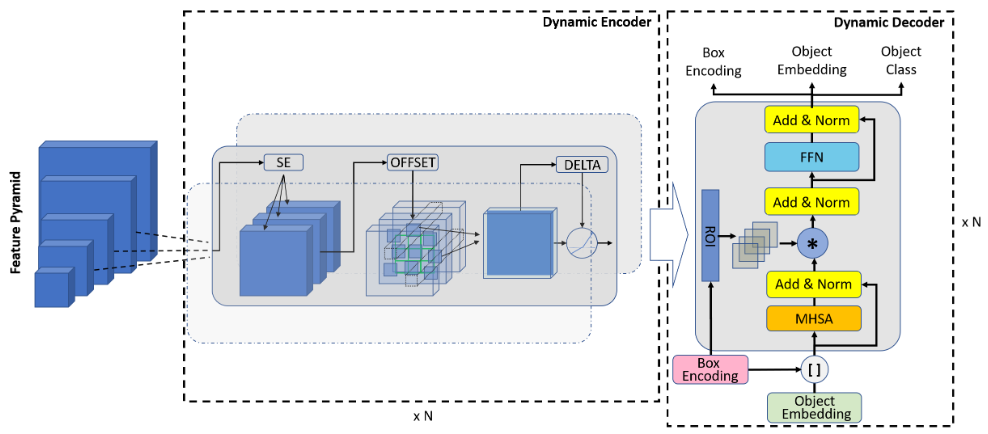}
    \caption{The architecture of dynamic DeTR \citep{dai_dynamic_2021}.}
    \label{fig_obj_det_Dynamic-DETR}
\end{figure*}

Dynamic DETR addresses the same issues by utilizing a deformable convolution-based FPN to learn multiscale feature representation \citep{dai_dynamic_2021}. The model replaces the transformer encoder with a convolution-based encoder to attend to various spatial features and channels. Moreover, an ROI-based dynamic attention is introduced at the transformer decoder, allowing the model to focus on the region of interests. This modification allows the model to effectively detect small objects and converge faster during training. The architecture of dynamic DETR is shown in Figure \ref{fig_obj_det_Dynamic-DETR}. DETR utilizes bipartite matching between the ground truth and the predicted objects to assign each ground truth object to a unique prediction, limiting the localization supervision. In \citep{jia_detrs_2023}, a hybrid matching scheme is introduced, combining the original one-to-one matching with an auxiliary one-to-many matching during training, improving training efficiency and detection accuracy without adding inference complexity. A training scheme known as Teach-DeTR is proposed to improve the overall performance of DeTR \citep{huang_teach-detr_2023}. The training scheme leverages the predicted bounding boxes by other object detection models during the training by calculating the loss of one-to-one matching between the object queries and the predicted boxes. Table \ref{table_summary_objdet_studies} lists the summary of state-of-the-art object detection.

\begin{table*}[h!]
\centering
\caption{Summary of state-of-the-art object detection.}
\scriptsize
\begin{tabularx}{\textwidth}{p{1.5cm} X X}
 \hline
 Ref. & Description & Results (Datasets and Metrics) \\ 
 \hline
 \citep{husham_al-badri_adaptive_2023}, 2023 & This study introduced an adaptive NMS method that dynamically adjusts suppression criteria based on intersection over union (IoU) values, applying an additional iteration to improve the distinction between closely positioned objects and preserve high-confidence bounding boxes, reducing false positives & Rumex weeds\newline IoU: 91.2, mAP: 90 \\ 
 \hline
 \citep{jiang_non-maximum_2024}, 2024 & This study proposed a learning-based NMS that integrates the NMS process into the model's learning framework through a novel NMS-aware loss function, which adjusts bounding box weights using IoU to reduce false positives, while introducing a regression-assisted classification branch to enhance classification by leveraging bounding box prediction relationships & CrowdHuman\newline mAP: 90.1 \\
 \hline
 \citep{shi_similarity_2024}, 2024 & This study proposed a similarity distance metric to evaluate bounding box similarity by considering location and shape, dynamically adapting to different datasets and object sizes, automatically learning hyperparameters to improve label assignment, and enhancing detection performance for tiny objects & AI-TOD\newline mAP: 26.5, AP50: 57.7, AP75: 20.5\newline SODA-D\newline mAP: 32.8, AP50: 59.4, AP75: 31.3\newline VisDrone2019\newline mAP: 28.7, AP50: 50.3 \\
 \hline 
 \citep{dai_dynamic_2021}, 2021 & This study introduced Dynamic DETR, which enhances the original DETR by introducing dynamic attention mechanisms in both the encoder and decoder stages. This approach addresses limitations related to small feature resolution and slow training convergence, resulting in improved performance and efficiency in object detection tasks & COCO2017\newline mAP: 49.3, AP50: 68.4, AP75: 53.6 \\
 \hline 
 \citep{jia_detrs_2023}, 2023 & This study proposed a method that combines a one-to-one matching branch with an auxiliary one-to-many matching branch during training to improve the detection accuracy of DETR & COCO2017\newline mAP: 59.4, AP50: 77.8, AP75: 65.4 \\
 \hline 
 \citep{huang_teach-detr_2023}, 2023 & This study proposed a training scheme to enhances DETR by leveraging predicted bounding boxes from other teacher models including RCNN-based and DETR-based detectors using knowledge distillation. The method does not introduce extra parameters or computational overhead during inference & COCO2017\newline mAP: 58.5, AP50: 77.4, AP75: 64.8 \\
 \hline 
\end{tabularx}
\label{table_summary_objdet_studies}
\end{table*}

\subsubsection{Image Segmentation}
Image segmentation is another important task in which deep learning has a significant impact. One of the earliest deep learning models for image segmentation is the fully convolutional network \citep{long_fully_2015}. A fully convolutional network consists of only convolutional layers which accepts an input of an arbitrary size and produce the predicted segmentation map of the same size. The authors adopted the AlexNet, VGG16 and GoogleNet, replace their fully connected layers with convolutional layers and append a 1×1 convolutional layer, followed by bilinear up-sampling to match the size of the input. The model was considered a significant milestone in image segmentation, demonstrating the feasibility of deep learning for semantic segmentation trained in end-to-end manner. Deconvolution network is another popular deep learning model for semantic segmentation \citep{noh_learning_2015}. The model architecture consists of two parts: encoder and decoder. The encoder takes an input image and uses the convolutional layers to generate the feature maps. The feature maps are fed to the decoder composed of un-sampling and deconvolutional layers to predict the segmentation map. SegNet is another encoder-decoder model for semantic segmentation \citep{badrinarayanan_segnet_2017}. The encoder is a sequence of convolutional (with ReLU) and maximum pooling blocks which is analogous to a convolutional neural network. The decoder is composed of up-sampling layers which up-samples the inputs using the memorized pooled indices generated in the encoder phase, and convolutional layers without non-linearity. The encoder progressively reduces the resolution of the input data while extracting abstract features through a series of convolutional and pooling layers. This process causes the loss of fine-grained information, degrading the overall performance of segmentation. LinkNet mitigates this limitation by passing the feature maps at several stages generated by the encoder to the decoder, hence reducing information loss \citep{chaurasia_linknet_2017}. The model architecture of LinkNet is similar to SegNet, but utilizes ResNet as the encoder.

While Faster R-CNN is a significant approach in object detection task, it has been extended to perform instance segmentation task. One such extension is Mask R-CNN which is based on Faster R-CNN, introduces an additional branch for predicting the segmentation mask \citep{he_mask_2017}. Similar to Faster R-CNN, Mask R-CNN utilizes the RPN to generate region proposals and then the region of interest alignment is applied to extract more accurate features from the proposed regions. Mask R-CNN does not leverage the multiscale feature representation which may degrade the overall performance of segmentation. To overcome this limitation, Path Aggregation Network (PANet) incorporates the FPN and introduces a bottom-up pathway to facilitate the propagation of the low-level information \citep{liu_path_2018}. The pathway takes the feature maps of the previous stage as input and performs 3×3 convolution with stride 2 to reduce the spatial size of the feature maps. The generated feature maps are then fused with the feature maps from the FPN through the lateral connection. The model adopts the three branches as in Mask R-CNN. MaskLab is an instance segmentation model based on Faster R-CNN, consisting of object detection, segmentation, and instance (object) center direction prediction branches \citep{chen_masklab_2018}. The direction prediction provides useful information to distinguish instances of the same semantic label, allowing the model to further refine the instance segmentation results.

Attention mechanisms have been integrated into the segmentation models to learn the weights of multiscale features at each pixel location. A multistage context refinement network introduces a context attention refinement module that is composed of two parts, context feature extraction and context feature refinement \citep{liu_multi-stage_2023}. The context feature extraction captures both local and global context information, fuses both contextual information and passes it to the context feature refinement while the context feature refinement removes redundant information and generates a refined feature representation, improving the utilization of contextual information. The context attention is added to the skip connection between the encoder and the decoder. Handcrafted features are often abandoned for automatic feature extraction using convolutional networks. However, it is argued that the interpretability and domain-specific knowledge embedded in handcrafted features can provide valuable insights. To this end, an attention module based on the covariance statistic is introduced to model the dependencies between local and global context of the input image \citep{liu_covariance_2022}. Two types of attention are introduced: spatial covariance attention focuses on the spatial distribution and channel covariance attention attends to the important channels. Furthermore, the covariance attention does not require feature shape conversion, hence significantly reducing the space and time complexity of the model.

\begin{figure*}[h!]
    \centering
    \includegraphics[scale=0.7]{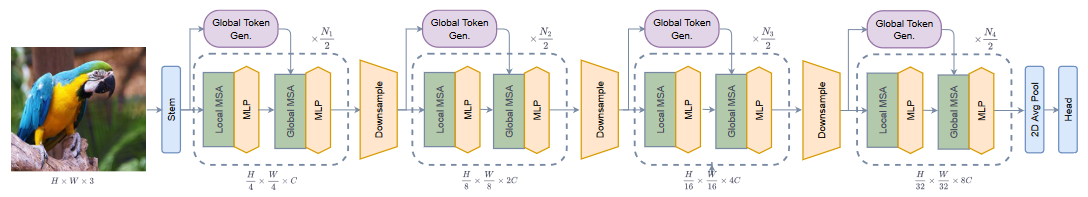}
    \caption{The architecture of global context ViT \citep{hatamizadeh_global_2023}.}
    \label{fig_img_seg_GC-ViT}
\end{figure*}

The convolutional layers use local receptive fields to process input data which can be effective for exploiting spatial patterns and hierarchical features but may find it difficult to capture global relationships across the entire image. The ViT has been leveraged to mitigate this issue in semantic segmentation \citep{strudel_segmenter_2021}. Specifically, the input image is divided into patches and treated as input to the transformer to capture the long-range relationship between the patches, significantly improving the prediction of the segmentation map. Global context ViT aims to address the lack of ViT’s ability to leverage local spatial features \citep{hatamizadeh_global_2023}. As shown in Figure \ref{fig_img_seg_GC-ViT}, the transformer consists of local and global self-attention modules. The role of global self-attention is to capture the global contextual information from different image regions while the short-range information is captured by the local self-attention. Multiscale feature representation is crucial for accurate semantic segmentation. However, the transformer often combines the features without considering their appropriate (optimal) scales, thus affecting the segmentation accuracy. Transformer scale gate is a module proposed to address the issue of selecting an appropriate scale based on the correlation between patch-query pairs \citep{shi_transformer_2023}. The transformer takes attention (correlation) maps as input and calculates the weights of the multiscale features for each image patch, allowing the model to adaptively choose the optimal scale for each patch. Table \ref{table_summary_imgseg_studies} lists the summary of state-of-the-art image segmentation.

\begin{table*}[htbp]
\centering
\caption{Summary of state-of-the-art image segmentation.}
\scriptsize
\begin{tabularx}{\textwidth}{p{1.5cm} X X}
 \hline
 Ref. & Description & Results (Datasets and Metrics) \\ 
 \hline
 \citep{liu_multi-stage_2023}, 2023 & The study proposed a multistage context refinement network with a context attention refinement module that improves feature representation by capturing local and global context, fusing information, and removing redundancies, with attention integrated into the encoder-decoder skip connections & PASCAL VOC2012\newline mIoU: 79.05\newline ADE20K MI\newline mIoU: 41.25\newline Cityscapes\newline mIoU: 79.42 \\ 
 \hline
 \citep{liu_covariance_2022}, 2022 & This study introduced a covariance attention to model the local and global dependency for the feature maps by formulating a covariance matrix, providing complementary information to improve segmentation performance & Cityscapes\newline mIoU: 82.9\newline Pascal-Context\newline mIoU: 51.7\newline ADE20K\newline mIoU: 43.78 \\
 \hline
 \citep{strudel_segmenter_2021}, 2021 & This study introduced a transformer for semantic segmentation & Cityscapes\newline mIoU: 81.3\newline Pascal-Context\newline mIoU: 59.0\newline ADE20K\newline mIoU: 53.63 \\
 \hline 
 \citep{hatamizadeh_global_2023}, 2023 & This study proposed an improved vision transformer architecture that employes alternating global and local self-attentions to effectively capture both local and global spatial information & ADE20K\newline mIoU: 49.2 \\
 \hline 
 \citep{shi_transformer_2023}, 2023 & This study introduced TSG module that uses attention maps to calculate multiscale feature weights for each image patch, enabling the model to adaptively select the optimal scale based on patch-query correlation & Cityscapes\newline mIoU: 83.6\newline Pascal-Context\newline mIoU: 64.9\newline ADE20K\newline mIoU: 56.93 \\
 \hline 
\end{tabularx}
\label{table_summary_imgseg_studies}
\end{table*}

\subsubsection{Image Generation}
Image generation refers to the process of creating images based on input texts. Generally, the task can be divided into three stages. The first stage is extracting features from the input text, followed by generating the image and finally controlling the image generation process to ensure the output meets specific criteria and constraints. This section focuses on the progress made in the development of deep learning models of the second stage (image generation) since it directly impacts the quality of the generated images. Variational autoencoder is one of the earliest deep learning models that is capable of generating images \citep{kingma_auto-encoding_2013}. Variational autoencoder learns to generate data by capturing the underlying (Gaussian) distribution of the training data. During the generation process, the distribution parameters are sampled and passed to the decoder to generate the output image. Although the generated images are blurry and unsatisfactory, it has shown a lot of potential in image generation tasks. The introduction of GAN significantly improved the quality of generated images. GAN consists of two connected neural networks, a generator and a discriminator that are trained simultaneously in a competitive manner \citep{goodfellow_generative_2014}. The generator learns to generate realistic images to fool the discriminator, while the discriminator learns to distinguish between fake and real images. The generated images are less blurry and more realistic. Several enhanced models have been proposed to improve its usability and overall performance such as conditional GAN \citep{odena_conditional_2017} which allows us to tell what image to be generated, and the deep convolutional GAN (DCGAN) \citep{radford_unsupervised_2015} which provides a more stable structure for image generation. DCGAN is the basis of many subsequent improvements in GANs.

StackGAN divides the process of image generation into two stages \citep{zhang_stackgan_2017}. Stage-I generates a low-resolution image by creating basic shapes and colors and the background layout using the random noise vector. Stage-II completes the details of the image and produces a high-resolution photo-realistic image. StackGAN++ is the enhanced model of StackGAN whereby it consists of multiple generators with shared parameters to generate multiscale images \citep{zhang_stackgan_2018}. The generators have a progressive goal with the intermediate generators generating images of varying sizes and the deepest generator producing the photo-realistic image. HDGAN is a generative model featuring a single-stream generator with hierarchically nested discriminators at intermediate layers \citep{zhang_photographic_2018}. These layers, each connected to a discriminator, generate multiscale images. The lower resolution outputs are used to learn semantic image structures while the higher resolution outputs are used to learn fine-grained details of the image. StackGAN heavily relies on the quality of the generated image in Stage-I. DM-GAN incorporates a memory network for image refinement to cope with badly generated images in Stage-I \citep{zhu_dm-gan_2019}. The memory network dynamically selects the words that are relevant to the generated image, and then refines the details to produce better photo-realistic images.

\begin{figure*}[h!]
    \centering
    \includegraphics[scale=0.6]{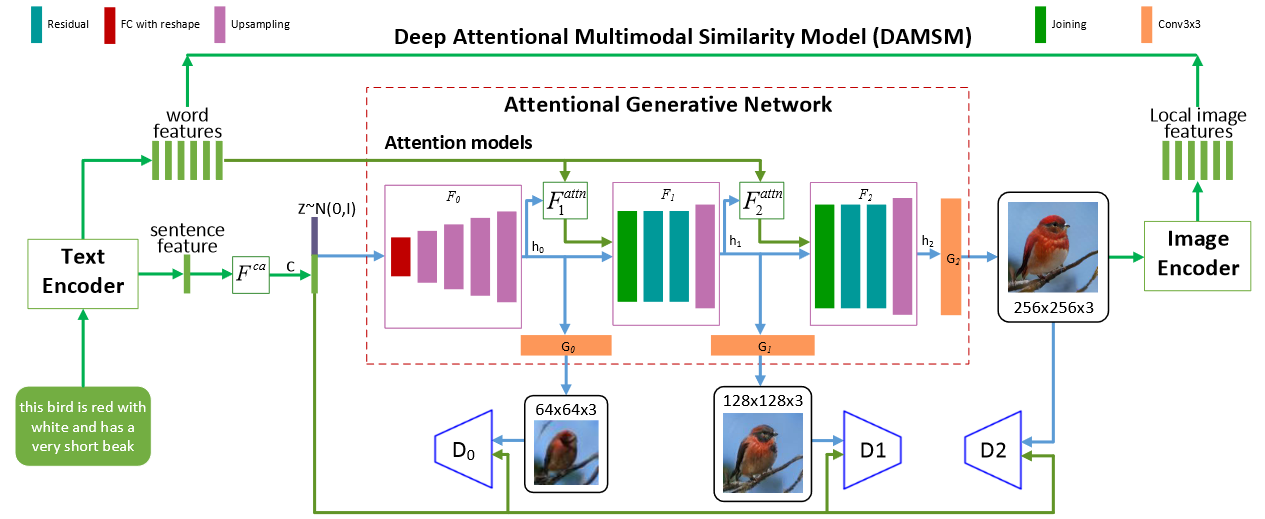}
    \caption{The architecture of AttnGAN \citep{xu_attngan_2018}.}
    \label{fig_image_gen_AttnGAN}
\end{figure*}

AttnGAN is the first to incorporate attention mechanisms into the multiple generators to focus on words that are relevant to the generated image \citep{xu_attngan_2018}. To this end, in addition to encoding the whole sentence into a global sentence vector, the text encoder encodes each word into a word vector as shown in Figure \ref{fig_image_gen_AttnGAN}. Then, the image vector is used to attend to the word vector using the attention modules at each stage of the multistage generators. Furthermore, AttnGAN introduces a loss function to compute the similarity between the generated image and the associated sentence, improving the performance of image generation. A similar work is reported whereby the model known as ResFPA-GAN, incorporates attention modules into the multiple generators \citep{sun_resfpa-gan_2019}. Specifically, a feature pyramid attention module is proposed to capture high semantic information and fuse the multiscale feature, enhancing the overall performance of the model. DualAttn-GAN improves AttnGAN by incorporating visual attention modules to focus on important features along both spatial and channel dimensions \citep{cai_dualattn-gan_2019}. This allows the model to better understand and capture both the context of the input sentence and the fine details of the image, resulting in more realistic image generation.

\begin{figure*}[h]
    \centering
    \includegraphics[scale=0.6]{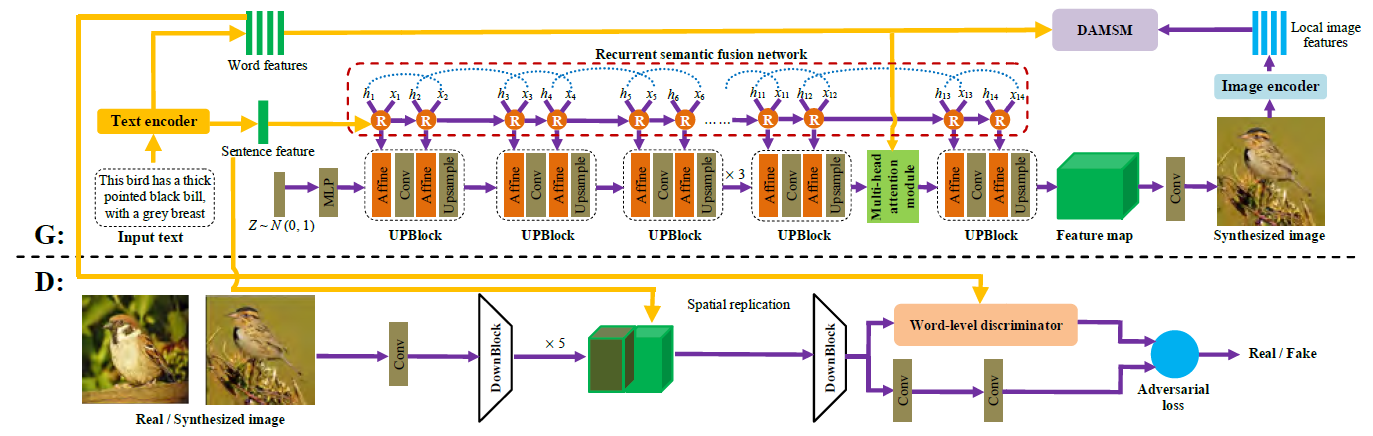}
    \caption{The architecture of DMF-GAN \citep{yang_dmf-gan_2024}.}
    \label{fig_image_gen_DMF-GAN}
\end{figure*}

Although multistage generators improve image generation performance by leveraging multiscale representation, the generated images may contain fuzzy shapes with coarse features. DF-GAN replaces the multistage generators with a single-stage deep generator featuring residual connections and trained with hinge loss \citep{tao_df-gan_2022}. Furthermore, DF-GAN introduces a regularization strategy on the discriminator that applies a gradient penalty on real images with matching text, allowing the model to generate more text-matching images. DMF-GAN an improved DF-GAN, incorporates three novel components designed to leverage semantic coherence between the input text and the generated image \citep{yang_dmf-gan_2024}. The first component is the recurrent semantic fusion module, which models long-range dependencies between the fusion blocks. The second component is the multi-head attention module, which is placed towards the end of the generator to leverage the word features, forcing the generator to generate images conditioned on the relevant words. The last component is the word-level discriminator, which provides fine-grained feedback to the generator, facilitating the learning process and improving the overall quality of the generated images. Figure \ref{fig_image_gen_DMF-GAN} shows the architecture of DMF-GAN. The process of image generation involves feeding a noise vector to the generator at the very beginning of the network. However, as the generator goes deeper, the noise effect may be diminished, affecting the diversity of the image generation results. To mitigate this issue, DE-GAN incorporates a dual injection module into the single-stage generator \citep{jiang_-gan_2024}. The dual injection module consists of two text fusion layers followed by a noise broadcast operation. The text fusion layer takes the sentence embedding and fuses it with the input feature map using the fully connected layer. Then noise is injected into the output feature map to retain the randomness in the generation process, improving diversity and generalization of the model. Table \ref{table_summary_imggen_studies} lists the summary of state-of-the-art image generation.

\begin{table*}[htbp]
\centering
\caption{Summary of state-of-the-art image generation.}
\scriptsize
\begin{tabularx}{\textwidth}{p{1.5cm} X X}
 \hline
 Ref. & Description & Results (Datasets and Metrics) \\ 
 \hline
 \citep{tao_df-gan_2022}, 2022 & This study introduced a one-stage GAN architecture, employing a Target-Aware Discriminator to enhance text-image semantic consistency and utilizes deep text-image fusion blocks for effective feature integration, resulting in improved image authenticity and relevance & CUB\newline IS: 5.10, FID: 14.81\newline COCO\newline FID: 19.32 \\ 
 \hline
 \citep{yang_dmf-gan_2024}, 2024 & This study proposed a deep multimodal fusion generative adversarial networks that allow effective semantic interactions for fusing text information into the image synthesis process & CUB\newline IS: 5.42, FID: 13.21\newline COCO\newline IS: 36.72, FID: 15.83 \\
 \hline
 \citep{jiang_-gan_2024}, 2024 & This study proposed a deep learning-based method which utilizes BERT to extract sentiment information and analyze sentiment at the sentence and word level with SVM for sentiment classification & CUB\newline IS: 4.86, FID: 18.94\newline COCO\newline IS: 18.33, FID: 28.79 \\
 \hline 
\end{tabularx}
\label{table_summary_imggen_studies}
\end{table*}

\subsection{Natural Language Processing}
Natural language processing (NLP) refers to the field of AI that concerns with enabling computers to process, analyze and interpret human languages to extract useful information. Some of the common tasks in NLP are machine translation, text classification and text generation. Deep learning has been widely applied to solve real-world NLP problems. This section presents the recent advancements in deep learning models that have been designed for NLP over the past few years.

\subsubsection{Text Classification}
Text classification, known as text categorization, is a task that involves assigning predefined categories or labels to a piece of text based on its content. The task is commonly used in various applications such as document classification, sentiment analysis and spam filtering. Numerous deep learning models have been proposed for text classification in the past few decades, and multilayer perceptron is one of the earliest architectures adopted to classify documents \citep{calvo_intelligent_2000, yu_latent_2008}. The model typically has a single hidden layer with a number of units between 15 and 150. Text data is inherently sequential, as it is composed of a series of words and symbols arranged in a specific order. This property makes RNN and its variants particularly well-suited for processing and analyzing text data. In \citep{arevian_recurrent_2007}, RNN with two hidden layers, each with 6 units, is used to classify news documents into eight classes. A study was conducted to investigate the variants of RNN such as LSTM and GRU for text classification \citep{huang_optimization_2020}. The input to the model is a sequence of words of fixed length. The input sequence is also sliced into smaller subsequences of fixed length and passed to an independent model for parallelization. A convolutional layer can extract local features, allowing the model to leverage hierarchical temporal information in textual data. A hybrid model of convolutional and LSTM architecture is proposed for text classification \citep{wang_convolutional_2019}. Two parallel convolutional layers are used to extract features from word embeddings, followed by maximum pooling layers to reduce the feature dimensions. The reduced features are then concatenated and passed to LSTM for prediction.

Although CNN and RNN provide excellent results on text classification tasks, the models lack the ability to attend to specific words based on their importance and context. To address this limitation, an attention mechanism is incorporated into the model to focus on the important features, enhancing the text classification accuracy. In \citep{liu_bidirectional_2019}, two attention modules are introduced to capture the contextual information of the feature sequence extracted by bidirectional LSTM. The first attention module attends the sequence in forward direction, while the backward sequence is attended by the second attention module. The convolutional layers are used before the bidirectional LSTM to extract features from the word embedding. The attention modules require sequential processing using RNN-based architecture such as LSTM and GRU, which may lead to information loss and distorted representations, particularly in long sequence. Furthermore, the attention modules focus on inter-sequence relationships between the input sequence and the target, ignoring the intra-sequence relationships or the dependencies between the words. In \citep{lin_structured_2017}, the deep learning model is integrated with self-attention to capture the intra-sequence relationships between the features in the sequence. A multilayer of bidirectional LSTMs is utilized to extract the feature sequence from the word embedding before the self-attention module attends the feature sequence to compute the attention weights. To further improve the overall performance, a multichannel features consisting of three input pipelines is introduced \citep{li_bidirectional_2020}. Each pipeline concatenates the word vector with a feature vector derived from the input sequence such as the word position, part-of-speech and word dependency parsing. The input pipeline is connected to bidirectional LSTM, followed by a self-attention module to learn the dependencies between the features in the sequence.

The transformer is a deep learning architecture that transforms sequential data using self-attention mechanisms, allowing long-range dependencies and complex patterns to be captured. The architecture is the basis of various advanced deep learning models and the Bidirectional Encoder Representations from Transformers popularly known as BERT is one of the examples that leverage transformer for pre-training on large scale textual data \citep{devlin_bert_2018}. BERT is a bidirectional transformer encoder that is designed for various NLP tasks, capable of capturing the contextual information from both preceding and succeeding words in the input sequence. Several improvements have been made to BERT to enhance its overall performance, such as ALBERT \citep{lan_albert_2019}, RoBERTa \citep{liu_roberta_2019} and DeBERTa \citep{he_deberta_2020}. The improvements are centered around refining the pre-training approaches such as dynamic masking of the training instances, training with a block of sentences and representing each input word using two vectors, both content and position of the word. Most of the recent works leverage BERT and its variants to capture effective feature representation of the input sequence. In \citep{rodrawangpai_improving_2022}, BERT and its variants are leveraged to capture the long-range dependencies of the input tokens. The features are then passed to a layer normalization and a linear fully connected layer with dropout for classification. Similar work is reported in \citep{murfi_bert-based_2024} whereby BERT is used to extract the features and the features are then passed to a hybrid of convolutional and recurrent neural networks. The traditional machine learning algorithms have been used to classify the features extracted by BERT \citep{hao_sentiment_2023}. The study shows machine learning algorithms can effectively leverage the rich contextual features extracted by BERT for downstream classification tasks. 

\begin{figure*}[h!]
    \centering
    \includegraphics[scale=0.65]{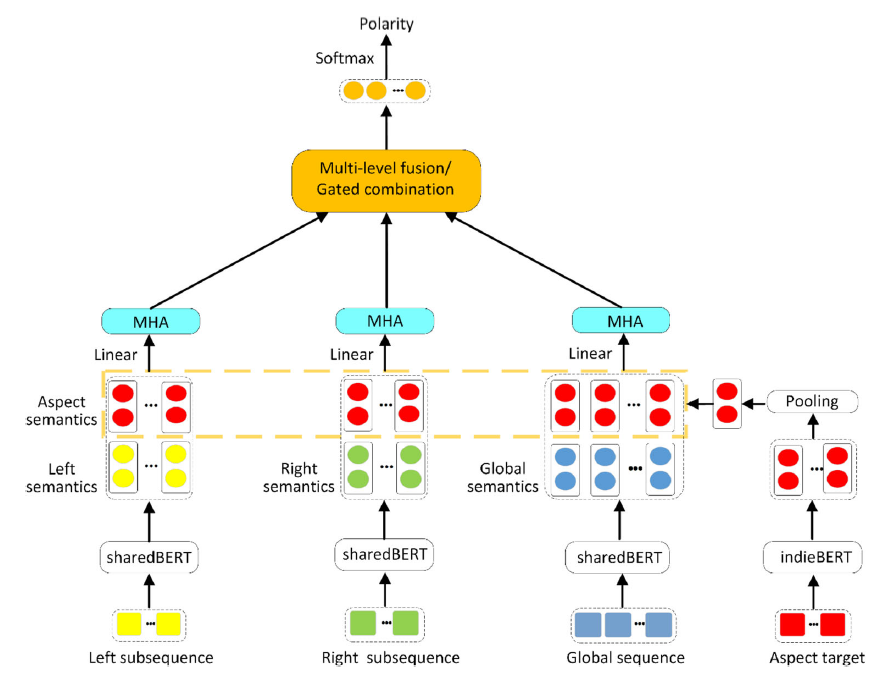}
    \caption{The architecture of BERT-MSL \citep{zhu_bert-based_2023}.}
    \label{fig_text_clf_BERT-MSL}
\end{figure*}

In text classification, the text labels can help in capturing the words relevant to the classification. The label-embedding attentive model is one of the earliest attempts to joint learn the label and word embeddings in the same latent space and measure the compatibility between labels and words using cosine similarity \citep{wang_joint_2018}. The joint embedding allows the model to capture more effective text representations, increasing the overall performance of the model. LANTRN is a deep learning model that leverages label embedding extracted by BERT and entity information e.g. person name and organization name for text classification \citep{yan_r-transformer_bilstm_2023}. The entity recognition module is based on bidirectional LSTM and conditional random field layers to calculate the probability of each word in each entity label. The model introduces a label embedding bidirectional attention to learn the attention weights of token-label and sequence-label pairs. Furthermore, a transformer consisting of RNN and a multi-head self-attention mechanism is introduced to learn local short-term dependencies of multiple short text sequences and long-term dependencies of the input sequence. Aspect refers to a specific attribute of an entity within the text, and incorporating this information enhances the model’s understanding of the nuances of the text. BERT-MSL is a multi-semantic deep learning model with aspect-aware enhancement and four input pipelines: left sequence, right sequence, global sequence and aspect target \citep{zhu_bert-based_2023}. The aspect-aware enhancement module takes the features extracted by BERT, and performs average pooling followed by a linear transform. Then the output is concatenated with the outputs produced by the local and global semantic learning modules. The concatenated features are then jointly attended by a multi-head attention for text classification. Figure \ref{fig_text_clf_BERT-MSL} shows the architecture of BERT-MSL. Table \ref{table_summary_textclf_studies} lists the summary of state-of-the-art text classification.

\begin{table*}[h!]
\centering
\caption{Summary of state-of-the-art text classification.}
\scriptsize
\begin{tabularx}{\textwidth}{p{1.5cm} X X}
 \hline
 Ref. & Description & Results (Datasets and Metrics) \\ 
 \hline
 \citep{rodrawangpai_improving_2022}, 2022 & The study proposed an enhanced RoBERTa by incorporating layer normalization and dropout to enable smoother gradients, faster training and reduced overfitting, thus, improving model generalization & NEISS\newline Accuracy: 0.98 \\ 
 \hline
 \citep{murfi_bert-based_2024}, 2024 & The study employed IndoBert to extract features from text data which were then passed to a hybrid deep learning model for sentiment analysis & Shopee\newline Accuracy: 0.8612 (CNN-GRU)\newline Tokopedia\newline Accuracy: 0.8768 (LSTM-CNN)\newline Lazada\newline Accuracy: 0.8710 (LSTM-CNN) \\
 \hline
 \citep{hao_sentiment_2023}, 2023 & This study proposed a deep learning-based method which utilizes BERT to extract sentiment information and analyze sentiment at the sentence and word level with SVM for sentiment classification & NLPCC-SCDL\newline Accuracy: 95.12\% \\
 \hline
 \citep{yan_r-transformer_bilstm_2023}, 2023 & This study proposed LANRTN model for text classification, integrating an R-Transformer with label embedding, attention mechanisms, and an entity recognition model to capture both global dependencies and label-aware contextual information & Reuters Corpus Volume I\newline Micro F1-score: 0.893, Arxiv Academic Paper Dataset\newline Micro F1-score: 0.718 \\
 \hline
 \citep{zhu_bert-based_2023}, 2023 & The paper proposed a BERT-MSL model for aspect-based sentiment analysis, integrating multiple semantic learning modules and merging them using multi-head self-attention and linear transformation layers for enhanced classification & 14Lap\newline Accuracy: 82.24\%, Macro-average F1-score: 78.98\%\newline 14Rest\newline Accuracy: 89.11\%, F1-score: 84.07\%\newline 15Rest\newline Accuracy: 88.53\%, Macro-average F1-score: 71.41\%\newline 16Rest\newline Accuracy: 93.09\%, Macro-average F1-score: 81.24\%\newline Twitter\newline Accuracy: 74.89\%, Macro-average F1-score: 73.43\% \\
 \hline
\end{tabularx}
\label{table_summary_textclf_studies}
\end{table*}

\subsubsection{Neural Machine Translation}

\begin{figure*}[h]
    \centering
    \includegraphics[scale=0.45]{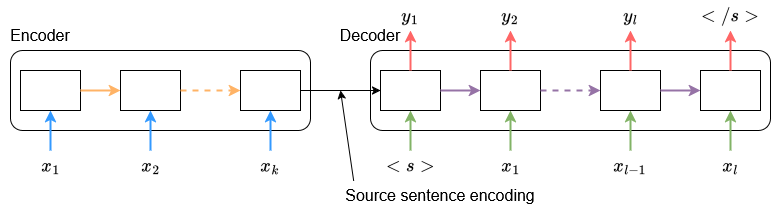}
    \caption{The architecture of an encoder-decoder. Adapted from \citep{stahlberg_neural_2020}.}
    \label{fig_nmt_encoder_decoder}
\end{figure*}

Neural machine translation (NMT) refers to the automated process of translating text from one language to another language. Numerous deep learning models have been proposed for NMT, which can be categorized into RNN-based and CNN-based models. One of the first successful RNN-based models is the encoder-decoder \citep{cho_learning_2014, sutskever_sequence_2014}. The model consists of two connected subnetworks (the encoder and the decoder) for modelling the translation process, as shown in Figure \ref{fig_nmt_encoder_decoder}. The encoder reads the source sentence word by word and produces a fixed-length context vector (final hidden state). This process is known as source sentence encoding, as shown in the figure. Given the context vector, the decoder generates the target sentence (translation) word by word. This modelling of the translation can be seen as a mapping between the source sentence to the target sentence via the intermediate context vector in the semantic space. The context vector represents the summary of the input sequence’s semantic meaning, providing a compressed representation that captures the essence of the source sentence. However, the compression process can sometimes result in the loss of information, especially those early in the sequence. Bidirectional RNN may mitigate the loss of information by modelling the sequence in reverse order. However, the problem can persist, particularly in cases where the input is a long sequence.

\begin{figure*}[h!]
    \centering
    \includegraphics[scale=0.5]{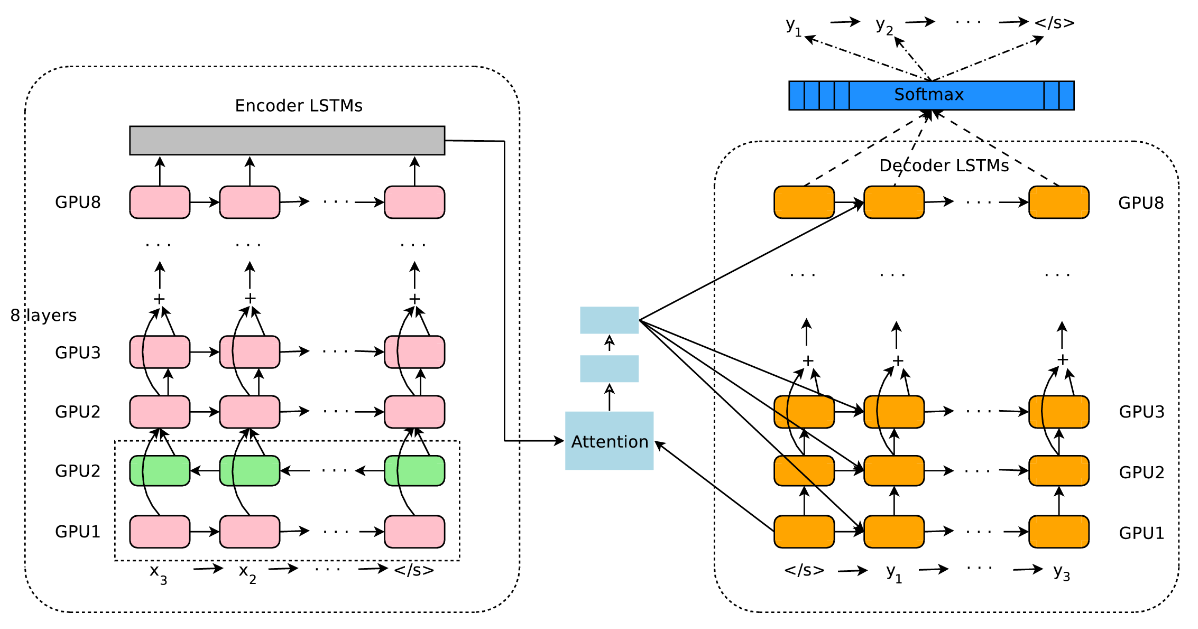}
    \caption{The architecture of Google Neural Machine Translation \citep{wu_googles_2016}.}
    \label{fig_nmt_GNMT}
\end{figure*}

Attention mechanism was introduced to solve the problem of learning long input sequences \citep{bahdanau_neural_2014}. Attention alleviates this issue by attending on different words of the input sequences when predicting the target sequences at each time step. Unlike the standard encoder-decoder model, attention derives the context vector from the hidden states of both the encoder and decoder, and the alignment between the source and target. This mechanism allows the model to focus on the important words, increasing the overall accuracy of the translation. Several alignment score functions have been proposed for calculating the attention weights. Some popular functions are additive \citep{bahdanau_neural_2014}, dot-product, location-based \citep{luong_effective_2015}, and scaled dot-product \citep{vaswani_attention_2023}. The attention weights are calculated by attending to the entire hidden states of the encoder. This attention, also known as global attention, is computationally expensive. Instead of attending to all hidden states, local attention attends to a subset of hidden states, thus reducing the computational cost \citep{luong_effective_2015}. Google Neural Machine Translation is a popular encoder-decoder model with an attention mechanism that significantly improves the accuracy of machine translation \citep{wu_googles_2016}. As shown in Figure \ref{fig_nmt_GNMT}, the model consists of a multilayer of LSTMs with eight encoder and decoder layers and an attention connection between the bottom layer of the decoder to the top layer of the encoder. Furthermore, to deal with the challenging words to predict, a word is tokenized into subwords e.g. feud is broken down into ``fe'' and ``ud'', allowing the model to generalize well to new and uncommon words. A year later, the self-attention mechanism was proposed, significantly improving the overall accuracy of machine translation \citep{vaswani_attention_2023}. Self-attention, also known as intra-attention, allows the deep learning model to capture the dependencies between the input words. The self-attention mechanism is the fundamental building block of the transformer model, which has since become a cornerstone in natural language processing and other domains.

Despite the success of transformer, the model falls short in capturing nuances of human language and struggles with tasks requiring deeper understanding of context. This can be especially challenging when the tasks involve formality, colloquialism, and subtle cultural references that may not directly equivalent in the target language, resulting in inaccurate translation or losing the original meaning. One of the approaches to include context into the input sequence is concatenating the current source sentence with the previous (context) sentences and feeding the whole input to the transformer \citep{lupo_encoding_2023}. The model is trained to predict the translated sentence including the context translation. At inference time, only the translation is considered while the context translation is discarded. Furthermore, the approach encodes the sentence position and segment-shifted position to improve the distinction between current sentences and context sentences. In \citep{rippeth_improving_2023}, the source sentence is prefixed with the summary of the document to contextualize the input sentence. The summary is the set of salient words that represents the essence of the document, resolving ambiguity associated with the translation. A study was conducted to determine the optimal technique of aggregating contextual features \citep{wu_study_2022}. Three techniques were studied namely concatenation mode, flat mode and hierarchical mode, and the experimental results indicate that concatenation mode achieved the best results. In \citep{kim_towards_2023}, a training method is introduced to train the deep learning machine translation model to generate translation involving honorific words. The training method indicates the honorific context in the target sentence using an honorific classifier to guide the model to attend to the related tokens. Unlike other studies where the context features are included by concatenation, the training method assigns weights to the context tokens indicated by the honorific classifier. This allows the model to generate a more accurate translation with honorifics. 

The performance of transformers relies on large-scale training data. However, for the vast majority of languages, only limited amounts of training data exist. To mitigate this problem, recent studies introduce shallow transformer architectures \citep{gezmu_transformers_2022}, explore the effect of hyperparameter finetuning \citep{araabi_optimizing_2020} and leveraging visual input as contextual information for the translation task \citep{meetei_cues_2023}. The semi-supervised neural machine translation is used for translating low resource Egyptian Arabic dialects to modern standard Arabic \citep{faheem_improving_2024}. The study utilizes three datasets: bilingual Egyptian-standard Arabic and two monolingual Egyptian Arabic and standard Arabic. First, a transformer-based model is trained in a supervised manner using the bilingual dataset. Then, the model is trained in an unsupervised manner using both monolingual datasets. The unsupervised approach employs an Encoder-Decoder model with Byte Pair Encoding for tokenization and handling unknown words. The monolingual corpora are merged and used to improve the model by iteratively generating synthetic sentence pairs between Egyptian Arabic and standard Arabic, allowing the model to learn the correspondence between them. In \citep{li_towards_2024}, the authors exploit monolingual corpus to enhance the bilingual dataset for model training. Furthermore, a new loss function is proposed as a replacement for traditional cross-entropy loss, allowing the model to learn with uncertainty in the presence of noise. Additionally, contrastive re-ranking is employed to refine translation results by selecting the most confident output from multiple candidates. Table \ref{table_summary_nmt_studies} lists the summary of state-of-the-art neural machine translation.

\begin{table*}[htbp]
\centering
\caption{Summary of state-of-the-art neural machine translation.}
\scriptsize
\begin{tabularx}{\textwidth}{p{1.5cm} X X}
 \hline
 Ref. & Description & Results (Datasets and Metrics) \\ 
 \hline
 \citep{lupo_encoding_2023}, 2023 & This study incorporated context into the input sequence by concatenating the current source sentence with previous context sentences, and the model is trained to predict the translation with context, while encoding sentence positions to distinguish between current and context sentences & En$\rightarrow$De\newline ContraPro: 82.54\newline En$\rightarrow$Ru\newline Voita: 75.94 \\ 
 \hline
 \citep{rippeth_improving_2023}, 2023 & The paper proposed a method to enhance translation accuracy of ambiguous words by prefixing source sentences with salient words extracted from related pseudo-documents, thereby providing additional context without altering standard model architectures & En$\rightarrow$De\newline BLEU: 22.0, COMET: 0.785 \\
 \hline
 \citep{wu_study_2022}, 2022 & The paper investigated three most common methods to aggregate the contextual features: concatenation mode, flat mode and hierarchical mode & En$\rightarrow$De\newline BLEU: 30.89\newline De$\rightarrow$En\newline BLEU: 36.84\newline En$\rightarrow$Zh\newline BLEU: 20.06\newline Zh$\rightarrow$En\newline BLEU: 20.50 \\
 \hline
 \citep{kim_towards_2023}, 2023 & The paper introduced formality classifier to incorporate formality-related contexts into the model training & AI-HUB (En$\rightarrow$Ko)\newline BLEU: 27.90\newline OpenSubtitles201 (En$\rightarrow$Ko)\newline BLEU: 20.70\newline BSD, AMI and OpenSubttles2018 (En$\rightarrow$Ja)\newline BLEU: 13.67 \\
 \hline
 \citep{faheem_improving_2024}, 2024 & This study employed a semi-supervised neural machine translation approach to translate low-resource Egyptian Arabic dialects to Modern Standard Arabic, utilizing bilingual and monolingual datasets, with a transformer-based model trained in both supervised and unsupervised manners to generate synthetic sentence pairs and improve translation accuracy & Egyptian-standard Arabic\newline BLEU: 24 (300 word embedding)\newline BLEU: 29.5 (512 word embedding) \\ 
 \hline
 \citep{li_towards_2024}, 2024 & The study proposed a new loss function to handle uncertainty in noisy data, and used contrastive re-ranking to select the most confident translation from multiple candidates & Zh$\rightarrow$Ma\newline BLEU: 28.12\newline Ma$\rightarrow$Zh\newline BLEU: 23.53\newline Zh$\rightarrow$In\newline BLEU: 28.91\newline In$\rightarrow$Zh\newline BLEU: 22.76 \\ 
 \hline
\end{tabularx}
\label{table_summary_nmt_studies}
\end{table*}

\subsubsection{Text Generation}
Text generation refers to the process of creating texts based on a given input, whereby the input can be in the form of texts, images, graphs, tables or even tabular data. Due to the various forms of inputs, text generation has a wide range of applications, including creative writing, image captioning and music generation. This section focuses on the progress made in text-to-text generation tasks such as question answering, dialogue generation and text summarization. The recurrent neural network and its variants play an important role in text generation tasks for their strong ability to model sequential data. One of the earliest works on question answering is based on the RNN-based encoder-decoder model, whereby the encoder takes the question embedding and processes it using bidirectional LSTM, and the decoder generates the corresponding answer \citep{nie_attention-based_2017}. Additionally, to prevent semantic loss and enable the model to focus on the important words in the input sequence, a convolution operation is applied to the word embedding, and an attention mechanism is then used to attend to the output of the convolution operation. Similar work is reported in \citep{yin_neural_2015} in which a knowledge-based module is introduced to calculate the relevance score between the question and the relevant facts in the knowledge base. This improves the text (answer) generation by the decoder. Another work is described in \citep{li_deep_2016} where an encoder-decoder with attention for dialogue generation is optimized using reinforcement learning. The model is first trained in a supervised learning manner and then improved using the policy gradient method to diversify the responses. Ambiguous content in question answering sentences is a challenge in text generation and can lead to incorrect and uncertain responses. Cross-sentence context aware bidirectional model introduces a parallel attention module to compute the co-attention weights at the sentence level, accounting for the relationships and similarities in the question and the answer \citep{wu_building_2020}.

The transformer has been leveraged for text generation tasks. An incremental transformer-based encoder is proposed to incrementally encode the historical sequence of conversations \citep{li_incremental_2019}. The decoder is a two-pass decoder that is based on the deliberation network, generates the next sentence. The first pass focuses on contextual coherence of the conversations, while the second pass refines the output of the first pass. BERT and ALBERT have been used as pre-trained models for question answering task \citep{alrowili_biom-transformers_2021}. The study found that the performance of the models is sensitive to random assignment of the initial weights especially on small datasets \citep{alrowili_exploring_2022}. T-BERTSum is a model based on BERT, designed to address the challenge of long text dependence and leveraging latent topic mapping in text summarization \citep{ma_t-bertsum_2021}. The model integrates a neural topic module to infer topics and guide summarization, uses a transformer network to capture long-range dependencies and incorporates a multilayer of LSTM for information filtering. 

The generated texts often lack diversity and may exhibit repetitive patterns. To address this issue, feature-aware conditional GAN (FA-GAN) is proposed for controllable category text generation \citep{li_feature-aware_2023}. The generator consists of BERT, a category encoder, a relational memory core (RMC) decoder. BERT acts as a feature encoder, improves contextual representation and mitigates mode collapse while the category encoder embeds categorical information for text generation. The RMC decoder utilizes a self-attention mechanism to capture interactions between features, generating more expressive and diverse texts. The discriminator includes an additional classification head to ensure the generated texts match specified categories. Traditional text generation has largely focused on binary style transfers, but real-world applications require capturing diverse styles, which existing methods struggle to achieve. In \citep{kwon_class_2024}, a multi-class conditioned text generation model is proposed using a transformer-based decoder with an adversarial module, a style attention module and a generation module. The adversarial module extracts style-excluded representation from the input text. The style attention mechanism then introduces the desired style into the text representation through concat attention mechanism, producing a conditioned representation. Finally, the generation module utilizes the conditioned representation to generate stylistically diverse text.

Exploiting domain knowledge is essential in reducing the semantic gap between the deep learning models and the text corpus. KeBioSum is a knowledge infusion framework to inject domain knowledge into the pre-trained BERTs for text summarization \citep{xie_pre-trained_2022}. In the framework, the relevant information is detected and extracted from the domain knowledge, generating label sequences of the sentences. The label data is then used to train the text summarization model using discriminative and generative training approaches, infusing the knowledge into the model. Large language models (LLMs) have been used for text generation with exceptional quality and diversity. LLMs, trained on extensive corpus data, have a deep understanding of human language, allowing them to interpret and generate texts. In \citep{hajipoor_gptgan_2025}, GPTGAN is introduced that leverages an LLM as a guiding mentor to a GAN-Autoencoder model for text generation. The approach involves the GPT model to generate a sequence of words given a subset of the input sequence, and the generated text is mapped into latent space by the transformer-based encoder. The latent representation is then used by the transformer-based decoder to generate the text. Furthermore, local discriminators are introduced to refine the text generation because the generated words by the GPT model may be inaccurate. Table \ref{table_summary_textgen_studies} lists the summary of state-of-the-art text generation.

\begin{table*}[htbp]
\centering
\caption{Summary of state-of-the-art text generation.}
\scriptsize
\begin{tabularx}{\textwidth}{p{1.5cm} X X}
 \hline
 Ref. & Description & Results (Datasets and Metrics) \\ 
 \hline
 \citep{ma_t-bertsum_2021}, 2021 & The paper proposed a text summarization model that integrates neural topic model with BERT to guide the text summarization based on inferred topics. It leverages transformers for long-term dependencies and combines LSTM layers with a gated network to enhance extractive and abstractive summarization of social texts & CNN/Daily Mail\newline Rouge-1: 43.58, Rouge-2: 20.45, Rouge-L: 34.60\newline Xsum\newline Rouge-1: 39.90, Rouge-2: 17.48, Rouge-L: 29.85 \\ 
 \hline
 \citep{li_feature-aware_2023}, 2023 & This study proposed FA-GAN featuring BERT for contextual representation, a category encoder for embedding categorical information, and an RMC decoder for generating diverse texts, while the discriminator ensures category consistency & MR-10\newline BLEU-2: 0.560\newline MR-20\newline BLEU-2: 0.674\newline AM-30\newline BLEU-2: 0.748  \\
 \hline
 \citep{kwon_class_2024}, 2024 & This study proposed a multi-class conditioned text generation model, combining a transformer-based decoder with an adversarial module, style attention, and a generation module to produce stylistically diverse text & Amazon (neg$\rightarrow$pos)\newline self-BLEU: 0.257, S-ACC: 0.375 \newline Amazon (pos$\rightarrow$neg)\newline self-BLEU: 0.096, S-ACC: 0.191 \newline YELP (neg$\rightarrow$pos)\newline self-BLEU: 0.257, S-ACC: 0.198\newline YELP (pos$\rightarrow$neg)\newline self-BLEU: 0.081, S-ACC: 0.207\newline *S-ACC (style accuracy) \\
 \hline
 \citep{xie_pre-trained_2022}, 2022 & The paper proposes a knowledge-based summarization model that integrates medical knowledge into the pretrained models using a lightweight knowledge adapter. It employs generative and discriminative training to predict and reconstruct PICO elements, enhancing domain-specific text summarization & CORD-19\newline Rouge-1: 32.04, Rouge-2: 12.61, Rouge-L: 29.10\newline PubMed-Long\newline Rouge-1: 36.39, Rouge-2: 16.27, Rouge-L: 33.28 \\
 \hline
 \citep{hajipoor_gptgan_2025}, 2025 & This study proposed GPTGAN, an approach that enhances adversarial text generation using a GPT model as a mentor to the GAN model and employing a multiscale discriminator framework to balance text quality and diversity & MSCOCO\newline BLEU: 98.7\newline WMTNews\newline BLEU: 93.4\newline Persian COCO\newline BLEU: 97.9 \\ 
 \hline
\end{tabularx}
\label{table_summary_textgen_studies}
\end{table*}

\subsection{Time Series and Pervasive Computing}
Pervasive computing, often referred to as ubiquitous computing, is the process of integrating computer technology into everyday objects and surroundings so that they become intelligent, networked, and able to communicate with one another to offer improved services and functionalities \citep{weiser1991computer}. According to He et al. \citep{he2020developing}, the role of pervasive computing is foremost in the field where it provides the ability to distribute computational services to the surroundings where people work, leading to trust, privacy, and identity. Examples of pervasive computing applications include smart homes with connected appliances, wearable devices that monitor health and fitness, smart cities with sensor networks for traffic management, and industrial applications that utilize the Internet of Things (IoT) for monitoring and control. Generally, the continuous interaction of interconnected devices in pervasive computing often results in time series data, which captures the evolution of various parameters over time.

For instance, medical sensors, such as electrocardiograms (ECG) and electroencephalograms (EEG), generate time series data that contain critical diagnostic information, which deep learning can use to detect anomalies, predict diseases, and classify medical conditions with improved accuracy. Furthermore, devices such as accelerometers, magnetometers and gyroscopes, among others, can be used to capture human activity signals, which are often represented as time series of state changes \citep{ige2022survey}. In traditional machine learning, features such as mean, variance, and others are manually extracted from times series of state changes before human activity classification. However, deep learning models automatically extract features \citep{mohd_noor_feature_2021}. Furthermore, in other fields such as finance, which entail time series data, deep learning has been instrumental in stock price prediction \citep{singh2017stock}, fraud detection \citep{zhang2021hoba}, and algorithmic trading \citep{lei2020deep}, among others. Generally, deep learning networks excel at capturing intricate temporal relationships within time-series data, enabling more precise predictions and improved decision-making. Based on this, several deep learning models have been employed for feature learning across various time series and pervasive computing domains.

\subsubsection{Human Activity Recognition} 
Human activity recognition (HAR) finds application across various domains including intelligent video surveillance, environmental home monitoring, video storage and retrieval, intelligent human-machine interfaces, and identity recognition, among many others. It includes various research fields, including the detection of humans in video, estimating human poses, tracking humans, and analyzing and understanding time series data \citep{zhang2019comprehensive}. Despite the advancements in vision-based HAR, there exist inherent limitations. Generally, vision-based approaches heavily rely on camera systems, which may have restricted views or be affected by lighting conditions, occlusions, and complex backgrounds \citep{ige2022survey}. Additionally, vision-based HAR struggles with identifying actions that occur beyond the range of the camera or actions that are visually similar.

Wearable sensors offer a promising alternative to overcome these limitations. By directly capturing data from the individual, wearable sensors provide more comprehensive and accurate information about human activities. The signals obtained from wearable sensors typically represent time series data reflecting state changes in activities. Deep learning models can effectively learn from these signals, allowing for robust and accurate recognition of human activities. Moreover, wearable sensors offer the advantage of mobility, enabling activity recognition in various environments and situations where vision-based systems may be impractical or ineffective \citep{dang2020sensor}. Generally, the time series nature of signals from wearable sensors presents an excellent opportunity for deep learning models to excel in recognizing human activities with high accuracy and reliability.

Several researchers have proposed the use of CNN, RNN, and Hybrid models for deep learning-based feature learning in wearable sensor HAR. For instance, using two-dimensional CNN (Conv2D), several researchers, as seen in \citep{gao2021danhar}, \citep{gupta2021deep} and \citep{erdacs2021human}, among others, have developed deep learning models for wearable sensor HAR, despite the time series nature of the data. This is often done by treating the time series signals from wearable sensors as 2D images by reshaping them appropriately. To achieve this, researchers often organize each time series signal into a matrix format, with time along one axis and sensor dimensions along the other, before creating a pseudo-image representation, which allows the matrix to be fed into Conv2D layers for feature extraction. Conv2D layers excel at capturing spatial patterns and relationships within images, and by treating the time series data as images, these layers can learn relevant spatial features that contribute to activity recognition. The convolution operation performed by Conv2D filters across both the time and sensor dimensions, allowing the network to identify patterns and features that may be indicative of specific activities.

Even though Conv2D can effectively capture spatial dependencies within the data, it often struggles to capture temporal dependencies inherent in time series data. Since Conv2D processes data in a grid-like fashion, it does not fully leverage the sequential nature of the time series, potentially leading to less effective feature extraction for wearable sensor HAR tasks. For this reason, recent HAR architectures have leveraged one-dimensional CNN (Conv1D) and other RNNs for automatic feature extraction. Conv1D layers are specifically designed to capture temporal dependencies within sequential data. They operate directly on the time series data without reshaping it into a 2D format, allowing them to capture temporal patterns more effectively. Conv1D layers are better suited for extracting features from time series data, making them a more natural choice for wearable sensor HAR \citep{mohd_noor_feature_2021}.

For instance, Ragab et al. \citep{ragab2020random} proposed a random search Conv1D model, and evaluated the performance of the model on UCI-HAR dataset. The result showed that the model achieved a recognition accuracy of 95.40\% when classifying the six activities in the dataset. However, the model exhibited extended training times due to the dynamic nature of some activities within the dataset. To address this, \citep{banjarey2022human} proposed the use of varying kernel sizes in Conv1D layers to recognize various activities, including sitting, standing, walking, sleeping, reading, and tilting. Furthermore, a few Conv1D layers were stacked to streamline the time optimization process for training the neural network. In \citep{baraka_similarity_2023}, a signal segmentation method is proposed based on similarity between two subsequent windows. The method can not only achieve more accurate segmentation, but also distinguish between transitional and non-transitional windows. Based on the distinction, two deep learning models with convolutional and fully connected layers are trained: one for transitional activities and the other for non-transitional activities. In \citep{baraka_deep_2024}, signal segmentation is treated as a binary classification task, distinguishing windows as either transitional or non-transitional. To this end, a deep learning model with parallel Conv1D pipelines is proposed to capture temporal dependencies within the window sequence. Furthermore, some researchers have proposed models that combine machine learning algorithms with Conv1D in HAR, as seen in Shuvo et al. \citep{shuvo2020hybrid}. Their work presented a two-stage learning process to improve HAR by classifying activities into static and dynamic using Random Forest, before using Support Vector Machine to identify each static activity, and Conv1D to recognize dynamic activities. The result shows that the method achieved an accuracy of 97.71\% on the UCI-HAR dataset.

Following these advancements, several researchers have further explored Conv1D architectures with various modifications, to enhance feature learning in activity recognition systems. For example, Han et al. \citep{han2022human} developed a two-stream CNN architecture as a plug-and-play module to encode contextual information of sensor time series from different receptive field sizes. The module was integrated into existing deep models for HAR at no extra computation cost. Experiments on OPPORTUNITY, PAMAP2, UCI-HAR and USC-HAD datasets show that the module improved feature learning capabilities. A similar research is reported in \citep{ige2023wsense} proposed the WSense module to address the issue of differences in the quality of features learnt, regardless of the size of the sliding window segmentation, and experimented on PAMAP2 and WISDM datasets. The results show that by plugging the WSense module into Conv1D architectures, improved activity features can be learned from wearable sensor data for human activity recognition.

\begin{figure*}[h!]
    \centering
    \includegraphics[scale=0.5]{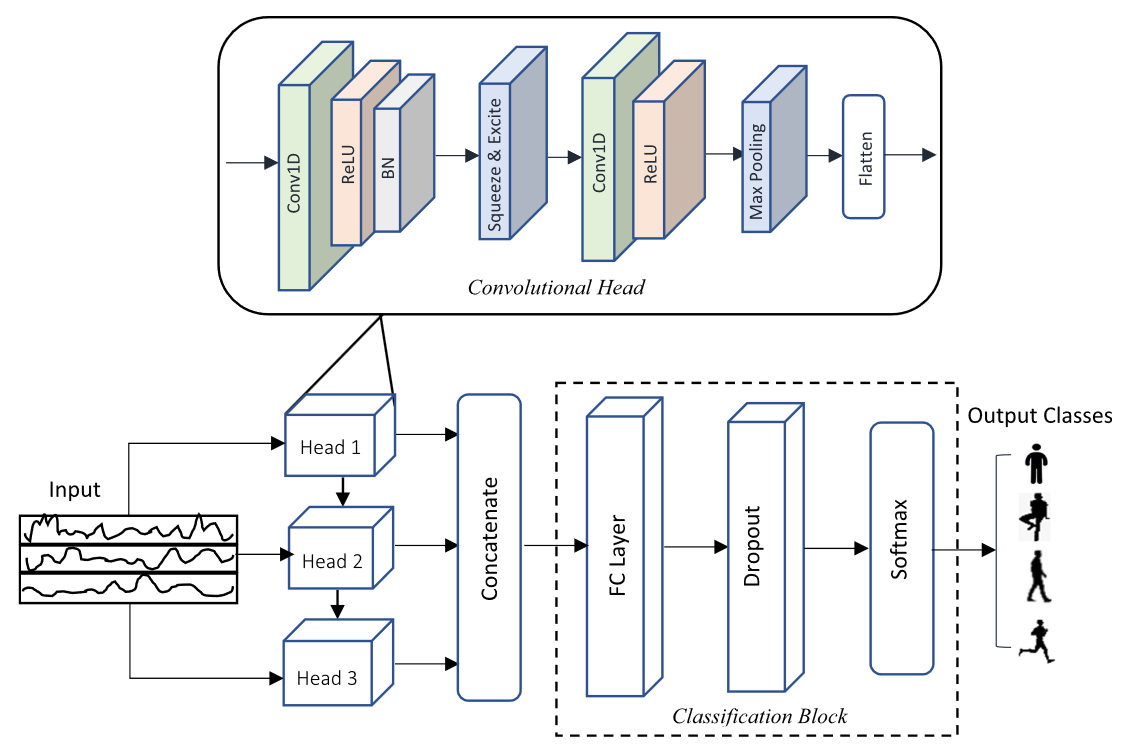}
    \caption{The architecture of multi-head CNN model \citep{khan_attention_2021}.}
    \label{fig_har_multi-head_cnn}
\end{figure*}

\textbf{Hybrid Models}: Researchers have also proposed the use of standalone RNNs in HAR, and a hybrid of Conv1D architectures with RNNs such as LSTMs \citep{deep_hybrid_2019}, BiLSTMs \citep{luwe_wearable_2022, shi_novel_2023}, GRUs \citep{dua_inception_2023} and BiGRUs \citep{imran_smart-wearable_2024} to fully harness the feature learning capabilities of both CNN and RNNs. For instance, Nafea et al. \citep{nafea_sensor-based_2021}, leveraged Bi-LSTM and Conv1D with increasing kernel sizes to learn features at various resolutions. Human activity features were extracted using the stacked convolutional layers with a Bi-LSTM layer, before including a flattening layer and a fully connected layer for subsequent classification. However, the model had issues extracting quality features of dynamic activities compared to static activities. To address such issues, some research works have incorporated attention mechanisms in Conv1D-based architectures to improve feature learning of dynamic and complex activities from time series signals obtained from wearable sensors. For example, Khan and Ahmad \citep{khan_attention_2021} designed three lightweight convolutional heads, with each specialized in feature extraction from wearable sensor data. Each head comprised stacked layers of Conv1Ds, along with embedded attention mechanisms to augment feature learning as shown in Figure \ref{fig_har_multi-head_cnn}. The results demonstrated that integrating multiple 1D-CNN heads with SE attention can enhance feature learning for Human Activity Recognition. 

Ayo and Noor \citep{ige_deep_2023} designed three feature learning pipelines, each pipeline consisting of two concurrent layers of Conv1D and LSTM with maximum pooling, which are then concatenated and processed using a channel-wise attention mechanism to enhance feature learning. In \citep{gao_danhar_2021}, a sequential channel-temporal attention is proposed for multi-modal activity recognition. The channel attention is similar to SE attention, but with average pooling and maximum pooling pipelines to squeeze to temporal dimension. The pooled features are then combined through addition operation. As for the temporal attention, average pooling and maximum pooling are applied along the channel dimension, producing two pooled features which are then combined through concatenation. The attention module is incorporated into the residual blocks of ResNet-like model to improve feature extraction. Similar work is reported in \citep{agac_resource-efficient_2024} whereby a hybrid sequential channel-spatial attention is proposed for a lightweight activity recognition model. The lightweight model consists of four convolutional blocks integrated with the attention module and two LSTM layers. Although the proposed attentions have shown a strong potential in improving model performance, their sequential processing inherently prioritize one of the feature map dimensions, impeding the model's ability to capture holistic feature representations. In \citep{tang_triple_2022}, a triple cross-domain attention is proposed to blend three attention branches to improve feature extraction in HAR. For each attention branch, z-pooling \citep{misra_rotate_2020} and convolution operations are applied along the corresponding dimension to generate the attention weights. Then, the feature maps are combined via average operation. The model consists of three residual blocks where each block is integrated with the attention module. These diverse modifications and adaptations showcase the versatility and potential of deep learning models in achieving state-of-the-art in HAR systems.

\textbf{Transformers}: Transformers have been employed in human activity recognition to capture long-range dependencies in sensor data, enabling more accurate activity classification. In \citep{chen_transformer_2022}, a transformer-based model is enhanced by integrating a bidirectional GRU and a linear fully connected layer into each encoder block. The decoder is reduced to a fully connected layer with softmax function. This hybrid model leverages the transformer's ability to capture long-range dependencies in sensor data and the bidirectional GRU's strength in modeling temporal sequences, enhancing the model performance in classifying complex and collaborative activities. Similar work is reported in \citep{sun_efficient_2024} whereby the encoder blocks of the transformer are integrated with 1D convolutional layers instead of fully connected layers, enabling the model to extract local features. Furthermore, self-supervised contrastive learning is employed to learn from unlabeled data before the model is fine-tuned using a smaller dataset. Although transformer-based models have shown remarkable performance, they suffer from high computational cost and memory requirements, making them less suitable for real-time or resource-constrained activity recognition. A study was conducted to determine the applicability of transformer models on a low-power ESP32 microcontroller \citep{lattanzi_are_2025}. The study concludes that transformer models are not suitable for tiny devices. Furthermore, the results show that a tiny transformer model with two encoder blocks achieved lower accuracy compared to the standard LSTM model, with a difference up to 14\%. However, transformer models run 3x faster than LSTM, making them a viable option provided the architecture of transformers is optimized for low-power platforms.

\textbf{Generative Models}: A reliable deep learning model requires a large amount of training data to learn the underlying patterns of the data. However, in HAR, data collection is expensive, and the available datasets are often limited in the number of samples. To overcome this limitation, researchers employ GAN models to generate synthetic data, thus augmenting the training set. In \citep{chan_unified_2021}, a conditional GAN is proposed to generate realistic sensor data of different activities for human activity classification. The generator consists of four 1D convolutional and maximum pooling blocks followed by two layers of LSTM and fully connected layers, while the discriminator is a standard CNN model. An enhanced conditional GAN is proposed to improve synthetic data generation for HAR \citep{jimale_fully_2022}. This architecture integrates 1D convolutional layers with multiple fully connected networks at the generator's input and discriminator's output, aiming to produce higher-quality synthetic samples compared to existing CGAN models. Similar work is reported in \citep{lupion_data_2024}, where the conditional Wasserstein GAN is employed to generate synthetic data for HAR. 

In \citep{kia_human_2024}, sensor data is transformed into its frequency spectrum, forming an RGB-based feature space for human activity classification. Then, BiGAN is employed to generate synthetic frequency spectrum images to increase minority sample classes, thus diversifying the training set. The generator and discriminator of the proposed BiGAN are built using 2D convolutional and fully connected layers. In \citep{mohammadzadeh_cgan-based_2025}, a conditional GAN is employed to generate synthetic sensor data for therapeutic activity recognition. The generator utilizes Inception-like modules and transposed convolutional layers, while the discriminator consists of two classification pipelines: one classifies raw signals and the other pipeline classifies the Fourier-transformed signal (frequency spectrum images). Both predictions are averaged to obtain the final prediction. Table \ref{table_summary_har_studies} lists the summary of state-of-the-art human activity recognition.

\begin{table*}[htbp]
\centering
\caption{Summary of state-of-the-art human activity recognition.}
\scriptsize
\begin{tabularx}{\textwidth}{p{1.5cm} X X}
 \hline
 Ref. & Description & Results (Datasets and Metrics) \\ 
 \hline
 \citep{dua_inception_2023}, 2023 & This study proposed an Inception inspired model, combining convolutional layers with GRU. The convolutional blocks use multiple-sized filters to extract multiscale feature representations & MHEALTH\newline Accuracy: 99.25\%\newline PAMAP2\newline Accuracy: 97.64\% \\ 
 \hline
 \citep{ige_deep_2023}, 2023 & This study proposed a model with three parallel feature learning pipelines, each pipeline has two sub-feature learning pipelines consisting of convolutional layers and bidirectional LSTM, and channel-wise attention before the classifier & PAMAP2\newline Accuracy: 98.52\%\newline WISDM\newline Accuracy: 97.90\% \\
 \hline
 \citep{agac_resource-efficient_2024}, 2024 & This study proposed A lightweight model consists four convolutional blocks integrated with a sequential channel-spatial module and two LSTM layers & PAMAP2\newline Accuracy: 98.52\%\newline WISDM\newline Accuracy: 97.90\% \\
 \hline
\citep{sun_efficient_2024}, 2024 & A modified transformer integrated with convolutional layer instead of fully connected layer & UCI-HAR\newline Accuracy: 95.49\%\newline Skoda\newline Accuracy: 87.88\%\newline Mhealth\newline Accuracy: 98.43\% \\
 \hline
 \citep{lupion_data_2024}, 2024 & This study employed a conditional Wasserstein GAN to generate synthetic human activity recognition accelerometry signals & Local Ulster University dataset\newline Accuracy: 0.7453 \\
 \hline
 \citep{kia_human_2024}, 2024 & This study converted sensor data into its frequency spectrum to create an RGB-based feature space, employed BiGAN to generate synthetic frequency spectrum images to balance minority classes, and used 2D convolutional and fully connected layers in the generator and discriminator & Up-Fall\newline Accuracy: 99.1\%\newline Opportunity\newline Accuracy: 86.8\%\newline WISDM\newline Accuracy: 99.12\% \\ 
 \hline
 \citep{mohammadzadeh_cgan-based_2025}, 2025 & This study  employed a conditional GAN to generate synthetic sensor data for therapeutic activity recognition which has Inception-like generator and a discriminator with two classification pipelines, one for raw signals and the other for Fourier-transformed signals & Local Sharif University of Technology dataset \newline Avg. F1-score: 0.897 \\ 
 \hline
\end{tabularx}
\label{table_summary_har_studies}
\end{table*}

\subsubsection{Speech Recognition}
Speech, as the primary mode of human communication, has captivated researchers for over five decades, especially since the inception of AI \citep{Nassif2019}. From the earliest endeavors to understand and replicate the complexities of human speech, to contemporary advancements leveraging cutting-edge technologies, the quest for accurate and efficient speech recognition systems has been relentless. In recent years, the emergence of deep learning techniques has revolutionized the speech recognition field. Deep learning has demonstrated unparalleled success in processing and extracting intricate patterns from vast amount of data. When applied to the realm of speech recognition, deep learning have surpassed traditional approaches by learning intricate features directly from raw audio signals, circumventing the need for handcrafted features and complex preprocessing pipelines. This paradigm shift has significantly advanced the state-of-the-art in speech recognition, enabling systems to achieve unprecedented levels of accuracy and robustness across various languages, accents, and environmental conditions. Generally, deep learning has been extended to other essential applications of speech recognition, such as speaker identification \citep{Tirumala, app11083603}, emotion recognition \citep{khalil2019}, language identification \citep{singh2021spoken}, accent recognition \citep{jiao2016accent}, age recognition \citep{sanchez2022age} and gender recognition \citep{alnuaim2022speaker}, among many others.

Prior to the adoption of deep learning in speech recognition, the foundation of traditional speech recognition systems was the use of Gaussian Mixture Models (GMMs), which are often combined with Hidden Markov Models (HMMs) to represent speech signals \citep{srivastava2022speech}. This is because a speech signal can be thought of as a short-term stationary signal. The spectral representation of the sound wave is modelled by each HMM using a mixture of Gaussian. However, they are considered statistically inefficient for modelling non-linear or near non-linear functions \citep{padmanabhan2015review,Nassif2019}. This is because HMMs rely on a set of predefined states and transition probabilities, making assumptions about the linearity and stationarity of the underlying data. While suitable for modelling certain aspects of speech, HMMs often fall short when tasked with representing the intricate nonlinearities and variability present in speech signals. Speech, by nature, exhibits nonlinear and dynamic characteristics, with features such as intonation, rhythm, and phonetic variations challenging the simplistic assumptions of traditional statistical models like HMMs. In other words, GMM-HMM approach had limitations in capturing complex acoustic patterns and long-term dependencies in speech \citep{mukhamadiyev2022automatic}.

\textbf{Sequence-to-Sequence Models}: In recent times, CNN and RNNs have been leveraged for automatic speech recognition in order to consider a longer or variable temporal window for context information extraction \citep{lu2020automatic}. Generally, CNNs are well-suited for capturing local patterns and hierarchical features in data, making them effective for modelling acoustic features in speech. By directly learning features from raw speech signals, CNNs bypassed the need for handcrafted features used in traditional GMM-HMM systems. Additionally, CNNs can capture long-range dependencies in the data, which is crucial for understanding the context of speech. Likewise, the RNNs are suitable choice for exploring extended temporal context information in one processing level for feature extraction and modelling. 

Based on this, several researchers have proposed the use of both CNN and variants of RNNs for automatic speech recognition and for other speech related tasks. For instance, \citep{HEMA2023109492}, used CNN to classify speech emotions and benchmarked on a dataset consisting of seven classes (anger, disgust, fear, happiness, neutral, sadness, and surprise). However, CNNs lack the ability to model temporal dependencies explicitly. In speech recognition, understanding the temporal context of speech is essential for accurate transcription. Furthermore, speech signals are inherently sequential, and information from previous time steps is crucial for understanding the current speech segment. CNNs, by design, do not inherently capture this sequential nature. For this reason, variants of RNNs have been leveraged to collect extended contexts in speeches. This is because RNNs are designed to model sequential data by maintaining hidden states that capture information from previous time steps. This allows them to capture temporal dependencies effectively, making them well-suited for ASR tasks. In \citep{shewalkar2019performance}, the authors evaluated the performance of RNN, LSTM, and GRU on a popular benchmark speech dataset (ED-LIUM). The results show that LSTM achieved the best word error rate, while the GRU optimization was faster and achieved word error rate close to that of LSTM.  

However, RNN architectures process input sequences sequentially, which limits their ability to capture global context information effectively. As a result, they may struggle to understand the entire context of a spoken utterance, leading to lower transcription accuracy, particularly in tasks requiring understanding beyond local dependencies. Furthermore, most CNN and RNN automatic speech recognition systems consist of separate acoustic, pronunciation, and language modelling components that are trained independently. Usually, the acoustic model bootstraps from an existing model that is used for alignment to train it to recognize context dependent (CD) states or phonemes. The pronunciation model, curated by expert linguists, maps the sequences of phonemes produced by the acoustic model into word sequences. For this reason,  Sequence-to-Sequence models are being proposed in automatic speech recognition to train the acoustic, pronunciation, and language modelling components jointly in a single system \citep{prabhavalkar2017comparison}. Sequence-to-Sequence models in automatic speech recognition are a class of models that aim to directly transcribe an input sequence of acoustic features such as speech spectrograms or Mel-frequency cepstral coefficients into a sequence of characters or words representing the recognized speech. There have been various sequence-to-sequence models explored in the literature, including Recurrent Neural Network Transducer (RNN-T) \citep{graves2012sequence}, Listen, Attend and Spell (LAS) \citep{chan2015listen}, Neural Transducer \citep{jaitly2016online}, Monotonic Alignments \citep{raffel2017online} and Recurrent Neural Aligner (RNA) \citep{sak2017recurrent}.

\begin{figure}[h!]
    \centering
    \includegraphics[scale=0.45]{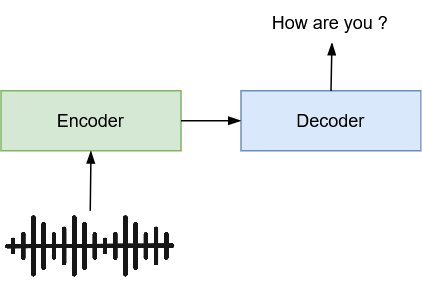}
    \caption{Sequence-to-Sequence}
    \label{fig_speech_seq2seq}
\end{figure}

As shown in Figure \ref{fig_speech_seq2seq},  the encoder component takes the input sequence of acoustic features and processes it to create a fixed-dimensional representation, often called the context vector. This representation captures the essential information from the input sequence and serves as the basis for generating the output sequence. The decoder component takes the context vector produced by the encoder and generates the output sequence. In ASR, this output sequence consists of characters or words representing the recognized speech. The decoder is typically implemented as a recurrent neural network (RNN), such as a Long Short-Term Memory (LSTM) or Gated Recurrent Unit (GRU) network, or it could be a transformer-based architecture. During training, the model learns to map input sequences to their corresponding output sequences by minimizing a suitable loss function, such as cross-entropy loss. This is typically done using techniques like backpropagation through time (BPTT) or teacher forcing, where the model is trained to predict the next token in the output sequence given the previous tokens. Thereafter, the trained model is used to transcribe unseen speech input. The encoder processes the input sequence to produce the context vector, which is then fed into the decoder to generate the output sequence. In some cases, beam search \citep{szucbeamsearch2019, li2018seq2seq} or other decoding strategies may be used to improve the quality of the generated output.

In \citep{chiugoogle}, the authors explored various structural and optimization enhancements to their LAS Sequence to Sequence model, resulting in significant performance improvements. They introduce several structural enhancements, including the utilization of word piece models instead of graphemes and the incorporation of a multi-head attention architecture, which outperforms the commonly used single-head attention mechanism. Additionally, they investigate optimization techniques such as synchronous training, scheduled sampling, label smoothing, and minimum word error rate optimization, all of which demonstrate improvements in accuracy. The authors presented experimental results utilizing a unidirectional LSTM encoder for streaming recognition. On a 12,500-hour voice search task, they observe a decrease in Word Error Rate (WER) from 9.2\% to 5.6\% with the proposed changes, while the best-performing conventional system achieves a WER of 6.7\%. Moreover, on a dictation task, their model achieves a WER of 4.1\%, compared to 5\% for the conventional system. Similarly, the work of Prabhavalkar et al. \citep{prabhavalkar2017comparison} investigated a number of sequence-to-sequence methods in automatic speech recognition. These included the RNN transducer (RNN-T), attention-based models, a new model that augments the RNN-T with attention, and a Connectionist Temporal Classification (CTC) trained system that directly outputs grapheme sequences. According to their research, sequence-to-sequence approaches can compete on dictation test sets against state-of-the-art when trained on a large volume of training data. 

\textbf{Transformers}: Transformers have become the basis of state-of-the-art models in speech recognition. Wav2Vec 2.0 reads raw audio signals using multilayer convolutional encoder to generate latent speech representations \citep{baevski_wav2vec_2020}. The feature representation is subsequently fed to a transformer to capture the contextual information in the data. The transformer uses a convolutional layer for the positional encoding. Furthermore, the authors introduced vector quantization technique to convert the feature encoder output to discrete codes which are then used for contrastive learning during pretraining. Whisper is a robust speech recognition model proposed by OpenAI, trained on 680,000 hours of labeled audio data \citep{radford_robust_2022}. Of these, 117,000 hours cover 96 non-English languages and 125,000 hours consist of non-English to English translation data. The model is based on encoder-decoder transformer architecture, where the input data is first processed using convolutional layers followed by the positional encoding. Then, the input is fed into the encoder, which consists of a series of transformer blocks, each including multi-head self-attention mechanisms. The encoder's output is then fed into each decode transformer block to generate the final output. The decoder's transformer blocks consist of multi-head self-attention and cross attention mechanisms.Conformer is a Transformer-based encoder, combining self-attention with convolutional layers to better capture both global and local features in speech signals \cite{gulati_conformer_2020}. The encoder provides a balance approach leveraging the strengths of convolution and self-attention mechanism. Conformer is often used with CTC or RNN-T, or other decoding mechanisms.

Auto-regressive decoding in transformer-based models is computationally expensive, as it requires repeated processing of the complete encoded speech, resulting in slow operation. In \citep{stooke_aligner-encoders_2025}, Aligner-Encoder, a transformer-based encoder is proposed that perform internal alignment between audio and text, eliminating the need for complex decoding steps and complex dynamic programming during training. The approach simplifies the model architecture using a lightweight decoder without cross attention that processes embedding frames sequentially until an end-of-message token is generated. In \citep{zhang_breaking_2025}, the Spike Window Decoding (SWD) algorithm is introduced that leverages the spike property of CTC outputs where each spike represents a strong signal indicating the presence of a specific token. By focusing on these spikes within a fixed window, SWD reduces the complexity of decoding, enabling faster and more efficient processing. Although transformer-based models are highly effective, they often capture highly entangled feature representations, leading to a lack of clear interpretability. In \citep{wang_disentangled-transformer_2024}, the authors proposed the Disentangled-Transformer to enhance the interpretability of transformer-based models. The transformer disentangles the internal representations into sub-embeddings based on the various temporal characteristics of the speech signals. To this end, the loss function is improved by introducing time-invariant regularization terms for each time frame. A state-of-the-art speech recognition model called Samba-ASR is proposed utilizing Mamba architecture that is based on state-space-models \citep{shakhadri_samba-asr_2025}. This approach overcomes the limitations of transformer-based models such as inability to handle long-range dependencies and their quadratic scaling with input length, by efficiently capturing both local and global temporal dependencies through state-space-dynamics. Table \ref{table_summary_speech_studies} lists the summary of state-of-the-art speech recognition applications.

\begin{table*}[htbp]
\centering
\caption{Summary of state-of-the-art speech recognition.}
\scriptsize
\begin{tabularx}{\textwidth}{p{1.5cm} X X}
 \hline
 Ref. & Description & Results (Datasets and Metrics) \\ 
 \hline
 \citep{radford_robust_2022}, 2022 & This study proposed Whisper an automatic speech recognition (ASR) system that transcribes and translates spoken language using a robust deep learning model trained on diverse multilingual and multitask datasets & Various datasets (first three datasets only)\newline LibriSpeech Clean\newline WER: 2.7 \newline Artie \newline WER: 6.2\newline Common Voice\newline WER: 9.0 \\ 
 \hline
 \citep{stooke_aligner-encoders_2025}, 2025 & This study introduced Aligner-Encoder, a transformer-based automatic speech recognition model that internally aligns audio-text representations before decoding, enabling a simpler and more efficient architecture trained with frame-wise cross-entropy loss & LibriSpeech-960H\newline WER: 2.2 \newline Voice Search\newline WER: 3.7\newline YouTube videos\newline WER: 7.3 \\
 \hline
\citep{zhang_breaking_2025}, 2025 & An algorithm that leverages spike property of CTC outputs to reduce the complexity of WFST decoding process & AISHELL-1\newline CER: 3.89\newline In House\newline CER: 2.09 \\
 \hline
 \citep{wang_disentangled-transformer_2024}, 2024 & A transformer-based model architecture designed to disentangle internal representations and enhance model explainability. & LibriSpeech-100H\newline DER: 8.1\newline LibriMix 1\newline DER: 5.7\newline LibriMix 2\newline DER: 6.9\newline LibriMix 3\newline DER: 2.5\newline LibriMix 4\newline DER: 5.6 \\
 \hline
 \citep{shakhadri_samba-asr_2025}, 2025 & A speech recognition model based on state-space-models & GigaSpeech\newline WER: 9.12\newline LibriSpeech Clean\newline WER: 1.17\newline LS Other\newline WER: 2.48\newline SPGISpeech\newline WER: 1.84 \\ 
 \hline
\end{tabularx}
\label{table_summary_speech_studies}
\end{table*}

\subsubsection{Finance}
Over the past few decades, computational intelligence in finance has been a hot issue in both academia and the financial sector \citep{ozbayoglu2020deep}. Deep learning, especially RNN models have gained significant traction in the field of finance due to its ability to handle sequential data, since financial data often exhibit sequential dependencies, such as time series data for stock prices or historical transaction data. Within the financial industry, researchers have developed deep learning models for stock market forecasting \citep{singh2017stock}, algorithmic trading \citep{lei2020deep}, credit risk assessment \citep{shen2021new}, portfolio allocation \citep{wang2020portfolio}, asset pricing \citep{chen2024deep}, and derivatives markets \citep{ahnouch2023model}, among others and these models are intended to offer real-time operational solutions. In exchange rate prediction, 
Sun et al. \citep{SUN2020101160} developed an ensemble deep learning technique known as LSTM-B by combining a bagging ensemble learning algorithm with a long-short term memory (LSTM) neural network to increase the profitability of exchange rate trading and produce accurate exchange rate forecasting results. In comparison to previous methodologies, the authors' estimates proved to be more accurate when they looked at the potential financial profitability of exchange rates between the US dollar (USD) and four other major currencies: GBP, JPY, EUR, and CNY. 

The authors in \citep{abedin2021deep} proposed a Bi-LSTM-BR technique, which combined Bagging Ridge (BR) regression with Bi-LSTM as base regressors. The pre-COVID-19 and COVID-19 exchange rates of 21 currencies against the USD were predicted using the Bi-LSTM BR, and experiments showed that the proposed method outperformed ML algorithms such as DT and SVM. However, exchange rate data can be noisy and subject to non-stationarity, which can pose challenges for predictive modelling. While bagging techniques can help mitigate the effects of noise to some extent, they may struggle to capture long-term trends or sudden shifts in the data distribution, leading to suboptimal performance. To address this, Wang et al. \citep{wang_sudf-rs_2023} presented an approach for one-day ahead of time exchange rate prediction that concurrently considers both supervised and unsupervised deep representation features to enhance Random Subspace. Two crucial phases in the SUDF-RS technique are feature extraction and model building. First, LSTM and deep belief networks, respectively, extract the supervised and unsupervised deep representation features. To produce high-quality feature subsets, an enhanced random subspace approach was created that integrates a random forest-based feature weighting mechanism. Then, the matching base learner is trained using each feature subset, and the final outputs are generated by averaging the outcomes of each base learner. Experiments on EUR/USD, GBP/USD and USD/JPY showed that improved accuracy was achieved using the model.

In stock market prediction, several deep learning architectures have been proposed in the literature. For instance, \citep{nikou2019stock}, conducted a comparative study between the ANN, SVR, RF and an LSTM model. As compared to the other models discussed in the study, the LSTM model outperformed the others in predicting the closing prices of iShares MSCI United Kingdom. Similarly, using stock market historical data and financial news, Cai et al. \citep{cai2018financial} used CNN and LSTM forecasting methods to generate seven prediction models. The seven models were then combined into a single ensemble model in accordance with the ensemble learning approach to create an aggregated model. However, the accuracy of all the models' predictions was low. Gudelek et al. \citep{gudelek2017deep} proposed a CNN model which used a sliding window technique and created pictures by capturing daily snapshots within the window's bounds. With 72\% accuracy, the model was able to forecast the prices for the following day and was able to generate 5 times the starting capital. In Eapen et al. \citep{eapen2019novel}, a CNN and Bi-LSTM model with numerous pipelines was proposed, utilising an SVM regressor model on the S\&P 500 Grand Challenge dataset, and results showed enhanced prediction performance by over a factor of 6\% compared to baseline models. As presented, deep learning has undeniably achieved state-of-the-art performance across various domains within finance. In \citep{chen_deep_2024}, a multi-modal deep learning model is proposed for stock market trend prediction by integrating daily stock prices, technical indicators and sentiment from daily news headlines. The model architecture consists of BERT-based branch fine-tuned on financial news to capture textual sentiment and an LSTM branch that captures temporal patterns in the time series data, including stock prices and technical indicators. Both feature representations are combined through concatenation and then passed to a fully connected layer for predictions.

A study was conducted to analyze the performance of Prophet, LSTM and Transformer. Prophet is a parametric, additive regression model based on time series decomposition \citep{mozaffari_predictive_2024}. The LSTM model consists of two hidden layers, with a hidden size of 64 and a linear fully connected layer is used to generate the output. The Transformer model follows the original transformer architecture. The results show that transformer outperformed LSTM and Prophet in terms of forecasting accuracy, particularly for datasets with complex temporal dependencies. However, transformer models have high memory usage and quadratic complexity due to self-attention mechanism, which makes them inefficient for very long sequence time-series forecasting. To overcome the limitations, a StockFormer is proposed which is based on Informer \citep{li_transformer_2025}. Informer is a transformer that was designed for time-series forecasting, utilizing the ProbSparse self-attention mechanism which attends to only the most important queries for each key, reducing the computational complexity \citep{zhou_informer_2021}. Furthermore, self-attention distilling technique is introduced to reduce redundancy and highlighting crucial information in the feature maps, helping the model to focus more on the relevant parts of the input sequence. In \citep{berti_tlob_2025}, a transformer-based model is proposed for stock trend prediction using limit order book data (LOB). The model consists of a series of transformer LOB (TLOB) blocks which leverage dual attention mechanisms to capture both spatial and temporal dependencies, allowing the model to adaptively focus on the market microstructure of the input data. Furthermore, a multilayer of perceptron LOB (MPLOB) is introduced, comprising two blocks of two fully connected layers which operate along the feature and temporal dimensions. Table \ref{table_summary_finance_studies} lists the summary of state-of-the-art finance applications.

\begin{table*}[htbp]
\centering
\caption{Summary of state-of-the-art finance applications.}
\scriptsize
\begin{tabularx}{\textwidth}{p{1.5cm} X X}
 \hline
 Ref. & Description & Results (Datasets and Metrics) \\ 
 \hline
 \citep{wang_sudf-rs_2023}, 2023 & This study introduces SUDF-RS technique for one-day-ahead exchange rate prediction by combining supervised and unsupervised deep representation features with an enhanced random subspace approach, leveraging LSTM and deep belief networks for feature extraction and using a random forest-based weighting mechanism to improve prediction accuracy & Data collected from investing.com\newline RMSE: 0.448 (EUR/USD), 0.676 (GBP/USD), 0.374 (USD/JPY) \\ 
 \hline
 \citep{chen_deep_2024}, 2024 & A multi-modal deep learning model integrates stock prices, technical indicators, and news sentiment using a BERT-based branch for textual analysis and an LSTM branch for temporal patterns, combining their features for stock market trend prediction & Data collected using Yahoo Finance and  EODHD API\newline F1-Score: 0.45 (ATVI), 0.51 (AAPL), 0.48 (AMT), 0.53 (PLD), 0.50 (NDAQ), 0.45 (SCHW), 0.43 (BIO), 0.47 (JNJ) \\
 \hline
\citep{mozaffari_predictive_2024}, 2024 & This study analyzes stock closing price prediction using LSTM, Prophet, and Transformer models & American Airlines Group Inc.\newline MSE: 0.0085 (Transformer), 0.1972 (LSTM), 16.9321 (Prophet) \newline Atlantic American-Life Insurance\newline MSE: 0.0118 (Transformer), 0.0095 (LSTM), 17.8601 (Prophet) \\
 \hline
 \citep{li_transformer_2025}, 2025 & The study proposed a model based on Informer which utilizes ProbSparse self-attention and self-attention distilling to reduce the computation complexity & Combined data from Yahoo, alphavantage.co, alpaca.markets and polygon.io \newline Percent profit: 1.7550 \\
 \hline
 \citep{berti_tlob_2025}, 2025 & The study proposed a transformer-based model with dual attention mechanisms and multilayer of perceptrons for feature mixing along feature and temporal dimensions & FI-2010\newline F1-score: 92.81\newline TSLA \newline F1-score: 60.50 \newline INTC \newline F1-score: 80.15 \\ 
 \hline
\end{tabularx}
\label{table_summary_finance_studies}
\end{table*}

\subsubsection{Electrocardiogram (ECG) Classification}
Disorders pertaining to the heart or blood vessels are collectively referred to as Cardiovascular Diseases (CVD) \citep{LIU2021107187}. According to the American Heart Association's 2023 statistics, CVD has emerged as the leading cause of death worldwide. In 2020, 19.05 million deaths were recorded from CVD globally, which signifies an increase of 18.71\% from 2010, and it is believed that this number will rise to 23.6 million by 2030 \citep{tsao2023heart}. Blood clots and vascular blockages caused by CVDs can cause myocardial infarction, stroke, or even death \citep{LIU2021107187}. Generally, early diagnosis has been shown to reduce the mortality rate of CVDs, and Electrocardiogram (ECG) signals play a crucial role in diagnosing various cardiac abnormalities and monitoring heart health. However, ECG signal has characteristics of high noise and high complexity, making it time-consuming and labor-intensive to identify certain diseases using traditional methods. The traditional approach is tedious and requires the expertise of a medical specialist. Over the past decades, the task of Long-term ECG recording classification has been significantly facilitated for cardiologists through the adoption of computerized ECG recognition practices. Throughout this period, feature extraction methods have predominantly relied on manual techniques, encompassing diverse approaches such as wave shape functions \citep{LLamedo2011}, wavelet-based features \citep{MATHEWS201853}, ECG morphology \citep{zhu2019}, hermite polynomials \citep{desai2021low}, and Karhunen-Loeve expansion of ECG morphology \citep{crippa2015multi}, among others. These extracted features are subsequently subjected to classification using various machine learning algorithms. 

More recently, the advent of deep learning has revolutionized the field by enabling automatic feature learning directly from ECG signals. This advancement holds significant promise in the realm of automated ECG classification, offering clinicians a tool for swift and accurate diagnosis. Based on this, several deep learning architectures have been proposed for feature learning of ECG signals. For instance, Acharya et al. \citep{acharya2017deep} developed a 9-layer CNN model to automatically identify five categories of heartbeats in ECG signals. A similar model was also developed in \citep{baloglu2019classification}. However, ECG signals often vary significantly in length, as they may contain different numbers of heartbeats. CNNs typically require fixed-length inputs, which may necessitate preprocessing steps such as padding or truncation, potentially losing important temporal information. For this reason, several architectures have leveraged RNN in ECG classification, as seen in \citep{singh2018classification}, \citep{prabhakararao2020attentive} and \citep{wang2023single}, among others. While RNNs are capable of handling sequential data, they also have limitations in capturing local patterns or short-term dependencies effectively. In ECG signals, local features such as specific waveforms or intervals can be crucial for classification. For this reason, recent works have proposed hybrid models which combine the strengths of both CNNs and RNNs to overcome some of these limitations \citep{sowmya2022contemplate}.

\textbf{Hybrid Models}: The work of Rai et al. \citep{rai2022hybrid} developed a hybrid CNN-LSTM network to evaluate the optimum performing model for myocardial infarction detection using ECG signals. The authors then experimented on 123,998 ECG beats obtained from the PTB diagnostic database (PTBDB) and MIT-BIH arrhythmia database (MITDB), and the result showed that by combining the capabilities of both CNN and LSTM, improved classification accuracy can be achieved. Also, in \citep{Banerjee2020}, a CNN architecture was developed to extract morphological features from ECG signals. For the purpose of determining the degree of heart rate variability, another composite structure was designed using LSTM and a collection of manually created statistical features. Following that, a hybrid CNN-LSTM architecture is built using the two independent biomarkers to classify cardiovascular artery diseases, and experiments were carried out on two distinct datasets. The first is a partly noisy in-house dataset collected using an inexpensive ECG sensor, and the other is a corpus taken from the MIMIC II waveform dataset. The hybrid model proposed in the work achieved an overall classification accuracy of 88\% and 93\%, respectively, which surpasses the performance of standalone architectures. 

An automated diagnosis method based on Deep CNN and LSTM architecture was presented in \citep{kusuma_ecg_2022} to identify Congestive Heart Failure (CHF) from ECG signals. Specifically, CNN was used to extract deep features, and LSTM was employed to exploit the extracted features to achieve the CHF detection goal. The model was tested using real-time ECG signal datasets, and the results show that the AUC was 99.9\%, the sensitivity was 99.31\%, the specificity was 99.28\%, the F-Score was 98.94\%, and the accuracy was 99.52\%. Similar work is reported in \citep{alamatsaz_lightweight_2024} whereby CNN is combined with LSTM for ECG classification. The hybrid model consists of three blocks of convolutional and maximum pooling layers with dropout followed by an LSTM with drop. Furthermore, Shapley is utilized to determine the contribution of each ECG sample on the prediction to improve model interpretability. However, since ECG signals can vary in length due to differences in recording durations or patient conditions. LSTMs are capable of handling variable-length sequences, but traditional CNNs typically require fixed-length inputs. Therefore, fusing these features effectively in a hybrid model can be challenging. Furthermore, Hybrid CNN-RNN models can be computationally intensive, especially when processing long ECG sequences or large datasets. For this reason, recent research works have proposed the use of attention mechanisms to reduce the computational burden by enabling the model to selectively attend to informative features, focusing computational resources where they are most needed. Likewise, attention mechanisms can enable the model to attend to informative segments of the ECG signal, regardless of their length, allowing for more flexible processing of variable-length sequences. 

Several researchers have leveraged attention mechanisms in standalone and hybrid architectures for improved performance. For instance, in the work of Chun-Yen et al. \citep{chen_automated_2022}, CNN layers were used to extract main features, while LSTM and attention were included to enhance the model's feature learning capabilities. Experiments on a 12-lead KMUH ECG dataset showed that the model had high recognition rates in classifying normal and abnormal ECG signals, compared to hybrid models without attention mechanisms. Wang et al. \citep{wang_automated_2021} presented a 33-layer CNN architecture with non-local convolutional block attention module (NCBAM). To extract the spatial and channel information, preprocessed ECG signals were first fed into the CNN architecture. A non-local attention further captured long-range dependencies of representative features along spatial and channel axes. Similarly, a spatio-temporal attention-based convolutional recurrent neural network (STA-CRNN) was presented in \citep{zhang_ecg-based_2020} with the aim of concentrating on representative features in both the spatial and temporal dimensions. The CNN subnetwork, spatiotemporal attention modules, and RNN subnetwork formed the STA-CRNN and according to findings, the STA-CRNN model was able to classify eight different forms of arrhythmias and normal rhythm with an average F1 score of 0.835. In \citep{sun_arrhythmia_2024}, the SE module is utilized in between convolutional layers and LSTM. This approach allows the model to focus on relevant channels before temporal dependencies are captured by the LSTM, improving the classification performance.

In \citep{huang_ecg_2024}, a guided spatial attention mechanism is introduced to incorporate domain knowledge into the model, enhancing ECG classification performance. The guided spatial attention mechanism has an autoencoder-like structure, where the feature maps are downsampled by the encoder and upsampled by the decoder. Additionally, skip connections are employed to improve information flow between the encoder and decoder. Furthermore, an attention loss term, based on the attention weights is introduced to jointly train the guided attention mechanisms. In \citep{aghaomidi_ecg-sleepnet_2024}, ECG signals are classified into sleep stages using deep learning. The approach is divided into three stages. In the first stage, a deep learning model consisting of convolutional layers and liquid time-constant network \citep{hasani_liquid_2021} is employed to compute kurtosis and skewness of the ECG signals. In the second stage, a deep learning model with convolutional layers and SE attention is used to enhance the prediction of the minority sleep stage class. In the final stage, the outputs of the models are concatenated and classified into sleep stage categories. Combining hybrid deep learning models with attention mechanisms for ECG feature learning is a promising approach that has already shown potential in ECG feature learning, according to reviewed literature. Future research can further explore semi-supervised and self-supervised learning techniques to leverage large amounts of unlabeled ECG data. This could involve pre-training models on large-scale unlabeled datasets using self-supervised learning objectives. 

\begin{figure*}[h]
    \centering
    \includegraphics[scale=0.4]{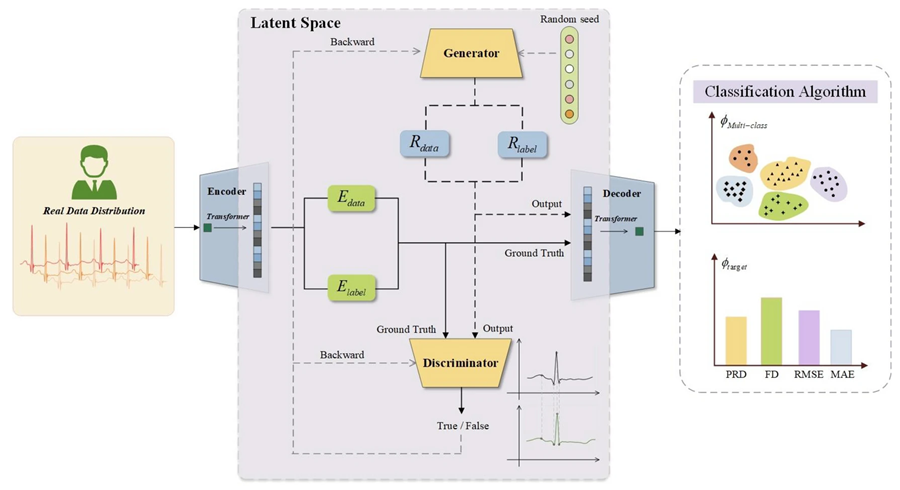}
    \caption{The architecture of CECG-GAN \citep{yang_data_2024}.}
    \label{fig_ecg_gan_ae}
\end{figure*}

\textbf{Generative Models}: Deep learning models have been leveraged in the generation of synthetic ECG signals to augment real signals, as seen in \citep{zhu_electrocardiogram_2019} where a GAN model was developed to generate ECG signals that correspond with available clinical data. The GAN model used two layers of BiLSTM  for the generator and CNN for the discriminator, and trained using the 48 ECG recordings of different users from the MIT-BIH dataset. The authors then compared their model with a Recurrent neural network autoencoder (RNN-AE) model and a recurrent neural network variational autoencoder (RNN-VAE) model, and the results show that their model exhibited the fastest convergence of its loss function to zero. Similar work is reported in \citep{yang_data_2024} in which GAN and autoencoder are employed to generate ECG signals to address the issue of imbalanced dataset as shown in Figure \ref{fig_ecg_gan_ae}. The generator and encoder create synthetic and real low-dimensional ECG samples, respectively. The discriminator classifies the synthetic and real low-dimensional ECG samples as real or fake, while the decoder reconstructs the ECG signals from the synthetic low-dimensional samples. This approach further refines data transformation and reconstruction quality, generating more realistic ECG signals. In \citep{msigwa_iot-driven_2024}, cluster-based GAN is employed to improve the diversity and quality of the generated synthetic ECG signals. $K$-means algorithm is first employed to partition the ECG signals into $k$ distinct groups. Then, the GAN is trained using the cluster centroids as initial reference points to guide the GAN in generating synthetic signals. Table \ref{table_summary_ecg_studies} lists the summary of state-of-the-art ECG classification.

\begin{table*}[htbp]
\centering
\caption{Summary of state-of-the-art ECG classification.}
\scriptsize
\begin{tabularx}{\textwidth}{p{1.5cm} X X}
 \hline
 Ref. & Description & Results (Datasets and Metrics) \\ 
 \hline
 \citep{sun_arrhythmia_2024}, 2024 & A hybrid model consists of three blocks of convolutional layer, SE module and an LSTM & MIT-BIH ECG\newline Accuracy: 98.5\% \\ 
 \hline
 \citep{alamatsaz_lightweight_2024}, 2024 & A lightweight hybrid model consists of three blocks of convolutional and maximum pooling layers and an LSTM & MIT-BIH and LTAF\newline Accuracy: 98.24\% \\
 \hline
\citep{chen_automated_2022}, 2022 & A hybrid model consists of three convolutional pipelines with different kernel size, two LSTM layers and an attention module & KMUH\newline Accuracy: 96.02\% \newline CPSC-2018\newline Accuracy: 94.05\% \\
 \hline
 \citep{huang_ecg_2024}, 2024 & This study proposed a guided spatial attention mechanism to enhance ECG classification performance & CPSC2018 and PTB-XL Chapman Cinc2017\newline F1-score: 77.56\% (STC), 88.53\% (PC), 82.54\% (WPW), 85.37\% (AF) \\
 \hline
 \citep{aghaomidi_ecg-sleepnet_2024}, 2024 & A novel three-stage approach for sleep stage classification extracts statistical features using a Feature Imitating Network, enhances minority sleep stage detection, and integrates outputs for five-class classification & MIT-BIH Polysomnographic\newline Accuracy: 80.79\% \\ 
 \hline
 \citep{yang_data_2024}, 2024 & This study proposed a generative model by combining conditional GAN and autoencoder & CSPC2020 dataset\newline F1-score: 0.9823 (Normal), 0.9824 (Premature ventricular), 0.9825 (Premature Supraventricular) \\ 
 \hline
 \citep{msigwa_iot-driven_2024}, 2024 & This study introduces a generative model by combining GAN and K-means & MIT-BIH arrhythmia\newline Accuracy: 99.66\%  \\ 
 \hline
\end{tabularx}
\label{table_summary_ecg_studies}
\end{table*}

\subsubsection{Electroencephalography (EEG) Classification}
Three-dimensional scalp surface electrode readings provide a dynamic time series that is called Electroencephalogram (EEG) signal \citep{Schirrmeister2017}. Brain waves obtained from an  EEG can effectively depict both the psychological and pathological states of a human. The human brain is acknowledged to be a fascinating and incredibly complicated structure. Numerous brain signals, including functional magnetic resonance imaging (fMRI), near-infrared spectroscopy (NIRS), electroencephalograms (EEGs), and functional near-infrared spectroscopy (fNIR), among others have been collected and used to study the brain \citep{gao2021complex}. Due to the EEG's non-invasive, affordable, accessible, and excellent temporal resolution characteristics, it has become the most utilized approach. However, the signal-to-noise ratio of EEG signal is low, meaning that sources with no task-relevant information frequently have a stronger effect on the EEG signal than those that do. These characteristics often make end-to-end feature learning for EEG data substantially more challenging \citep{Schirrmeister2017}. Based on this, several methods have been leveraged for improved feature extraction in EEG signals across several domains including Motor imagery \citep{Ang7802578}, anxiety disorder \citep{shen2022aberrated}, epileptic seizure detection \citep{boonyaki2020}, sleep pattern analysis and disorder detection \citep{sharma2021automated, vaquerizo2023explainable}, and Alzheimer's disease detection \citep{modir2023systematic}, and many others.

EEG Motor Imagery (MI) is a technique used to study brain activity associated with the imagination of movement. It involves recording electrical activity generated by the brain through electrodes placed on the scalp. MI tasks typically involve imagining performing a specific motor action, such as moving a hand or foot, without physically executing the movement, and has been leveraged in smart healthcare applications such as post-stroke rehabilitation and mobile assistive robots, among others \citep{altaheri2023deep}. Prior to the advent of deep learning, motor imagery EEG data are passed through various steps before classification using traditional ML techniques. Pre-processing, feature extraction, and classification are the three primary stages that traditional approaches usually take while processing MI-EEG signals. Pre-processing includes a number of operations, including signal filtering (choosing the most valuable frequency range for MI tasks), channel selection (identifying the most valuable EEG channels for MI tasks), signal normalization (normalizing each EEG channel around the time axis), and artefact removal (removing noise from MI-EEG signals). Independent component analysis (ICA) is the most often utilized technique for removing artefacts \citep{brunner2007spatial, delorme2007enhanced, jafarifarmand2019eeg}. In contrast to the traditional approach, deep learning architectures can automatically extract complex features from raw MI-EEG data without the need for laborious feature extraction and pre-processing. Based on this, several deep learning architectures have been proposed for MI-EEG feature learning, as seen in \citep{zhang2019novel}, \citep{kumar2016deep} and \citep{tibrewal2022}, among others. For instance \citep{Schirrmeister2017}, categorized MI-EEG signals using three CNNs with varying architectures, and the number of convolutional layers varied from two layers to a five-layer deep ConvNet to a thirty-one-layer residual network.

In Dai et al. \citep{dai2019eeg}, the authors proposed an approach for classifying MI-EEG signals which blend variational autoencoder with CNN architecture. The VAE decoder was used to fit the Gaussian distribution of EEG signals, and the time, frequency, and channel information from the EEG signal were combined to create a novel representation of input, and the proposed CNN-VAE method was optimized for the input.  Experiments showed that by combining both deep learning architectures, improved features were learnt, which led to a high classification performance on the BCI Competition IV dataset 2b. Li et al. \citep{LiMingaiLSTM} employed optimal wavelet packet transform (OWPT) for the generation of feature vectors from MI-EEG signals. These vectors were then utilized to train an LSTM network, which demonstrated satisfactory performance on dataset III from the BCI Competition 2003. However, the model has an excessively intricate structure. To address this, Feng et al. \citep{Feng2020novel} introduced a technique that merges continuous wavelet transform (CWT) with a simplified convolutional neural network to enhance the accuracy of recognizing MI-EEG signals. By employing CWT, MI-EEG signals were converted into time-frequency image representations. Subsequently, these image representations were fed into the SCNN for feature extraction and classification. Experiments on the BCI Competition IV Dataset 2b demonstrate that, on average, the classification accuracy across nine subjects reached 83.2\%. However, the computational complexity of the model was quite high, due to the processing of time-frequency image representations. The conversion of MI-EEG signals into time-frequency images using CWT requires significant computational resources. 

The authors in \citep{hwang_improving_2023} introduced a classification framework based on Long Short-Term Memory (LSTM) to improve the accuracy of classifying four-class motor imagery signals from EEG. The authors used sliding window technique to capture time-varying EEG signal data, and employed an overlapping-band-based Filter Bank Common Spatial Patterns (FBCSP) method to extract subject-specific spatial features. Experiments on the BCI Competition IV dataset 2a, showed that their model achieved an average accuracy of 97\%, compared to existing methods. Also, in the classification of Alzheimer's disease, Zhao et al. \citep{zhao_deep_2015} employed a deep learning network to analyze EEG data. The deep learning model was evaluated on a dataset that consisted of fifteen (15) patients with clinically confirmed Alzheimer's disease and fifteen (15) healthy individuals, and results showed that improved features were learnt and compared the results to the traditional methods. This has prompted the use of deep learning in Alzheimer's disease detection, as seen in \citep{xia_novel_2023}, where the authors used CNN for diagnosing Alzheimer's Disease. To address challenges posed by limited data and overfitting in deep learning models designed for Alzheimer's Disease detection, the authors explored the use of overlapping sliding windows to augment the EEG data collected from 100 subjects (comprising 49 AD patients, 37 mild cognitive impairment patients, and 14 healthy controls subjects). After assembling the augmented dataset, a modified Deep Pyramid Convolutional Neural Network (DPCNN) was used to classify the enhanced EEG signals. 

\textbf{Hybrid Models}: In \citep{hermawan_multi_2024}, the authors developed three deep learning architectures (CNN, LSTM, and hybrid CNN-LSTM), with each model chosen for its effectiveness in handling the intricate characteristics of EEG data for epilepsy detection. Each architecture offers distinct advantages, with CNN excelling in spatial feature extraction, LSTM in capturing temporal dynamics, and the hybrid model combining these strengths. The CNN model, consisting of 31 layers, attained the highest accuracy, achieving 91\% on the first benchmark dataset and 82\% on the second dataset using a 30-second threshold, selected for its clinical significance. In the work of Abdulwahhab et al. \citep{abdulwahhab_detection_2024}, EEG waves' time-frequency image and raw EEG waves served as input elements for CNN and LSTM models. Two signal processing methods, namely Short-Time Fourier Transform (STFT) and CWT, were employed to generate spectrogram and scalogram images, sized at 77 × 75 and 32 × 32, respectively. The experimental findings demonstrated detection accuracies of 99.57\% and 99.26\% for CNN inputs using CWT Scalograms on the Bonn University dataset and 99.57\% and 97.12\% using STFT spectrograms on the CHB-MIT dataset. Similarly, in emotion recognition, several deep learning models have been leveraged with EEG signals. For instance, in \citep{pandey_subject_2022},  a subject-independent emotion recognition model was proposed, which utilizes Variational Mode Decomposition (VMD) for feature extraction and DNN as the classifier. Evaluation against the benchmark DEAP dataset demonstrates superior performance of this approach compared to other techniques in subject-independent emotion recognition from EEG signals. 

Some researchers have also combined EEG signals with facial expression and speech in emotion recognition, as seen in \citep{hassouneh_development_2020}, \citep{pan_multimodal_2024}, and \citep{wang_multimodal_2023}, among others. In \citep{pan_multimodal_2024}, a multi-modal emotion recognition framework is proposed, utilizing three deep learning models to extract features from facial expressions, speech and EEG signals. To process EEG signals, the authors designed a tree-like LSTM model that extracts temporal features at different stages to capture multiscale feature representations. The output of each model is then weighted to generate the final predictions. In \citep{wang_multimodal_2023}, the feature maps extracted by the deep learning models are combined through concatenation, and then they are fed into a fully connected classifier to generate the predictions. For EEG signals, two deep learning models are designed: one for extracting local features and the other for extracting global features. Both models are implemented using convolutional layers. However, EEG signals can vary significantly across individuals, making it challenging to generalize models across different subjects.

\textbf{Generative Models}: GANs have been employed to enhance EEG signal classification. In \citep{song_eeggan-net_2024}, a conditional GAN model is proposed  to overcome the limited number of EEG samples. To improve data generation quality, the discriminator is integrated with SE attention and a cropped training strategy is employed to leverage the entire spectrum of the EEG signals. In \citep{corley_deep_2025}, a Wasserstein GAN model is employed to enhance the spatial resolution of EEG signals for EEG classification. The model takes low-resolution signals and generates the corresponding high-resolution signals while simulataneously interpolating the missing channels. EEG signals are often corrupted with noise and artifacts. DHCT-GAN is a dual-branch hybrid model designed to generate denoised EEG signals by processing both clean and noisy/artifact-corrupted signals \citep{cai_dhct-gan_2025}. The generator consists of dual-branch feature learning pipelines, where one accepts clean signals while the other handles noisy/artifact signals to learn feature representations. The feature maps are then fused using two fully connected layers with a tanh activation function to generate denoised EEG signals. Both feature learning branches are identical, consisting of convolutional layers with local and global self-attention mechanisms, while the discriminators are implemented using convolutional and fully connected layers. Table \ref{table_summary_eeg_studies} lists the summary of state-of-the-art EEG classification.

\begin{table*}[htbp]
\centering
\caption{Summary of state-of-the-art EEG classification.}
\scriptsize
\begin{tabularx}{\textwidth}{p{1.5cm} X X}
 \hline
 Ref. & Description & Results (Datasets and Metrics) \\ 
 \hline
 \citep{hermawan_multi_2024}, 2024 & A hybrid CNN-LSTM model consists of a convolutional layer to extract local features and two LSTM layers for temporal modeling & UPenn and Mayo Clinic's Seizure Detection Challenge\newline Accuracy: 89\% \newline F1-score (avg.): 87\% \\ 
 \hline
 \citep{pan_multimodal_2024}, 2024 & A tree-like LSTM to capture multiscale temporal feature representations from EEG signals & MAHNOB-HCI\newline Accuracy: 0.9450 (valence), 0.7628 (arousal) \\
 \hline
\citep{wang_multimodal_2023}, 2023 & Parallel convolutional pipelines for extracting local and global features from EEG signals & DEAP\newline Accuracy: 85.72\% (valence), 87.97\% (arousal)\newline MAHNO B-HCI\newline Accuracy: 85.98\% (valence), 85.23\% (arousal) \\
 \hline
 \citep{abdulwahhab_detection_2024}, 2024 & A framework comprises two separated deep learning-based models using a 2D Convolutional Neural Network and an LSTM to process EEG signal images and normalized EEG signals & Bonn\newline Accuracy: 99.75\% \newline CHB-MIT\newline Accuracy: 97.12\% \\
 \hline
 \citep{song_eeggan-net_2024}, 2024 & A conditional GAN integrated with SE attention in the discriminator and employing a cropped training strategy for improved data generation & BCI Competition IV 2a\newline Accuracy: 81.3\% \newline BCI Competition IV 2b\newline Accuracy: 90.3\% \\ 
 \hline
 \citep{corley_deep_2025}, 2025 & A Wasserstein GAN model is used to enhance the spatial resolution of EEG signals for classification by generating high-resolution signals from low-resolution inputs while simultaneously interpolating missing channels, addressing noise and artifact contamination & Dataset V (Berlin BCI)\newline Accuracy: 83.88\% (scale 2), 82.00\% (scale 4) \\ 
 \hline
 \citep{cai_dhct-gan_2025}, 2025 & A dual-branch hybrid model that denoises EEG signals by separately learning features from clean and noisy inputs, fusing them through fully connected layers, and leveraging convolutional layers with self-attention mechanisms & EEGdenoiseNet(EEG, EOG, EMG), MIT-BIH Arrhythmia (ECG) and semi-simulated EEG/EOG\newline $\eta$ (artifact reduction): 82.35\% (EMG), 91.80\% (EOG), 88.97\% (ECG), 82.07\% (EOG+EMG)  \\ 
 \hline
\end{tabularx}
\label{table_summary_eeg_studies}
\end{table*}

\subsection{Robotics}
Deep learning is a crucial component in modern robotics, enabling machines to perceive and interpret their surroundings. By leveraging sensors and cameras, robots can identify objects, navigate environments and interact with humans more effectively.

\subsubsection{Object Identification}
Deep learning plays a crucial role in object identification, allowing robots to recognize, classify, and differentiate between various objects in real-time. In \citep{le_application_2025}, an intelligent system, integrating deep learning with robotic kinematic control is proposed for waste classification. The system consists of a robotic arm with a gripper and a camera placed in front of the robot to capture real-time video. YOLO model is trained to recognize and localize objects, allowing the system to determine the location, pick them up and classify them. A similar system is reported in \citep{vukicevic_versatile_2025}, where a robotic arm is used to pick up objects on a conveyor and a camera is mounted above the conveyor to capture real-time video. The waste sorting process is divided into two tasks: localization of the waste objects and their classification. The object localization is performed using Segment Anything architecture (SAM) \citep{kirillov_segment_2023} which is based on vision transformer. Several lightweight SAM models are proposed to solve the localization task efficiently while minimizing computational requirements. For classification, object images are cropped using SAM's output masks and classified by a MobileNetV2.

A deep learning-based method is proposed to detect tile edges and corners for tile-paving mobile robotic systems \citep{liu_automatic_2025}. The camera is mounted on a tile-grabing platform that is parallel to the ground and approximately 10 to 20 cm above it. The system utilizes a YOLOv8 for semantic segmentation with rule-based post-processing techniques. An unmanned ground vehicle-based is proposed for automated structural damage inspection \citep{ge_deep_2025}. The robotic system utilizes LiDAR and a camera, integrating them via Robot Operating System for efficient control and data processing. The images are processed for the damage detection and segmentation using an improved YOLOv7 model, while the point cloud data is processed for real-time localization and 3D mapping using an enhanced Keep It Small and Simple-Iterative Closest Point algorithm. A similar system is reported in \citep{dai_advanced_2024}, where object detection and tracking approach is proposed to fuse point cloud and visual data. The approach leverage YOLOv8 real-time object detection capabilities, while LiDAR provides accurate 3D spatial information and distance measurements, enhancing perception reliability. A fusion method aligns LiDAR data with camera images, enabling accurate 3D bounding boxes and object tracking. Table \ref{table_summary_objidn_studies} lists the summary of state-of-the-art robot object identification.

\begin{table*}[htbp]
\centering
\caption{Summary of state-of-the-art robot object identification.}
\scriptsize
\begin{tabularx}{\textwidth}{p{1.5cm} X X X}
 \hline
 Ref. & Hardware platform & Model & Task \\ 
 \hline
 \citep{le_application_2025}, 2025 & Robotic arm, AI Stereo ZED 2 camera, gripper, Intel Xeon E5 series CPU @ 2.40 GHz & YOLOv4 & Waste classification \\ 
 \hline
 \citep{vukicevic_versatile_2025}, 2025 & PC workstation, industrial camera, AMD Threadripper 3970X 32 cores CPU @ 3.79 GHz, 128 GB RAM, two Titan RTX 24 GB GPUs & Segment Anything architecture (SAM) & Waste sorting \\
 \hline
\citep{kirillov_segment_2023}, 2023 & Mobile robot, camera, Nvidia RTX 3060 GPU & YOLOv8 & Tile edges and corners detection \\
 \hline
 \citep{ge_deep_2025}, 2025 & Husky Unmanned ground vehicle, Intel Realsense camera, Velodyne VLP-16 LiDAR, Intel i7 CPU & YOLOv7 & Structural damage inspection \\
 \hline
 \citep{dai_advanced_2024}, 2024 & Agilex Scout Mini robot, Velodyne LiDAR, Intel D435 camera & YOLOv8 & Object tracking \\ 
 \hline
\end{tabularx}
\label{table_summary_objidn_studies}
\end{table*}

\subsubsection{Path Extraction and Navigation}
Deep learning models have been applied for robot navigation systems. In \citep{alotaibi_deep_2024}, LiDAR and a camera are to capture point cloud and visual data for a comprehensive understanding and navigation in the environments. The study experiments with Faster R-CNN, YOLOv5 and YOLOv8 for object detection. Additionally, the paper compares the performance of these systems in various real-world environments, showcasing their potential to enhance autonomous navigation. Mobile robot navigation system is proposed utilizing semantic segmentation to determine the drivable paths \citep{misir_drivable_2024}. First, Deeplabv3 \citep{chen_rethinking_2017} model with a ResNet-50 backbone is applied to segment scene images to extract drivable area. A perspective transformation then maps the segmented images to real-world space. Following the transformation, the image is divided into grids to determine the optimal path while avoiding obstacles. Finally, a PID controller guides the robot along the smoothed path to ensure accurate navigation in the environment.

In \citep{cao_orchard_2024}, a method for orchard robot navigation utilizing a modified YOLOv8 model is proposed. The model locates vine tree trunks and identifies the center points of tree trunks at the lower end of the detection boxes. The least squares method is then applied to fit reference lines on both sides of the trunk. Furthermore, a multiscale attention mechanism is incorporated into the model to prioritize relevant features, enhancing the model performance. In \citep{liu_single-stage_2025}, deep learning is employed for navigation path extraction for agricultural robots. The authors propose Single-Stage Navigation Path Extraction Network (NPENet) to predict the road centerline in a single stage, eliminating the need for complex multi-stage processes such as line detection and segmentation. The proposed model uses residual blocks for feature extraction which includes batch normalization, leaky ReLU and convolutional layers, generating three key outputs: detecting if the robot is on the road, estimating the navigation line's angle and predicting the vanishing point of the road. These outputs enable the robot to determine its trajectory and real-time navigation in orchard environments. Table \ref{table_summary_nav_studies} lists the summary of path extraction and navigation.

\begin{table*}[htbp]
\centering
\caption{Summary of state-of-the-art path extraction and navigation.}
\scriptsize
\begin{tabularx}{\textwidth}{p{1.5cm} X X X}
 \hline
 Ref. & Hardware platform & Model & Task \\ 
 \hline
 \citep{alotaibi_deep_2024}, 2024 & Mobile robot, 2MP Logitech webcam, LiDAR, Raspberry Pi 4, 4 GB RAM & YOLOv8 & Semantic navigation \\ 
 \hline
 \citep{misir_drivable_2024}, 2024 & Mobile robot, camera, Nvidia Jetson Nano & Deeplabv3+ (ResNet50 backbone) & Path planning \\
 \hline
\citep{cao_orchard_2024}, 2024 & Mobile robot, Intel RealSense D455 camera, Intel Core i7 10870H CPU @ 2.20 GHz, Nvidia GeForce GTX 1650 Ti GPU  & YOLOv8 & Navigation line extraction \\
 \hline
 \citep{liu_single-stage_2025}, 2025 & Mobile robot, Intel RealSense cameras, Nvidia Jetson Xavier & NPENet & Navigation path extraction \\
 \hline
\end{tabularx}
\label{table_summary_nav_studies}
\end{table*}

\subsubsection{Human-Robot Interaction}
Deep learning has significantly advanced human-robot interaction, particularly through the integration of large language models (LLMs) designed for natural language processing. Despite their extensive knowledge, LLMs have limitations, as they can sometimes generate inaccurate information, a phenomenon known as hallucination \citep{huang_survey_2025}. To address this issue, LLMs need to be connected to the physical world by providing real-world contextual information to the models, enabling them to generate accurate responses that are relevant to the situation. This process is known as LLM grounding. In \citep{asuzu_humanrobot_2025}, the authors explore the use of LLMs in human-robot interaction, focusing on collaborative planning between humans and robots. The LLM is guided to generate outputs within the predefined set of robot skills using a few-shot prompting, while vision models such as SAM-CLiP and ViLD are used for object identification within the scene, enabling the robot to perform actions in the real-world. By leveraging LLMs, robots can understand and process natural language instructions and execute the corresponding actions, enabling efficient collaboration with human users. LLMs need to be aligned with human intentions to ensure effective communication and interaction between humans and robots. In \citep{chen_synergai_2024}, a 3D scene graph is employed to represent objects and environments captured by 3D instance segmentation and 2D image classification. Given the 3D scene graph, a complex task is decomposed using LLM through step-by-step reasoning by progressively retrieving relevant nodes from the graph. During the task solving, human intent is automatically identified, and the system generates responses accordingly.

\begin{figure*}[h!]
    \centering
    \includegraphics[scale=0.5]{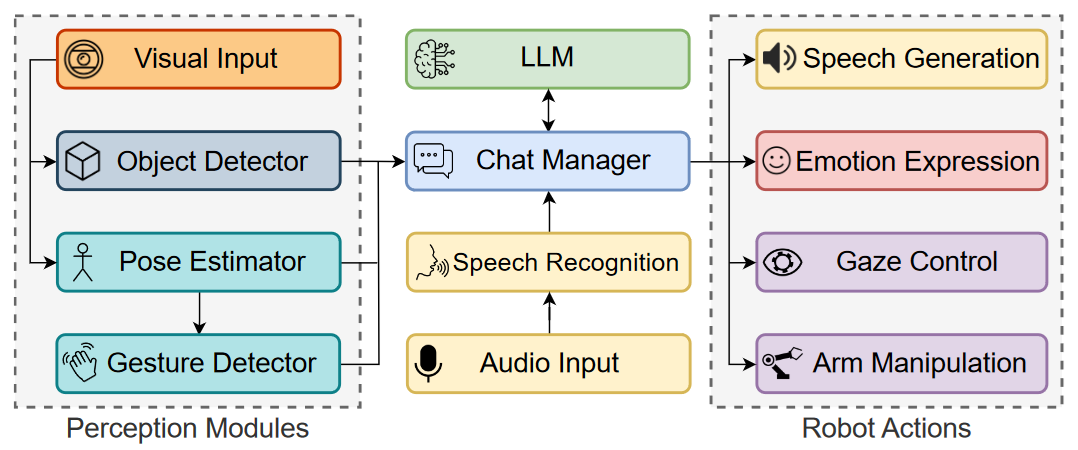}
    \caption{A grounded chat architecture \citep{allgeuer_when_2024}.}
    \label{fig_grounded_chat_architecture}
\end{figure*}

In \citep{allgeuer_when_2024}, a modular system architecture that can be easily extended is proposed to integrate LLM models for human-robot conversation and collaboration. Figure \ref{fig_grounded_chat_architecture} shows the system architecture, where the chat manager is the central to the system that coordinates the state and inference of the LLM model.  The robot perception such as recognizing objects and human gesture is handled separately by different deep learning models while robot actions are operated by speech generation model and actuators for emotion expression, gaze control and arm movement. This approach allows the robot to engage in open-ended conversation and collaborate naturally with humans. Grounding LLMs can be challenging due to robot's multi-modal perception system. Furthermore, the incoming data has different sampling rate which makes data alignment difficult, causing the robotic system to miss valuable information. In \citep{wang_i_2024}, a framework is proposed for processing multi-modal inputs to generate coherent narratives about robot experiences including its internal status, current observation, planning and critical alert. The framework is divided into three stages: key event selection, experience summarization and narration generation modules. First, multi-modal data across three categories: environment, internal and planning are aligned by sampling at a fixed rate and associating each frame at the nearest timestamp. Based on the aligned data, key events are selected by detecting changes in optical flow, RGB images and joint states using heuristic and normalization techniques to identify moment of interest. Using the detected events, the robotic data is converted into natural language summaries, and an LLM generates user-friendly explanation by filtering repetitive and irrelevant information. Table \ref{table_summary_hri_studies} lists the summary of human-robot interaction.

\begin{table*}[htbp]
\centering
\caption{Summary of state-of-the-art human-robot interaction.}
\scriptsize
\begin{tabularx}{\textwidth}{p{1.5cm} X X X}
 \hline
 Ref. & Hardware platform & Model & Task \\ 
 \hline
 \citep{asuzu_humanrobot_2025}, 2025 & Niryo Ned robot and Intel RealSense D435i camera & LLM: GPT-3.5,\newline Object detection: SAM-CLiP and ViLD & Robot planning \\ 
 \hline
 \citep{chen_synergai_2024}, 2024 & N.A. & LLM: GPT-4 turbo & Human collaboration \\
 \hline
\citep{allgeuer_when_2024}, 2024 & Neuro-Inspired COLlaborator (NICOL) robot  & LLM: GPT, Vicuna, Mistral\newline Human detection: YOLOX\newline Pose estimation: HRNet\newline Object detection: ViLD & Conversation and collaboration \\
 \hline
 \citep{wang_i_2024}, 2024 & MStretch SE3 robot, Intel RealSense D435i and D405 & Object segmentation: YOLO World & Robot behavior announcement \\
 \hline
\end{tabularx}
\label{table_summary_hri_studies}
\end{table*}
\renewcommand{\arraystretch}{1.0}  

\section{Research Challenges}\label{sec5}
Despite the success of deep learning in various applications, there are still fundamental challenges faced by researchers and practitioners. This section discusses the challenges ethical, technical and domain-specific perspectives.

\subsection{Ethical Issues}
Interpretability and explainability is crucial for building trust and understanding how predictive models make decisions, especially in high-stake applications such as healthcare and medical image analysis \citep{tonekaboni_what_2019}. However, as deep learning models become more intricate and complex with numerous layers, subnetworks and a large number of parameters, the models are often perceived as a “black box” and difficult to explain in terms of decision-making processes. Therefore, it is crucial for the researchers to focus on methods that provide insights into how a deep learning model performs the prediction and how its decisions are influenced by the input data, making it more transparent and trustworthy. Numerous methods have been proposed for interpreting and explaining the decisions of deep learning models, which can be categorized into visualization (feature attribution), model distillation and intrinsic (explainable by itself). Visualization methods involve the use of scientific visualization such as saliency maps or heatmaps to express the explanation by highlighting the degree of association between the inputs and the predictions \citep{tjoa_enhancing_2023}. The heatmaps identify the saliency of the input features influencing the model’s predictions. The visualization approach is simple and intuitive and can be applied to tabular data and image data. Furthermore, it can be used to identify and debug issues in deep learning models, leading to improved performance and robustness.

Model distillation is an approach to approximating a complex model by fitting a simpler model using the training set. The simpler model is built typically using a simpler or interpretable algorithm such as linear regression, decision tree or rule-based methods \citep{li_hybrid_2024}. In this approach, the simpler model is trained to resemble the predictive behavior of the complex model. Then, the simpler model may serve as the proxy or surrogate model for explaining the complex model. Model distillation can be used together with visualization to further enhance the interpretability of the complex model \citep{termritthikun_explainable_2023}. Model distillation seeks explanations of the models that were never designed to be explainable. Ideally, the explanation of a deep learning model’s prediction should be included as part of the model output, or the explanation can be derived from the architecture of the model. This is because an intrinsic model can learn not only the mapping between the input and output, but also generate an explanation of the prediction that is faithful to the model’s behavior. Attention mechanisms are the key to this approach, providing a form of attention weights that can be used to explain why the model made a particular decision \citep{xiong_explainable_2022}. Another type of intrinsic approach is to train the model to simultaneously perform the prediction task and generate the explanation for its predictions \citep{fernandes_intrinsic_2023}. This “additional task” can be in the form of a text explanation or model prototype which embeds the semantic meaning of the prediction. However, the intrinsic approach is more difficult to apply because the user needs additional knowledge and understanding of the model's architecture and inner workings.

Deep learning models are increasingly being deployed in making high-stake decision including recruitment \citep{freire_e-recruitment_2021}, criminal justice \citep{dass_detecting_2023} and credit scoring \citep{gicic2023intelligent}. There are several advantages of deep learning-based systems in which, unlike humans, machines are able to process vast amounts of data and applications quickly and consistently. However, deep learning-based systems have the risk of being prone to biases present in the data used for training which can lead to unfairness and injustice. Numerous efforts have been made to mitigate this issue which can be categorized into modelling bias detection and modelling bias mitigation. Detection of modelling bias refers to the process of identifying and quantifying biases that may present in predictive models. This approach involves the use of statistical analysis, fairness metrics, counterfactual testing and human review to detect bias in the models. For instance, visualization-based methods such as attribution maps are used to indicate which regions are significant to the predictions \citep{schaaf_towards_2021}. This in turn can be used to detect and quantify bias using metrics such as Relevance Mass Accuracy, Relevance Rank Accuracy and Area over the perturbation curve (AOPC). In \citep{giloni_benn_2022}, two modules are presented for estimating bias in predictive models. The first module utilizes an unsupervised deep neural network with a custom loss function to generate hidden representation of the input data called bias vectors, revealing the underlying bias of each feature. The second module combines these bias vectors into a single vector representing the bias estimation of each feature, achieved by aggregating them using the absolute averaging operation. 

Bias mitigation refers to the process of reducing the presence of bias in predictive models, which can be done in three stages. The first stage combats bias by modifying the training data, either relabeling the labels or perturbing the feature values \citep{iosifidis_fairness-enhancing_2019,kehrenberg_tuning_2020}. The second stage addresses bias during the training of the model by applying regularization terms to the loss function to penalize discrimination. In \citep{jain_increasing_2023}, a loss function based on bias parity score (BPS) is introduced to measure the degree of similarity of a statistical measure such as accuracy across different subgroups. The BPS term is added to the loss function as a regularizer to the original prediction task. The last stage mitigates bias after the predictive models have been successfully trained. This stage applies post-processing approaches such as reinforcement learning to obtain a fairer model \citep{yang_algorithmic_2023}. For instance, the detection of minority classes is rewarded to prevent bias towards the majority class. This allows the model to generalize well across different patient demographics.

\subsection{Technical Issues}
Even though deep learning architectures have achieved state-of-the-art across various computer vision tasks, they often come with large model parameters \citep{raiaan_systematic_2024}. The architecture and complexity of a deep learning network determine the number of model parameters. The deeper the network, the larger the number of model parameters. However, deep learning models with large parameters often suffer limitations when deploying on end devices. For instance, a deep learning model developed for security monitoring by analyzing video data using 3D-CNNs might suffer deployment issues when deploying such models on low-resourced systems like smartphones or small-scale IoT devices. Model training and inference for deep learning models with large parameters demands substantial processing power. As the number of parameters increases, so does the computational complexity, resulting in longer model training duration and more hardware needs. Furthermore, large parameter sizes translate to increased memory requirements, limiting their deployment on end devices. This is because these end devices often have battery, processor, or memory capacity limitations. To address these challenges, it is important to develop sophisticated but lightweight architectures that can achieve state-of-the-art with few model parameters. Such lightweight models will be characterized by their ability to deliver competitive performance while mitigating computational complexity and memory requirements, making them well-suited for deployment on resource-constrained devices. An approach would be to develop novel lightweight plug-and-play modules that can be plugged to a few layered deep learning architectures to improve feature learning without incurring additional model complexity. Other approaches could involve leveraging model compression techniques to reduce the size and computational complexity of deep learning models. Researchers can focus on improving pruning methods \citep{pruning8794944}, which can identify and eliminate redundant parameters or connections, thereby reducing the model's footprint without compromising performance. Furthermore, quantization techniques \citep{Yang_CVPR} can be further explored to reduce the precision of weights and activations, therefore, enabling efficient representation with lower memory requirements. Also, knowledge distillation techniques \citep{NEURIPS2021_376c6b9f} can be further investigated to facilitate the transfer of knowledge from a complex teacher model to a simpler student model, therefore, enabling compact yet effective representations. 

Deep learning training involves complex processes that require efficient optimization to ensure fast convergence and resource management. As models grow in size and complexity, challenges such as slow convergence, vanishing gradients, and computational limitations become more pronounced. Researchers have explored various methods to overcome these issues, proposing innovative approaches like predicting parameter change and incorporating them into training, thus reducing training time and improving the model performance \citep{ying_enhancing_2024}. Researchers also leverage prior experience to optimize parameter adjustments to reduce forward and back propagation steps, thus reducing computational costs and making training more efficient \citep{wang_optimizing_2024}. Others have proposed alternative training algorithms such as alternating minimization \citep{yan_triple-inertial_2025} and random search based on the annealing method \citep{krasnoproshin_random_2024}. Overall, the researchers aim to accelerate learning and improve efficiency either by decomposing the optimization problem into sequential sub-problems or by refining the search space based on parameter value ranges.

Adversarial attacks and defense mechanisms in deep learning represent a critical area of research and development, particularly as deep learning models become increasingly integrated into various applications. Adversarial attacks involve the deliberate manipulation of input data to mislead or deceive deep learning models, leading to incorrect predictions or behavior \citep{adversarial2018}. Szegedy et al. \citep{szegedy2013intriguing} was the first to identify this intriguing shortcoming of deep neural networks in image classification. They showed that even with their great accuracy, deep learning models are surprisingly vulnerable to adversarial attacks that take the form of tiny image changes that are (almost) invisible to human vision systems. A neural network classifier may radically alter its prediction about an image as a result of such an attack. Furthermore, such a model can indicate high confidence in wrong predictions, which can be catastrophic for deep learning models deployed in medical or security fields, among many others. In generative models, several studies have investigated how adversarial attacks affect autoencoders and GANs, as seen in Tabacof et al. \citep{tabacof2016adversarial} where a method to manipulate input images in a way that deceives variational autoencoders into reconstructing an entirely different image was introduced. In recent times, the focus of adversarial attack research has been on images, but studies have shown that adversarial attacks are not limited to image data; they can also affect other types of data such as text, signals, audio, and video \citep{zhang2020adversarial, jiang2019black, esmaeilpour2019robust}.

\subsection{Domain-specific Issues}
Building and employing deep learning models face several challenges. The training of deep learning requires a large number of instances (examples) to achieve high accuracy and generalization \citep{munappy_data_2022}. Furthermore, the complexity of deep neural networks may lead to overfitting, where the model performs well on training data but fails to generalize on new, unseen data. This phenomenon frequently arises when the models are trained on insufficient data, highlighting the importance of diverse and extensive datasets. However, the data collection is time-consuming and often require domain experts, specialized training and standardization \citep{luca_impact_2022}. Moreover, this process is prone to error and has the risk of introducing biases into the dataset which can significantly impact the performance of the trained model. One of the approaches to address this issue is transfer learning. Transfer learning involves the use of a deep learning model (known as pre-trained model) that is trained on a large dataset for solving a specific task (with a small dataset) \citep{zhuang_comprehensive_2020}. The pre-trained model serves as a basis for the model training by fine-tuning the weights of the pre-trained model and adapting it to the new prediction task. This approach helps to mitigate the lack of training data in the target domain. Furthermore, transfer learning reduces computational resources required to train the model and helps faster convergence. 

Another approach that can be employed to address the lack of data is data augmentation. Data augmentation is a convenient method that increases the number of instances by performing transformation functions on the existing instances without changing the labels \citep{mumuni_data_2022}. In the domain of computer vision, image transformation such as rotation, translation and cropping. However, it is important to consider the output of the transformation because the resultant may not represent the actual data. For example, flipping or adding noise to a signal may introduce distortion or changing the characteristics (trend, seasonality and cyclic variations) of the signal. Thus, careful consideration must be given to ensure that the generated instances still accurately represent the underlying patterns present in the data. Data augmentation can also be realized by generating synthetic data to supplement the training set. Synthetic data is artificially created data that resembles real data but is generated using statistical methods or deep generative models \citep{hu_survey_2023,murtaza_synthetic_2023}. The generated data can complement the less-diverse, limited datasets, providing a broader range of examples for the model to learn from. However, generating synthetic data that accurately reflects the characteristics of the real-world data is challenging. Careful consideration must be given to the choice of models and parameters used to ensure the synthetic data is realistic and representative of the real-world data.

Data annotation is expensive and time-consuming. The problem is exacerbated when the data is of low quality corrupted with noise which may lead to bad, resulting in unreliable training data. One of the approaches addresses this issue is active learning. Active learning is an approach where the learning algorithm selectively queries the most informative data points for labeling. The aim is to improve predictive model efficiency with fewer training data, thus reducing the overall cost of data annotation. The approach involves training a model using an initial training set, and then a subset of unlabeled data is selected for labeling by external annotators. The newly labeled instances are appended to the training set for retraining or fine-tuning. This process is repeated until the model performance reaches a desired threshold or the labeling budget has exhausted. The core principle behind active learning is the query strategy or, essentially how to select the most informative data for labeling that will be beneficial to the model training. In general, the query may select the most ambiguous instances based on the model predictions, instances that will have a significant impact on the model performance, or instances that cover the distribution of the entire feature space \citep{li_survey_2024}. The dynamic and sequential nature of real-world applications presents different challenges, which led to the development of online active learning. This approach was introduced to address specific issues, including data streams, concept drift and environmental changes \citep{cacciarelli_active_2024}. Recently, researchers have investigated the use of LLMs for data annotation. The study presents methods for generating annotations, assessing annotations and utilizing annotations \citep{tan_large_2024}.

Self-supervised learning is another approach that was introduced to mitigate the issue of limited labeled datasets. In self-supervised learning, the model learns from unlabeled data by generating its own labels from the input data \citep{gui_survey_2024}. This approach essentially creates a pretext task which the model solves without requiring manual annotations or labeled datasets. By solving the pretext task, the model leverages the underlying structure in the data and learn useful representations that can be used for specific tasks. For example, in computer vision, a model might predict the speed \citep{altabrawee_stclr_2025} or repeated scenes \citep{altabrawee_repeat_2024} in a video. In natural language processing, it could predict missing words in sentences to learn the dependencies between words \citep{lee_predicting_2021}. The key advantage of self-supervised learning is that it allows the model to take advantage of large amounts of unlabeled data, which is often much more abundant and cheaper to collect than labeled data. The learned representations can then be used for downstream tasks like classification, regression, or segmentation.

\section{Summary and Future Directions}\label{sec6}
We have discussed the state-of-the-art applications and challenges of deep learning in computer vision, natural language processing and time series analysis. In this section, we summarize the advancements made which can be grouped into model architecture, contextual enhancement, and loss function and optimization. Finally, we discuss possible future works in these domains.

\textbf{Model Architecture}: The recent advancements in computer vision focus on vision transformers (ViT) for scalable representation learning \citep{dosovitskiy_image_2021}. In image classification and image segmentation, ViTs are utilized to capture global context and relationships between different parts of an image, while in object detection, ViTs are used to streamline the detection process by eliminating the traditional components like anchor boxes and non-maximum suppression \citep{carion_end_end_2020}. However, ViTs lack the ability to exploit local spatial features and struggle with hyperparameter sensitivity and performance on smaller datasets. To address this issue, researchers have developed hybrid deep learning architectures such as conformer \citep{peng_conformer_2023} and MaxViT \citep{tu_maxvit_2022} that combine both transformers and convolutional neural networks to capture both local and global features. Similar trend is observed in modern natural language processing whereby the basis of the deep learning models is transformer architectures such as BERT \citep{devlin_bert_2018} and its variants to achieve remarkable accuracy in tasks such as text classification, machine translation and text generation.

While recent advancements in computer vision and natural language processing have predominantly relied on transformer architectures, the field of time series analysis focuses on a different trajectory, where transformers research is lagging due to fundamental challenges in data structure, sequence length and dataset size and availability. Notably, areas such as HAR and financial prediction have seen significant progress through the development of hybrid architectures that do not depend on transformers. Instead, these advancements leverage recurrent neural networks and convolutional neural networks, often incorporating attention mechanisms. The hybrid architectures allow the models to capture both local features and temporal dependencies in the data, which is crucial for time series analysis \citep{khan_attention_2021, ige_deep_2023}. The hybrid attention mechanisms improve the feature extraction process by dynamically weighting important channel, spatial and temporal features, enhancing model performance \citep{gao_danhar_2021, agac_resource-efficient_2024, tang_triple_2022}. GAN models are used to address data scarcity in HAR, where the generator consists of convolutional layers \citep{lupion_data_2024, kia_human_2024}, with LSTM \citep{chan_unified_2021} while the discriminator consists of convolutional layers.

Similar trends can be observed in ECG and EEG classification, where convolutional and recurrent neural networks are commonly used to build hybrid models \citep{alamatsaz_lightweight_2024, hermawan_multi_2024}, with attention mechanisms \citep{sun_arrhythmia_2024, chen_automated_2022}. Generative models such as variational autoencoder \citep{ dai2019eeg} and generative adversarial network are used to generate data to address challenges posed by limited datasets \citep{yang_data_2024, msigwa_iot-driven_2024, song_eeggan-net_2024}, poor quality \citep{corley_deep_2025}, noisy signals \citep{cai_dhct-gan_2025}, and overfitting \citep{zhu_electrocardiogram_2019}. The generative approach employs GAN only or with encoder-decoder \citep{yang_data_2024}, where the generator consists of convolutional layers with attention mechanisms or bidirectional LSTM, while the discriminator consists of convolutional layers with attention mechanisms.

\textbf{Contextual Enhancement}: Recent advancements in text classification focus on capturing contextual information utilizing approaches such as joint embeddings of labels \citep{wang_joint_2018} and words as well as aspect-aware methods that enhance feature extraction \citep{zhu_bert-based_2023}. Conversely, in machine translation, although transformers have proven effective, they often struggle to capture nuanced language features and contexts. To address this issue, various strategies have been introduced such as concatenating contextual sentences and context-sensitive training \citep{wu_study_2022, kim_towards_2023}. In text generation, additional contexts are provided to improve both the quality and diversity of the generated sentences such as incorporating the category context \citep{li_feature-aware_2023} and style context \citep{kwon_class_2024} into text generation process.

In image generation, the researchers focus on improving the quality and relevance of generated images by enhancing the alignment between input text and images, while also addressing issues like fuzzy shapes and diversity through innovative deep learning architectures such as incorporating attention mechanisms and fusion modules into the model. Furthermore, recent models like DF-GAN \citep{tao_df-gan_2022} and DMF-GAN \citep{yang_dmf-gan_2024} showcase a shift towards single-stage generators with regularization strategies that maintain details while enhancing diversity. In text generation, BERT and its variants are the basis of the deep learning models. The research focuses on hierarchical architectures, capturing long text dependencies \citep{ma_t-bertsum_2021} and leveraging domain knowledge through label sequence \citep{xie_pre-trained_2022} and conversation sequence \citep{li_incremental_2019} to bridge semantic gaps. Overall, the advancements show a clear trend toward improving contextual awareness and feature representation.

LLMs are increasingly being leveraged in robotics for improving human-robot interaction. The main challenge of integrating LLMs is bridging the gap between robots and the physical world to provide real-world contexts, a process known as LLM grounding. The concept of LLM grounding refers to how machines make sense of abstract language such as words and ideas by connecting them to objects they can actually experience or sense in the real world. To this end, sensors such as cameras and microphones, along with object detection, pose estimation and speech recognition methods, are used for semantic and perceptual grounding. The robot data are streamed are at varying sampling rates, so they need to be aligned and carefully selected to identify key events \citep{wang_i_2024}. The detected objects and their relationships are then modeled using 3D scene graph for scene understanding \citep{chen_synergai_2024}. Central to the grounding system is a module that collates and constructs the text prompts which are sent to the LLM \citep{allgeuer_when_2024}. The text prompts are typically formulated by integrating a set of predefined robot skills that can be executed \citep{asuzu_humanrobot_2025}.

\textbf{Loss Function and Optimization}: Researchers are increasingly focused on refining loss functions to improve classification accuracy. For instance, in image classification, some studies introduce additive terms to the cross-entropy loss to reward well-classified instances \citep{zhao_well-classified_2022}, while others propose asymmetric polynomial loss functions that prioritize positive instances to tackle class imbalance \citep{huang_asymmetric_2023} and adaptive loss function that dynamically adjust the weights assigned to class-level components based on model performance \citep{maldonado2023owadapt}. Another approach to address class imbalanced is data augmentation through generating virtual samples \citep{zhu2024irda}, oversampling with target-aware autoencoders for estimating target values for new features \citep{belhaouari2024oversampling}. In time series analysis, similar approach has been proposed such as penalizing misclassification of minority classes \citep{wang2024class}, maximizing the minimum recall of the classes \citep{ircio2023minimum}. Improved training strategies have been proposed to address class imbalanced data such as iteratively selecting the most informative instances \citep{moles2024exploring} and contrastive learning to keep the instances of each fine-grained clusters away from the minority class \citep{zhu2024sfpl}.

\textbf{Future Directions}: Looking ahead, the future of deep learning presents numerous opportunities for growth and innovation. Future models could explore non-sequential hybrid architectures of transformer and convolutional neural networks to leverage the strengths of both approaches, enhancing performance in image classification, object detection and image segmentation. Furthermore, researchers could investigate simpler or lightweight architectures and new training schemes to address the long training time associated with transformers. Future research in image generation may explore advanced frameworks or architectures that further integrate semantic understanding, perhaps by employing hierarchical attention modules and/or fusion modules to capture both local and global features more effectively. Additionally, incorporating unsupervised learning or self-supervised learning approaches could reduce reliance on labeled datasets, allowing models to learn from a diverse range of inputs. 

In the area of natural language processing, future research in text classification and machine translation could explore the integration of external knowledge bases and domain-specific embeddings may further improve context understanding and label alignment. Additionally, refining context incorporation methods such as dynamic context updating during translation may enhance the model performance. In text generation, advancing towards more interactive and adaptive systems that can maintain context over extended dialogues and narratives is crucial. Furthermore, frameworks that can leverage multi-modalities (text, audio, and visual) for richer contextual understanding may facilitate the development of more advanced applications across these domains. Future research can also focus on exploring adversarial attacks in these domains and developing tailored defense mechanisms. Furthermore, researchers can further investigate the practical implications of adversarial attacks in real-world scenarios, such as in autonomous vehicles and medical imaging. Understanding the potential impact of adversarial attacks in these applications can inform the development of more robust and secure systems.

Real-time processing capabilities are paramount, future work can develop efficient models or algorithms for real-time processing of HAR, speech recognition, ECG and EEG signals. This could involve optimizing existing architectures and leveraging hardware acceleration techniques to enable real-time inference on resource-constrained devices such as wearable sensors and implantable devices. Furthermore, future models could explore lightweight transformers to enhance feature representation for time series analysis. Due to the sensitive nature of financial research, future work can focus on enhancing the interpretability of deep learning models in financial predictions. Researchers should explore techniques to explain the predictions of models, to improve trust and understanding of model decisions, which is essential for adoption in HAR, speech recognition and finance. 

Future research in data generation for time series analysis and pervasive computing should focus on developing more advanced techniques to address data scarcity and improve model generalization. In HAR, frameworks that can generate diverse signals for subjects of different ages, genders, as well as physical abilities, may facilitate the development of models that can generalize across different populations and real-world scenarios \citep{jimale_subject_2023}. Moreover, the generated data should capture sensor noise, variations in user behavior, and diverse environmental conditions. In finance, synthetic market data should model complex dependencies, volatility patterns and anomalies to improve trend and risk predictions. For EEG and ECG classification, researchers could explore deep generative models that can create physiologically meaningful synthetic signals while preserving individual variability. Future models could explore methods for adapting or personalizing models to account for inter-subject variability and improve performance on individual subjects. Furthermore, EEG electrodes cover only a fraction of the brain's surface, resulting in limited coverage of neural activity. Deep learning models could investigate strategies to infer activity from unobserved brain regions or integrate information from multiple modalities to provide more comprehensive coverage. These areas can still be further explored.

Future research in robot object identification and navigation will focus on enhancing real-time perception, generalization across diverse environments, and increasing robustness to occlusions and dynamic obstacles. Furthermore, researchers can explore multimodal learning, integrating vision with LiDAR and tactile sensing to improve scene understanding. Advances in self-supervised learning and few-shot learning could enable robots to identify and navigate complex environments with minimal labelled data. LLMs will play a crucial role in human-robot interactions, facilitating natural language understanding, intent recognition and context-aware decision-making \citep{zhang_large_2023}. Researchers should optimize LLMs for real-time processing on robots and improve their grounding in physical environments. Additionally, future work can explore a deep learning-based navigation with LLMs to develop more autonomous, assistive and socially intelligent robots.

The future of AI lies beyond deep learning, moving towards a conscious intelligent system that mimics human cognition \citep{butlin_consciousness_2023}. Achieving this requires a significant advancement in hardware and software architectures, enabling systems to process information with greater efficiency, adaptability and reasoning capabilities. Beyond mere pattern recognition, future of AI will need to integrate symbolic reasoning, causal inference and common sense to truly emulate human thought \citep{marra_statistical_2024, colelough_neuro-symbolic_2025, bhuyan_neuro-symbolic_2024}. Future research will likely focus on hybrid AI models that combine deep learning with symbolic AI, enabling machines to understand abstract concepts, reason about cause and effect and apply knowledge across domains. Quantum computing, with its unique property of entanglement could exponentially accelerate deep learning training and inference \citep{wu_survey_2025, klusch_quantum_2024}. Future models could explore quantum neural networks which could process high dimensional data more effectively and discover patterns that are not possible by classical methods. Quantum-based optimizers could also solve problems that are infeasible for classical graphical processing units or tensor processing units. A more radical approach is neuromorphic computing that uses specialized hardware to mimic the structure and function of the human brain, enabling low-power inference, parallel processing and real-time learning and adaptation \citep{shrestha_survey_2022, kudithipudi_neuromorphic_2025}. This results in autonomous AI systems that can solve complex tasks requiring quick response time and adapt to changing situations. Finally, multi-modal intelligence enables AI to process and integrate multiple types of sensory input, including vision, speech and touch, providing a more holistic view of the world, similar to human perception \citep{fei_towards_2022}. 

\section{Conclusion}\label{sec7}
Deep learning has become the prominent data-driven approach in various state-of-the-art applications. Its importance lies in its ability to revolutionize many aspects of research and industries and tackle complex problems which were once impossible to overcome. Numerous surveys have been published on deep learning, reviewing the concepts, model architectures and applications. However, the studies do not discuss the emerging trends in the state-of-the-art applications of deep learning and emphasize the important traits and elements in the models. This paper presents a structured and comprehensive survey of deep learning, focusing on the latest trends and advancements in state-of-the-art applications such as computer vision, natural language processing, time series analysis and pervasive computing, and robotics. It explores key elements and traits in modern deep learning models, highlighting their significance in addressing complex challenges across diverse domains. The discussion also covers future research directions in advancing these fields. Furthermore, this paper presents a comprehensive review of the deep learning fundamentals, which is essential for understanding the core principles behind modern deep learning models. The survey finishes by discussing the critical challenges and future directions in deep learning.

\section*{Acknowledgements}
This work has been supported in part by the Ministry of Higher Education Malaysia for Fundamental Research Grant Scheme with Project Code: FRGS/1/2023/ICT02/USM/02/2.

\section*{Statements and Declarations}
\begin{itemize}
\item \textbf{Funding} This work was supported by the Ministry of Higher Education Malaysia (FRGS/1/2023/ICT02/USM/02/2).
\item \textbf{Competing Interests} The authors declare that they have no known competing financial interests or personal relationships that could have appeared to influence the work reported in this paper.
\item \textbf{Ethics approval} Not applicable.
\item \textbf{Availability of Data and Materials} No data was used for the research described in the article.
\end{itemize}



\bibliographystyle{elsarticle-num} 
\bibliography{elsarticle-template-num_v4}






\end{document}